\documentclass[10pt]{article}
\ifdefined\TafsutReview
  \usepackage{tmlr}
  
\else
  \usepackage[preprint]{tmlr}
  
\fi

\usepackage{microtype}
\usepackage{tabularx,placeins}
\usepackage{amsmath,amssymb}
\usepackage{booktabs}
\usepackage[table]{xcolor}
\usepackage{graphicx}
\usepackage{subcaption}

\newcommand{\vect}[1]{\pmb{#1}}
\newcommand{\mat}[1]{\pmb{#1}}
\DeclareMathOperator{\vectorize}{vec}
\newcommand{\ind}{\mathbb{I}}
\newcommand{\reals}{\mathbb{R}}
\definecolor{reportaccent}{HTML}{24524F}
\definecolor{ablationbetter}{HTML}{007F78}
\definecolor{ablationworse}{HTML}{A43D40}
\definecolor{ablationneutral}{HTML}{646B73}

\newcommand{\reporttablehead}[1]{\textcolor{reportaccent}{\textbf{#1}}}
\newcommand{\reporttablestyle}[1][2]{%
  \setlength{\aboverulesep}{1pt}%
  \setlength{\belowrulesep}{1pt}%
  \ifodd#1\relax
    \rowcolors{#1}{reportaccent!7}{}%
  \else
    \rowcolors{#1}{}{reportaccent!7}%
  \fi
}
\newcommand{\reporttablegroup}[2]{%
  \rowcolor{white}\multicolumn{#1}{@{}l}{\reporttablehead{#2}}%
}

\usepackage{hyperref}
\usepackage{url}

\newcommand{\TafsutPaperTitle}{\textnormal{\textsc{Tafsut}}: Temporal Abstractions for Forecasting through Syntheticity, Universality, and Transfer}
\title{\texorpdfstring{\raisebox{-.18\height}{\includegraphics[height=1.22em,trim=31bp 43bp 29bp 29bp,clip]{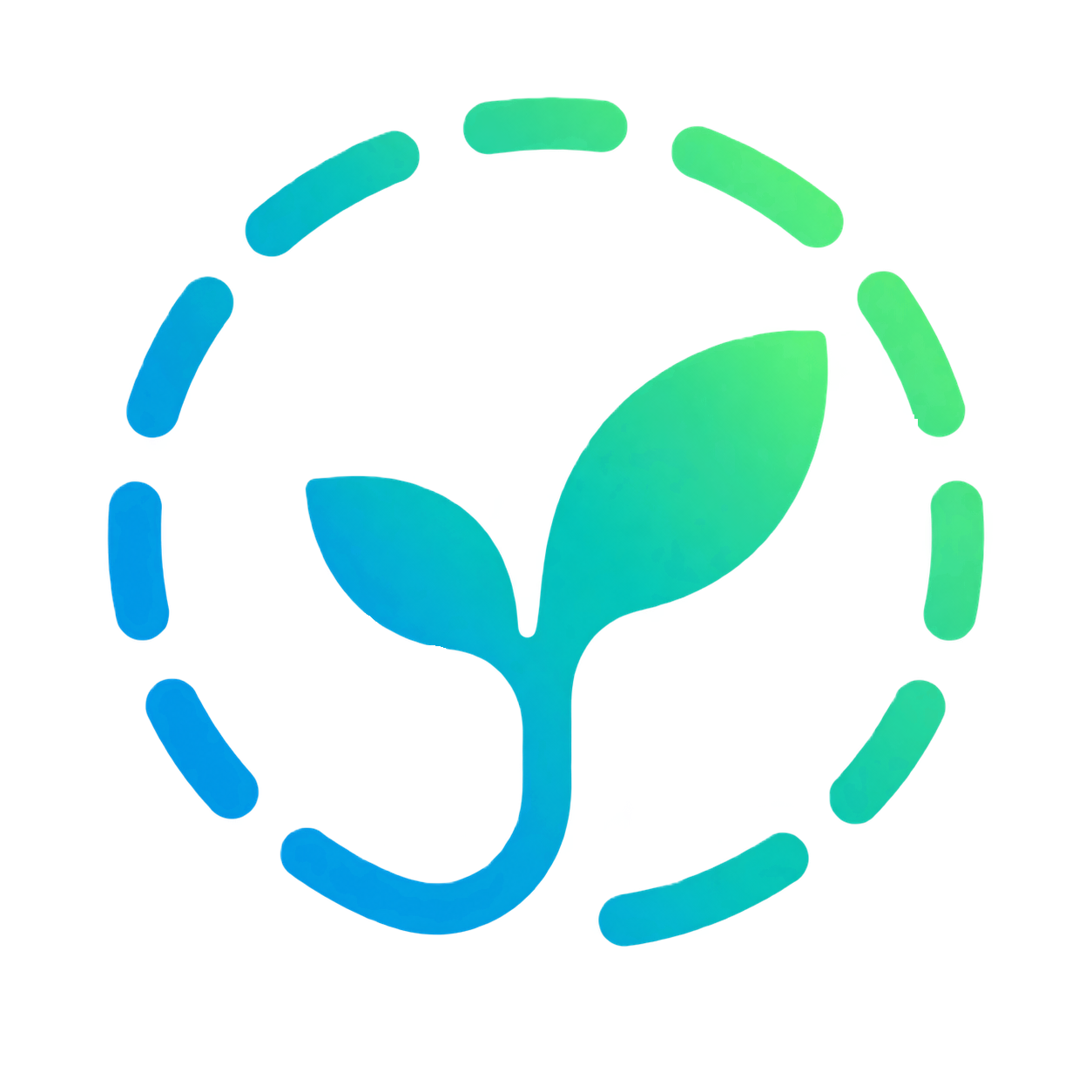}}%
\enspace}{}\TafsutPaperTitle}
\hypersetup{pdftitle={\TafsutPaperTitle},hidelinks}

\author{%
  \name Tareq Si Salem\thanks{Equal contribution.}\\
  \addr Huawei Fourier Research Center
  \AND
  \name Michel Doroch\footnotemark[1]\\
  \addr Huawei Fourier Research Center
  \AND
  \name Pietro Michiardi\\
  \addr EURECOM
  \AND
  \name Simone Rossi\\
  \addr EURECOM
}

\begin{document}
\maketitle
\begin{abstract}
We introduce \textsc{Tafsut}, a probabilistic time-series foundation model that combines synthetic pretraining with modular covariate adaptation. We release an open-source generator that composes white-box time-series processes and pretrain a univariate backbone exclusively on series it generates online. We freeze the backbone and train a separate adapter to refine its forecast quantiles using the target history and available covariates. The quantile interface lets the same adapter architecture work with different backbones that produce compatible forecasts. In univariate evaluations, the synthetic backbone ranks first for point-forecast and probabilistic accuracy among the fully synthetic models compared on both GIFT-Eval and FEV-Bench. After real-data adaptation, \textsc{Tafsut} is competitive with leading open-weight zero-shot forecasters. It ranks fourth for point-forecast accuracy and fifth for probabilistic forecast quality on GIFT-Eval, and eighth on both measures on FEV-Bench. We document the training recipe and data sources, including how we filter and audit separate adaptation corpora for overlap with each benchmark. Systematic ablations of generator composition, model architecture, and training choices explain how we selected the final \textsc{Tafsut} configuration.
\end{abstract}

\ifdefined\TafsutReview\else
  \noindent\href{https://github.com/Tafsut-FM/tafsut-timeseries-generator}{%
    \raisebox{-.25\height}{\includegraphics[height=3.6mm]{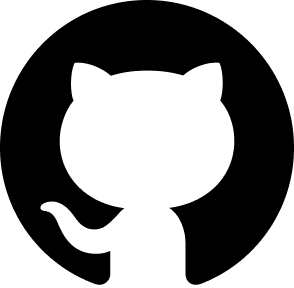}}%
    \enspace Synthetic Generator Code}%
    \qquad \href{https://github.com/Tafsut-FM/tafsut}{%
    \raisebox{-.25\height}{\includegraphics[height=3.6mm]{assets/github-logo.png}}%
    \enspace Inference Code}%
  \qquad\href{https://huggingface.co/Tafsut-FM}{%
    \raisebox{-.25\height}{\includegraphics[height=3.6mm]{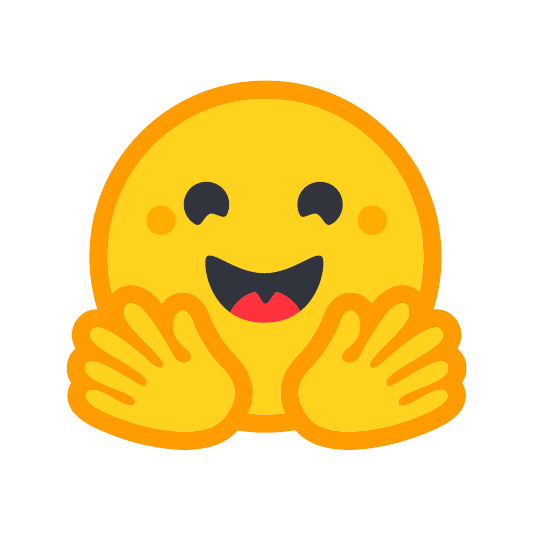}}%
    \enspace Model Weights}\par
\fi

\section{Introduction}
\label{sec:introduction}
Time-series foundation models aim to forecast across datasets without
task-specific training
\citep{garza2023timegpt,rasul2024lagllama,ansari2024chronos,das2024timesfm,woo2024moirai}.
Synthetic pretraining gives direct control over
the dynamics and observation artifacts a model encounters. Recent
forecasters such as Chronos-2, TiRex-2, TimesFM 3.0, Timestral, and TimeH2O
use both synthetic and real series in their training pipelines
\citep{ansari2025chronos2,podest2026tirex2,google2026timesfm3,petralia2026timestral,artley2026timeh2o}.
Evaluating synthetic-only models on real series measures transfer before
adaptation to real data.

Real-world forecasts can also depend on covariates such as weather
observations or planned promotions. Their value depends on what is
available at the forecast origin and how it relates to the target.
Evaluating transfer poses a separate challenge. Public corpora and
benchmarks can share series or resampled versions of the same source,
so zero-shot evaluation requires explicit overlap checks.

\begin{figure}[!t]
\centering
\captionsetup[subfigure]{skip=3pt,justification=centering}
\begin{subfigure}[t]{0.225\linewidth}
\centering\vspace{0pt}
\includegraphics[width=\linewidth]{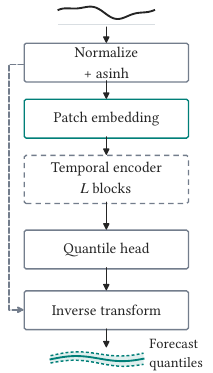}
\caption{Forecasting path}\label{fig:backbone-path}
\end{subfigure}\hfill
\begin{subfigure}[t]{0.225\linewidth}
\centering\vspace{0pt}
\includegraphics[width=\linewidth]{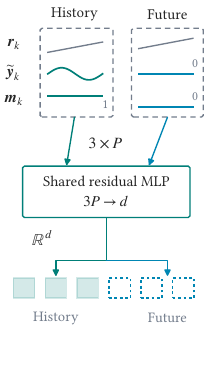}
\caption{Patch embedding}\label{fig:backbone-patches}
\end{subfigure}\hfill
\begin{subfigure}[t]{0.245\linewidth}
\centering\vspace{0pt}
\includegraphics[width=\linewidth]{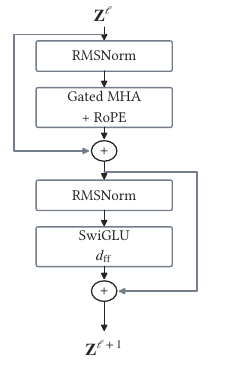}
\caption{Encoder block}\label{fig:backbone-block}
\end{subfigure}\hfill
\begin{subfigure}[t]{0.275\linewidth}
\centering\vspace{0pt}
\includegraphics[width=\linewidth]{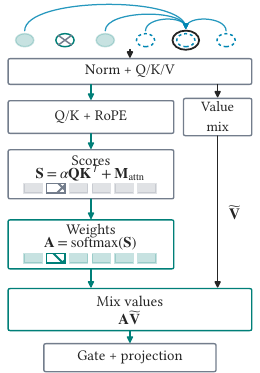}
\caption{Temporal attention flow}\label{fig:backbone-attention}
\end{subfigure}
\caption{\textsc{Tafsut}'s univariate backbone. A shared embedding maps historical
patches and blank future queries to tokens. The encoder attends across
both to forecast the full horizon. Panel (d) shows one attention head.
The ring marks a future query and the cross marks a masked key.
Table~\ref{tab:architecture-config}
lists the default settings.}
\label{fig:frozen-backbone-architecture}
\end{figure}

\begin{figure}[!t]
\centering
\captionsetup[subfigure]{skip=3pt,justification=centering}
\begin{subfigure}[t]{0.29\linewidth}
\centering\vspace{0pt}
\includegraphics[width=\linewidth]{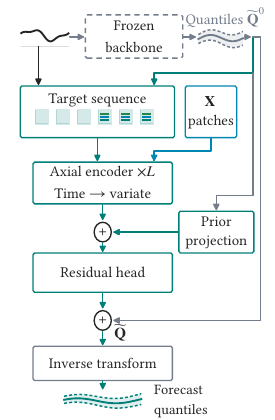}
\caption{Quantile flow and adaptation}\label{fig:adapter-path}
\end{subfigure}\hfill
\begin{subfigure}[t]{0.24\linewidth}
\centering\vspace{0pt}
\includegraphics[width=\linewidth]{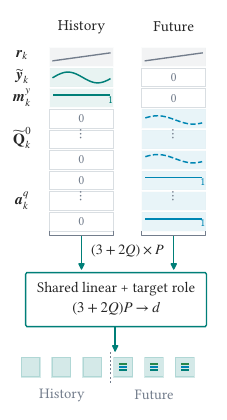}
\caption{Joint target patches}\label{fig:adapter-patches}
\end{subfigure}\hfill
\begin{subfigure}[t]{0.24\linewidth}
\centering\vspace{0pt}
\includegraphics[width=\linewidth]{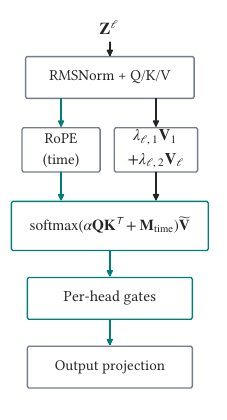}
\caption{Attention construction}\label{fig:adapter-attention}
\end{subfigure}\hfill
\begin{subfigure}[t]{0.20\linewidth}
\centering\vspace{0pt}
\includegraphics[width=\linewidth]{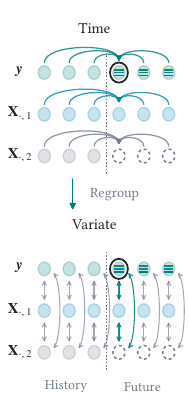}
\caption{Axial packing}\label{fig:adapter-axial-packing}
\end{subfigure}
\caption{\textsc{Tafsut}'s multivariate adapter. Frozen backbone quantiles enter the
future target patches and condition the residual head. The adapter attends
along each stream (c) and across variates at each patch position (d), then
adds its correction to the frozen forecast. Hollow tokens mark valid slots
with unavailable covariate values. Table~\ref{tab:architecture-config}
lists the default settings.}
\label{fig:joint-patch-adapter-architecture}
\end{figure}

We introduce \textsc{Tafsut}, a probabilistic forecasting model with a
patch-based univariate backbone pretrained on synthetic series. Its
generator composes temporal dynamics and observation mechanisms, and
separately defines relationships between targets and covariates. We
evaluate the synthetic backbone on real series and study what real-data
adaptation and covariate conditioning add. Figures~\ref{fig:frozen-backbone-architecture}
and~\ref{fig:joint-patch-adapter-architecture} summarize the backbone and
adapter. Appendix~\ref{sec:related-work} discusses prior work in more detail.

\paragraph{Contributions.}
\textbf{(i) Synthetic data generation.} We release the compositional
generator as open source. We show that our model, trained on this generated data, achieves the highest point-forecasting and probabilistic forecasting accuracy among models trained exclusively on synthetic data, as evaluated on GIFT-Eval and FEV-Bench.
(Fig.~\ref{fig:synthetic-normalized-mase-8k}).
\textbf{(ii) Modular covariate adaptation.} We train a separate adapter to refine the forecast quantiles of a frozen univariate backbone using target history and available covariates. The adapter is backbone-agnostic and can be applied to different forecasting models operating in univariate mode, without any additional training or model-specific adaptation. Remarkably, this shared adapter achieves performance on par with the models' native covariate-handling mechanisms and, in some cases, even surpasses them (Fig.~\ref{fig:fev-adapter-transfer}). On FEV-Bench, the adapted system outperforms Chronos-2 and TiRex-2 on both skill metrics, while enabling the adapter reduces aggregate MASE on tasks with dynamic covariates (Figs.~\ref{fig:fev-native-benchmark} and~\ref{fig:fev-covariate-effect}).
\textbf{(iii) Audited real-data adaptation.} We construct separate
adaptation corpora using the overlap rules for GIFT-Eval and FEV-Bench.
We evaluate each checkpoint only on its designated benchmark and
document source exclusions, shared histories, and the use of benchmark
development windows for model selection (Sec.~\ref{sec:adaptation}).
In comparisons with open-weight zero-shot releases, \textsc{Tafsut} ranks in
the top ten on both metrics on each benchmark
(\hyperref[sec:native-benchmarks]{Adapted Results}).
\textbf{(iv) Systematic ablations.} We examine generator composition,
augmentation, architecture, and training choices
(Sec.~\ref{sec:design-analysis}) that lead to our competitive model.

\section{Problem Formulation}
\label{sec:problem-formulation}

\paragraph{Notation.}
Scalar variables use ordinary math italic, vectors use bold lowercase
symbols, and matrices use bold uppercase symbols. For integers $a\leq b$,
$\vect{y}_{a:b}=(y_a,\ldots,y_b)$ denotes a vector indexed from $a$ to $b$,
including both endpoints. The scalar $X_{tj}$ is the entry in row $t$ and
column $j$ of $\mat{X}$. We write $\reals$ for the real numbers and
$\bar{\reals}=\reals\cup\{\infty\}$ for the space of recorded time-series
values, with $\infty$ marking a missing or unavailable entry.
We use $\mathcal{D}$ for a probability distribution, $\sim$ for a draw
from a distribution, and $\mathbb{E}$ for expectation. The indicator
$\ind\{A\}$ equals $1$ if the condition $A$ holds and $0$ otherwise.
The numeric suffix $\mathrm{K}$ denotes $1024$.

\paragraph{Time Series Forecasting.} We consider probabilistic forecasting of a
real-valued time series indexed relative to the forecast origin. The
historical context consists of $C$ time steps indexed by
$t\in\{-C,\ldots,-1\}$, and the forecast horizon consists of $H$ time
steps indexed by $t\in\{0,\ldots,H-1\}$.
A target window
$\vect{y}_{-C:H-1}\in\bar{\reals}^{C+H}$ consists of the history
$\vect{y}_{-C:-1}$ and the future labels $\vect{y}_{0:H-1}$ used for
training. For a value $v$, availability is $\ind\{v\in\reals\}$.

\paragraph{Forecasting Inputs.} We denote the target history by
$\mathcal{I}_y=\vect{y}_{-C:-1}$. In the covariate-conditioned setting,
we also use $K$ time-aligned covariates. The covariate matrix
$\mat{X}\in\bar{\reals}^{(C+H)\times K}$ has rows indexed by
$t\in\{-C,\ldots,H-1\}$ and columns indexed by $j\in\{1,\ldots,K\}$.
An entry $X_{tj}$ equals $\infty$ if it is unavailable. This includes all future entries of past-only
covariates. Known-future covariates such as calendar features could have
real values extending over the horizon. The forecasting input in this setting is
$\mathcal{I}=(\mathcal{I}_y,\mat{X})\in\bar{\reals}^C\times
\bar{\reals}^{(C+H)\times K}$. In the univariate setting, it is
$\mathcal{I}=\mathcal{I}_y\in\bar{\reals}^C$.

\paragraph{Forecasting Objective.} Let $\mathcal{D}(\,\cdot\,\mid\mathcal{I})$
denote the joint conditional distribution of the real-valued future target
on $\reals^H$. For each forecast time step $t$, let
$\mathcal{D}_t(\,\cdot\,\mid\mathcal{I})$ denote its marginal distribution.
We summarize these marginals at a finite set of quantile levels
$\mathcal{T}\subset(0,1)$, with $Q=|\mathcal{T}|$. For $\tau\in\mathcal{T}$,
the target quantile is
$q_{\tau,t}(\mathcal{I})=
\inf\{z\in\reals:\mathcal{D}_t((-\infty,z]\mid\mathcal{I})\ge\tau\}$.

For a residual $e$ defined as the target minus its predicted quantile,
the pinball loss is $\rho_\tau(e)=\max\{\tau e,(\tau-1)e\}$. The
conditional $\tau$-quantile minimizes the conditional expected pinball
loss. At $\tau=1/2$, $\rho_{1/2}(e)=\tfrac12|e|$, so the conditional
median supplies a point forecast that minimizes conditional expected
absolute error. 

\paragraph{Quantile Forecaster.}
Let $d_\theta$ be the number of trainable parameters and let
$\vect{\theta}\in\reals^{d_\theta}$ collect them. For a specified
horizon $H$, a quantile forecaster $f(\,\cdot\,;\vect{\theta})$ maps valid
inputs to prediction matrices in $\reals^{Q\times H}$. The entry of
$f(\mathcal{I};\vect{\theta})$ indexed by level $\tau$ and time step $t$
estimates $q_{\tau,t}(\mathcal{I})$.
\textsc{Tafsut} expresses these estimates in units normalized using
$\mathcal{I}_y$ and applies the inverse transform at inference
(Sec.~\ref{sec:framework-backbone}).

\paragraph{Training Objective.}
Let $\mathcal{D}_{\mathrm{train}}$ be the joint sampling distribution of
recorded target windows and their covariate matrices. The per-window pinball
loss $\ell$ scores available future labels after normalizing them using
the target history
(Sec.~\ref{sec:framework-backbone}). The optimization problem is
\begin{equation}
 \min_{\vect{\theta}}
 \mathop{\mathbb{E}}_{(\vect{y}_{-C:H-1},\mat{X})\sim\mathcal{D}_{\mathrm{train}}}
 \bigl[\ell\bigl(\vect{y}_{-C:H-1},f(\mathcal{I};\vect{\theta})\bigr)\bigr].
 \label{eq:joint-training-objective}
\end{equation}

\paragraph{Zero-shot Forecasting.} In the zero-shot setting, a pretrained model forecasts an unseen dataset
without additional training or adaptation on that dataset.

\section{Synthetic Data Generation}
\label{sec:synthetic-generator}
\label{sec:framework-synthetic}

Synthetic pretraining for cross-domain forecasting needs trajectories
that vary in their dynamics, observation artifacts, and dependencies.
SarSim0 simulates univariate series from
adapted seasonal autoregressive integrated moving-average (SARIMA)
models~\citep{oreshkin2026simulation}, and KernelSynth draws from
Gaussian processes with composed covariance kernels~\citep{ansari2024chronos}.
Temporal causal models describe dependencies through lagged structural
equations~\citep{runge2023causal}. CauKer combines structural causal models
with Gaussian-process primitives~\citep{xie2026cauker}.
Chronos-2 and TiRex-2 generate multivariate training examples by coupling
univariate source trajectories~\citep{ansari2025chronos2,podest2026tirex2}.
Tabby extends CauKer with a broader set of base processes in CauKerV2
and uses the generated channels as individual univariate training
examples~\citep{xie2026tabby}.

\textsc{Tafsut} composes white-box process models in the time domain and exposes controls over their shape, evolution, and observation. Its final univariate mixture contains 24 components spanning multiscale recurrence, autoregression, time-varying volatility, intermittent arrivals, and persistent regimes. A shared primitive, the atomic function, defines local shapes by interpolating sampled anchors. For jump-free linear interpolation, a bound on the slope between anchors limits how fast the function can change. An atom can define a seasonal cycle, a slot in a recurring calendar, or the decay of a shock. For example, we can repeat an atom to form a daily demand pattern, suppress activity during scheduled closures, and use a bounded stochastic state to vary its amplitude over time. We can then apply different noise levels or censoring thresholds to the same latent process. For multivariate examples, atoms can also define nonlinear maps from covariate values to their contributions to the target. Reusing these mechanisms lets us vary the combination, prevalence, and severity of forecasting conditions across the training mixture and test which choices improve transfer to real series.

\paragraph{Atomic Functions.}
We use a common primitive for seasonal cycles, calendar motifs, and
some impulse response profiles. For a duration $D>0$, an atomic function
$\varphi:[0,D]\to\reals$ interpolates anchors $(u_i,v_i)$ at sampled
positions $0=u_0<\cdots<u_m=D$.
In the continuous construction without jumps, an amplitude bound $A>0$
constrains $|v_i|\leq A$ and a slope bound $L>0$ constrains
$|v_i-v_{i-1}|\leq L(u_i-u_{i-1})$ for $i\geq1$.
Linear interpolation then gives a Lipschitz bound on the unscaled atom,
$|\varphi(u)-\varphi(u')|\leq L|u-u'|$ for $u,u'\in[0,D]$.
Piecewise cubic Hermite interpolation (PCHIP) joins the anchors with
continuous slopes~\citep{fritsch1984interpolation}, and piecewise-constant
interpolation gives plateaus and abrupt transitions. For continuous
interpolants, matching endpoint values $v_0=v_m$ removes the jump at a
repetition boundary. Figure~\ref{fig:generator-atomic} in
Appendix~\ref{app:generator-details} illustrates the anchor bounds,
interpolation, and endpoint bridging.

\paragraph{Time-Varying Controls.}
We generate some trends and time-varying parameters from bounded
stochastic states. We obtain a state by applying
$\tanh$ to a stationary Gaussian autoregression of order one. Its
persistence controls how gradually the state evolves. Scaling or
transforming this state gives time-varying controls for random-walk
drift, the level of mean reversion, innovation scale, and the
amplitude or local clock rate of recurring patterns.

\paragraph{Target Generation.}
For context length $C$, horizon $H$, and trajectory length $T\ge C+H$,
we draw a target trajectory from a mixture of $B$ component distributions
$\mathcal{D}_b^{(r)}(\,\cdot\,\mid C,H,T)$. The setting $r$ specifies
the active mechanisms, their parameter distributions, and mixture weights
$\vect{\pi}^{(r)}$, with $\pi_b^{(r)}\geq 0$ and
$\sum_{b=1}^{B}\pi_b^{(r)}=1$. The resulting distribution is
$\mathcal{D}_{\mathrm{syn}}^{(r)}(\,\cdot\,\mid C,H,T)
=\sum_{b=1}^{B}\pi_b^{(r)}\mathcal{D}_b^{(r)}(\,\cdot\,\mid C,H,T)$.
We extract the context and its immediate continuation from the trajectory.

\paragraph{Trajectory Composition.}
We arrange atomic functions into motifs that recur at sampled periods.
Latent calendars combine schedules at different periods. For example,
a gate $g_t\in[0,1]$ suppresses a recurring signal $s_t$ through
$g_ts_t$, with $g_t=0$ giving latent inactivity.
Seasonal autoregressive moving-average dynamics introduce
serial dependence~\citep{hyndman2021fpp}, conditional variance recursions
produce volatility clustering~\citep{bollerslev1986garch}, and
self-exciting arrivals produce bursts~\citep{hawkes1971selfexciting}.
The generator also includes changepoints, finite-range approximations
to long memory, discrete counts, and persistent regimes. A canonical
bank samples direct laws for processes such as counters, bounded stocks,
and pulse trains. Observation profiles vary the nominal signal-to-noise
ratio and the severity of recording effects such as censoring, quantization, and
saturation. The transforms respect prescribed properties such as
integer support and exact zeros.
Missing observations are encoded as $\infty$.

\paragraph{Covariate Generation.}
We reuse univariate source processes and sample mechanisms for
cross-series dependence separately to generate target-covariate
trajectories. The eight relationship families cover independent controls, shared
latent sources, transformed views, delayed linear effects, nonlinear
structural effects, cointegration, event gating, and amplitude modulation.
In the graph-based families, each channel combines private baseline
dynamics with lagged, filtered responses to its parents. Nonlinear
edge maps can reuse the atomic-function construction. Independent
distractors provide examples with irrelevant inputs. After generating
the complete trajectories, we set entries of the covariate matrix that
are unavailable at the forecast origin to $\infty$.

Appendix~\ref{app:generator-details} gives the component definitions,
sampling rules, composition order, and observation profiles.
Tables~\ref{tab:generator-defaults} and~\ref{tab:adapter-dag-configuration}
list the generator settings. Appendix~\ref{app:representation-analyses}
compares the final mixture with GIFT-Eval-S and FEV-Bench-S windows in
projected \textsc{Tafsut} representations (Figure~\ref{fig:generator-benchmark-coverage}).
Section~\ref{sec:design-analysis} tests changes to the generator
mixture, and Figure~\ref{fig:synthetic-normalized-mase-8k} reports
transfer from synthetic-only training.

\section{Model Architecture}
\label{sec:framework}

\textsc{Tafsut} composes a univariate backbone and a multivariate adapter as
$f(\mathcal{I};\vect{\theta})=
f_{\mathrm{mv}}(\mathcal{I},\,f_{\mathrm{uni}}(\mathcal{I}_y;\vect{\theta}_{\mathrm{uni}});\vect{\theta}_{\mathrm{mv}})$
with $\vect{\theta}=[\vect{\theta}_{\mathrm{uni}};\vect{\theta}_{\mathrm{mv}}]$.
We choose the backbone for its univariate forecasting performance and train
it on target-only windows (Eq.~\eqref{eq:backbone-pinball}). We then freeze it
and train the adapter on joint windows (Eq.~\eqref{eq:adapter-pinball}) to
refine its quantiles using the available covariates. The quantile forecast
is the interface between the two components, so the same adapter architecture
can be paired with different backbones that provide compatible forecasts.
We write $L$ for the number of blocks, $d$ for the model width, and
$d_{\mathrm{ff}}$ for the SwiGLU width. Table~\ref{tab:architecture-config}
gives the default settings for synthetic training.

\begin{table}[!htbp]
\centering
\caption{Training configurations. The adapter parameter count excludes the frozen backbone.}
\label{tab:architecture-config}
\footnotesize
\setlength{\tabcolsep}{4pt}
\renewcommand{\arraystretch}{0.95}
\reporttablestyle
\begin{tabular}{lrr}
\toprule
\reporttablehead{Setting} & \reporttablehead{Backbone} & \reporttablehead{Adapter}\\
\midrule
Blocks / model width / SwiGLU width & 14 / 768 / 2K & 8 / 640 / 1,504\\
Heads $\times$ head width, logit scale $\alpha$ & $12\times64,\ 1$ & $10\times64,\ 1/8$\\
Covariate scale floors $s_{\mathrm{abs}},s_{\mathrm{rel}}$ & \textemdash & $0.1, 0.1$\\
Trainable parameters & 102.7M & 50.1M\\
\midrule
Patch length / stride & \multicolumn{2}{r}{16 / 16}\\
Default quantile levels & \multicolumn{2}{r}{$0.1, 0.5, 0.9$}\\
Time scale $s_t$ / rotary base & \multicolumn{2}{r}{1K / 10,000}\\
Dropout rate / zero-scale replacement $\epsilon_s$ & \multicolumn{2}{r}{$0.1 / 10^{-6}$}\\
\bottomrule
\end{tabular}
\end{table}

\subsection{Univariate Backbone}
\label{sec:framework-backbone}
Following PatchTST~\citep{nie2023patchtst}, we group the target history into
patches. A Transformer encoder processes these patches together with blank
future queries to forecast the full horizon in one pass
(Figure~\ref{fig:frozen-backbone-architecture}).

\paragraph{History-Conditioned Normalization.}
\label{par:backbone-normalization}
We derive $m_t=\ind\{y_t\in\mathbb{R}\}$ from the target history
and compute the mean $\mu$ and population standard deviation $s$ over
positions with $m_t=1$. We replace a zero scale with a fixed
$\epsilon_s>0$ and use $(\mu,s)=(0,1)$ when the entire history is missing.
We transform observed values using
$T(v;\mu,s)=\operatorname{asinh}((v-\mu)/s)$, and map predicted quantiles
back to target units with $T^{-1}(u;\mu,s)=\mu+s\sinh(u)$.

\paragraph{Patch Representation.}
\label{par:backbone-patch-features}
We right-align the history on a non-overlapping grid of length-$P$
patches, using $\infty$ to pad an incomplete leading patch.
Let $\vect{r}_k\in\mathbb R^P$ contain the relative time coordinates
divided by a fixed scale $s_t$. Each historical patch concatenates the
time coordinates, transformed values, and observation indicators as
$\vect{a}_k^{\mathrm{hist}}=[\vect{r}_k;\widetilde{\vect{y}}_k;\vect{m}_k]$.
The indicators $\vect{m}_k$ are computed before zero-filling the missing
entries of $\widetilde{\vect{y}}_k$.
A future patch uses
$\vect{a}_k^{\mathrm{fut}}=[\vect{r}_k;\vect{0}_P;\vect{0}_P]$.
Both vectors lie in $\mathbb R^{3P}$ and share the residual MLP embedding
$E(\vect{a})=\mat{W}_2\operatorname{ReLU}(\mat{W}_1\vect{a}+\vect{b}_1)+\vect{b}_2+\mat{W}_s\vect{a}+\vect{b}_s$.
We insert a learned registration token between the history and future
embeddings.

\paragraph{Encoder and Shared Operators.}
\label{par:backbone-encoder}
Let $\mat{Z}^{\ell}$ stack token states as rows. Each encoder block applies
pre-normalized self-attention and a feed-forward sublayer, each with a
residual connection. The attention update is
$\mat{U}^\ell=\mat{Z}^\ell+\operatorname{Attn}_{\ell}
(\operatorname{RMSNorm}(\mat{Z}^\ell);\mat{M}_{\mathrm{attn}})$.
The feed-forward update is
$\mat{Z}^{\ell+1}=\mat{U}^\ell+\operatorname{FFN}_{\ell}
(\operatorname{RMSNorm}(\mat{U}^\ell))$.
RMSNorm~\citep{zhang2019rmsnorm} divides each vector by its RMS and
applies a learned coordinate-wise gain. We use a
SwiGLU~\citep{shazeer2020glu} feed-forward sublayer,
$\operatorname{FFN}(\vect{u})=\mat{W}_o[\operatorname{SiLU}(\mat{W}_g\vect{u})\odot\mat{W}_v\vect{u}]$.
We apply dropout to embeddings, the feed-forward sublayer, and residual
branches during training.

\phantomsection
\label{par:backbone-attention}
Each head $h$ projects the normalized states to queries $\mat{Q}_{\ell,h}$,
keys $\mat{K}_{\ell,h}$, and values $\mat{V}_{\ell,h}$. Rotary positional
embeddings~\citep{su2024roformer} act on queries and keys. The attention
weights are $\mat{A}_{\ell,h}=\operatorname{softmax}
(\alpha\mat{Q}_{\ell,h}\mat{K}_{\ell,h}^{\mathsf T}+\mat{M}_{\mathrm{attn}})$,
where $\alpha$ is the logit scale and softmax is applied row-wise.
The first layer uses
$\widetilde{\mat{V}}_{1,h}=\mat{V}_{1,h}$. For $\ell>1$, it uses
$\widetilde{\mat{V}}_{\ell,h}
=\lambda_{\ell,1}\mat{V}_{1,h}+\lambda_{\ell,2}\mat{V}_{\ell,h}$.
The value-mixing coefficients are learned per layer and initialized to
$(0,1)$. A sigmoid gate scales each query row of
$\mat{A}_{\ell,h}\widetilde{\mat{V}}_{\ell,h}$ before head concatenation
and output projection.

The attention mask $\mat{M}_{\mathrm{attn}}$ has entries $0$ for allowed
query-key pairs and $-\infty$ otherwise. Attention is bidirectional
across the history, registration token, and blank future queries. A history
patch is excluded as a key when all entries of $\vect{m}_k$ are zero.
Future target labels enter through the training objective.

\paragraph{Quantile Prediction and Training Objective.}
The sampler selects a context length $C$ and a supervised horizon $H$,
then cuts the context and its immediate continuation from a generated
trajectory (Sec.~\ref{sec:synthetic-generator}). Missing observations
carry $\infty$. At a fixed training stage,
$\mathcal{D}_{\mathrm{uni}}$ denotes the resulting distribution of target windows
$\vect{y}_{-C:H-1}\in\bar{\reals}^{C+H}$.
The backbone maps the historical input
$\mathcal{I}_y=\vect{y}_{-C:-1}$ to transformed
quantiles $f_{\mathrm{uni}}(\mathcal{I}_y;\vect{\theta}_{\mathrm{uni}})
\in\reals^{Q\times H}$, where $\vect{\theta}_{\mathrm{uni}}$ is the
vector of trainable backbone parameters. The output columns correspond to
forecast time steps $t\in\{0,\ldots,H-1\}$.
After final RMSNorm, a residual MLP emits $QP$ values per future patch,
which we assemble into this matrix. At inference, $T^{-1}$ maps the
quantiles back to target units. During training, we apply $T$ to finite
labels in the loss. If the query grid extends past $H$, we pad the labels
with $\infty$. Let $n$ be the number of finite labels in the horizon,
clamped below at one, so that
$n=\max\bigl(1,\sum_{t=0}^{H-1}\ind\{y_t\in\reals\}\bigr)$.
For a matrix $\mat{Q}$ of normalized predictions with entries
$\widehat{q}_{\tau,t}$, the per-window loss is
\[
 \ell(\vect{y}_{-C:H-1},\mat{Q})
 =\frac{1}{n}\sum_{\substack{0\le t<H\\y_t\in\reals}}
  \sum_{\tau\in\mathcal{T}}2\rho_\tau\bigl(T(y_t;\mu,s)-\widehat{q}_{\tau,t}\bigr).
\]
The pinball loss $\rho_\tau$ is defined in Sec.~\ref{sec:problem-formulation}.
A window without valid labels contributes zero. 

The backbone minimizes the expected per-window loss
\begin{equation}
 \min_{\vect{\theta}_{\mathrm{uni}}}
 \mathop{\mathbb{E}}_{\vect{y}_{-C:H-1}\sim\mathcal{D}_{\mathrm{uni}}}
 \bigl[\ell\bigl(\vect{y}_{-C:H-1},
 f_{\mathrm{uni}}(\mathcal{I}_y;\vect{\theta}_{\mathrm{uni}})\bigr)\bigr].
 \label{eq:backbone-pinball}
\end{equation}

\subsection{Multivariate Adapter}
\label{sec:multivariate-adapter}
\label{sec:adapter-architecture}
The adapter receives the target history, the frozen backbone quantiles,
and the covariate matrix on the context and forecast grid
(Figure~\ref{fig:joint-patch-adapter-architecture}).

\paragraph{Joint Target Representation.}
Target observations and backbone quantiles share the
\hyperref[par:backbone-normalization]{history-conditioned transform}
$T_y(\,\cdot\,)=T(\,\cdot\,;\mu_y,s_y)$. Let $s_{\mathrm{abs}}$ and
$s_{\mathrm{rel}}$ be the absolute and relative covariate scale floors.
Each covariate $j$ uses its own historical mean and scale
$s_j=\max(\sigma_j,s_{\mathrm{abs}},s_{\mathrm{rel}}r_j)$, where $\sigma_j$
and $r_j$ are the standard deviation and RMS over its observed history.
We derive covariate availability as $\ind\{X_{tj}\in\mathbb{R}\}$.
The historical statistics transform observed entries on the full grid.

Let $\mat{Q}^0$ be the frozen quantile matrix with entries $q^0_{\tau,t}$,
and let $\mat{Q}^0_k$ be its restriction to patch $k$. The target stream
extends the \hyperref[par:backbone-patch-features]{backbone patch features}
with $Q$ forecast channels and their availability indicators.
For a historical patch $k<N_c$, the input is
$\vect{a}^y_k=[\vect{r}_k;\widetilde{\vect{y}}_k;\vect{m}^y_k;\vect{0}_{QP};\vect{0}_{QP}]$.
For a forecast patch $k\geq N_c$, it is
$\vect{a}^y_k=[\vect{r}_k;\vect{0}_P;\vect{0}_P;
\vectorize(\widetilde{\mat{Q}}^0_k);\vect{a}^q_k]$.
Here $N_c$ counts historical patches, $\widetilde{\mat{Q}}^0$ contains
the transformed backbone quantiles, and $\vect{a}^q_k$ indicates which
slots contain backbone forecasts. The target observation indicators
$\vect{m}^y_k$ are derived as in the backbone. Both target patch vectors
lie in $\reals^{(3+2Q)P}$. A shared linear projection embeds history and
forecast patches, followed by a learned target-role embedding.
Each covariate uses an aligned stream of time, zero-filled transformed
values, and derived availability indicators, with separate history and
future projections and a shared covariate-role embedding.

\paragraph{Axial Information Exchange.}
Each adapter block extends the \hyperref[par:backbone-encoder]{backbone block}
with cross-variate attention. Temporal attention gives
$\mat{A}^\ell=\mat{Z}^\ell+\operatorname{Attn}_{\mathrm{time}}
(\operatorname{RMSNorm}(\mat{Z}^\ell);\mat{M}_{\mathrm{time}})$.
Cross-variate attention gives
$\mat{B}^\ell=\mat{A}^\ell+\operatorname{Attn}_{\mathrm{var}}
(\operatorname{RMSNorm}(\mat{A}^\ell);\mat{M}_{\mathrm{var}})$.
The feed-forward update is
$\mat{Z}^{\ell+1}=\mat{B}^\ell+\operatorname{FFN}
(\operatorname{RMSNorm}(\mat{B}^\ell))$.
Temporal attention groups tokens within each stream. Cross-variate
attention groups streams at the same patch position. Both use
the \hyperref[par:backbone-attention]{backbone attention rule} with logit
scale $\alpha$ and value residuals between corresponding sublayers.
Temporal attention also uses rotary positions.

Attention is bidirectional over the input grid. The attention masks isolate
independent packed tasks and exclude absent streams and padding, while
retaining aligned slots for unavailable future values in existing
streams. For $S$ streams and $N$ patches, this requires
$\mathcal O(SN^2+NS^2)$ attention interactions per block.

\paragraph{Quantile Prediction and Training Objective.}
For a normalized future target state $\vect{h}_k$, the readout uses
$\vect{b}_k=\vectorize(\widetilde{\mat{Q}}^0_k)$ to predict the correction
$\vect{\delta}_k=R_{\vect{\phi}}(\vect{h}_k+\mat{W}_b\vect{b}_k)$.
Here $\mat{W}_b$ projects the prior and $R_{\vect{\phi}}$ is a
LayerNorm-linear-GELU-linear head. The corrected quantiles are
$\widetilde{\mat{Q}}_k=\widetilde{\mat{Q}}^0_k+
\operatorname{reshape}_{Q\times P}(\vect{\delta}_k)$.
The multivariate generator draws joint target and covariate trajectories.
The sampler cuts aligned windows and sets covariate entries unavailable
at the forecast origin to $\infty$
(Sec.~\ref{sec:synthetic-generator}). Let $\mathcal{D}_{\mathrm{mv}}$
denote the resulting distribution of joint windows $(\vect{y}_{-C:H-1},\mat{X})$.
The forecasting input is $\mathcal{I}=(\mathcal{I}_y,\mat{X})$, and
the future target slice supplies the labels as in the univariate objective.
The frozen forecast is $\widetilde{\mat{Q}}^0=
f_{\mathrm{uni}}(\mathcal{I}_y;\widehat{\vect{\theta}}_{\mathrm{uni}})$,
where $\widehat{\vect{\theta}}_{\mathrm{uni}}$ denotes the fitted backbone
parameters held fixed during adapter training. With trainable adapter
parameters $\vect{\theta}_{\mathrm{mv}}$, the assembled $Q\times H$
matrix of corrected quantiles is
$\widetilde{\mat{Q}}=f_{\mathrm{mv}}(\mathcal{I},\widetilde{\mat{Q}}^0;\vect{\theta}_{\mathrm{mv}})$.\footnote{Let $\widetilde{q}_t^0$ and $\widetilde{q}_t$ denote the
quantile functions of the target-only marginal
$\mathcal{D}_t(\cdot\mid\mathcal{I}_y)$ and the covariate-conditioned
marginal $\mathcal{D}_t(\cdot\mid\mathcal{I})$, respectively, in
normalized units. For a shared $U\sim\operatorname{Uniform}(0,1)$,
the pair $(\widetilde{q}_t^0(U),\widetilde{q}_t(U))$ induces the
one-dimensional monotone transport coupling between these marginals
\citep[Sec.~2.6]{peyre2019computational}. The adapter's additive
correction thus estimates the corresponding quantile displacement
$\widetilde{q}_t(\tau)-\widetilde{q}_t^0(\tau)$ at each modeled
level $\tau\in\mathcal{T}$, using the pinball objective in
Eq.~\eqref{eq:adapter-pinball} evaluated on valid target observations.}
At inference, we sort the corrected quantiles and map them back to target
units. The adapter minimizes the same expected per-window loss
\begin{equation}
 \min_{\vect{\theta}_{\mathrm{mv}}}
 \mathop{\mathbb{E}}_{(\vect{y}_{-C:H-1},\mat{X})\sim\mathcal{D}_{\mathrm{mv}}}
 \bigl[\ell\bigl(\vect{y}_{-C:H-1},
 f_{\mathrm{mv}}(\mathcal{I},\widetilde{\mat{Q}}^0;\vect{\theta}_{\mathrm{mv}})
 \bigr)\bigr].
 \label{eq:adapter-pinball}
\end{equation}


\section{Domain Adaptation}
\label{sec:adaptation}
\label{sec:real-posttraining}
\label{sec:adaptation-data}
\paragraph{Datasets.} We adapt the final synthetic backbone separately for GIFT-Eval and
FEV-Bench. Both real-data corpora draw on Salesforce's GIFT-Eval-Pretrain
and Amazon's public Chronos datasets. GIFT-Eval-Pretrain comprises 88
datasets, 4.5 million series, and 230 billion data points~\citep{aksu2024gift}.
The original Chronos real-data training subset comprises 28 datasets,
about 890,000 univariate series, and 84 billion observations~\citep{ansari2024chronos}.
The GIFT-Eval run uses 178 selected source configurations. The FEV-Bench
run draws from a filtered inventory of 2,980 entries, including 2,792 BOOM members
(Appendix~\ref{app:adaptation-corpora}).

The GIFT-Eval archive covers 28 dataset roots and 97 forecasting tasks,
with 159,376 univariate series and 40.2 million finite target observations.
The FEV-Bench archive covers 56 dataset families and 100 tasks, including
tasks with dynamic covariates~\citep{shchur2025fev}, with 125,344 univariate
series and 8.93 million finite target observations. Series are counted
within each task and observations across forecast windows
(Appendix~\ref{sec:benchmarks}).

Because each adaptation corpus overlaps with the other
benchmark, we evaluate each checkpoint only on its designated benchmark
to preserve the zero-shot protocol. Both runs mix real and synthetic
windows and optimize the backbone objective in
Eq.~\eqref{eq:backbone-pinball}.

\paragraph{Overlap Checks.}
The GIFT-Eval corpus excludes the official benchmark training split and
the Alabama solar subset.  The verified shared intervals
fall outside the forecast targets (Table~\ref{tab:adaptation-source-selection}). For FEV-Bench, we treat exposure to the same dataset at the same frequency
as overlap, including truncated series. Nonetheless, the benchmark protocol permits
resampled variants at other frequencies~\citep{shchur2025fev}.
Table~\ref{tab:fev-adaptation-source-selection} records exclusions from our pretraining corpus.

We separately audit prepared in-domain histories for cross-task leakage
across forecast terms and frequencies (Appendix~\ref{app:benchmark-cross-task-audit}).
GIFT-Eval yields 38 exact same-frequency conflicts and 98 cross-frequency
candidates. In  FEV-Bench contexts, we find zero exact
same-frequency conflicts and 48 cross-frequency leaked peers.

\paragraph{Backbone Adaptation.}
For GIFT-Eval, we train on  200 additional epochs from
the synthetic checkpoint. And for FEV-Bench, we train on the filtered corpus in
Appendix~\ref{app:adaptation-corpora}.  We train for additional 100 epochs from the synthetic checkpoint.

\paragraph{Covariate Adaptation.}
For FEV-Bench tasks with dynamic covariates, we use the three-quantile
adapter (Sec.~\ref{sec:multivariate-adapter}) trained on the FEV-Bench
source corpus with synthetic relationship drivers. Tasks without dynamic
covariates use the adapted backbone alone.

\phantomsection
\label{sec:extended-quantile-heads}
\paragraph{Extended Quantile Heads.}
The full-benchmark probabilistic metrics use nine quantile levels
$\{0.1,0.2,\ldots,0.9\}$. Our extended quantile heads use forecast features
and quantile predictions from frozen models to produce forecasts on this
grid. The backbone head learns the surrounding quantiles with the original
median as an anchor. For the adapted FEV-Bench covariate route, the head
retains the adapter's $0.1$, $0.5$, and $0.9$ forecasts and learns the
relative spacings of the six intermediate quantiles within those intervals.
We train this head for 3,000 updates with an earlier backbone extended
quantile head and use the final frozen backbone head at evaluation.

\section{Benchmark Evaluation}
\label{sec:experimental-setup}

\phantomsection
\label{sec:data-and-protocols}
\paragraph{Benchmarks.}
We evaluate on GIFT-Eval~\citep{aksu2024gift} and FEV-Bench~\citep{shchur2025fev}.
Appendix~\ref{sec:benchmarks} summarizes the evaluation populations and
lists the benchmark tasks.

\paragraph{Evaluation Protocols.}
We compare the final synthetic \textsc{Tafsut} backbone
(Appendix~\ref{app:training-lineage}) with the released
Chronos-2 Synth checkpoint~\citep{ansari2025chronos2,chronos2synth} on all
archived windows at their native horizons. Both use target-only histories
capped at 8K steps. We also evaluate the same \textsc{Tafsut} checkpoint at
32K steps with identical targets, scales, and eligibility. On GIFT-Eval, we use the published TabPFN-TS~\citep{hoo2026tabpfnts}
scores and evaluate the released TempoPFN checkpoint~\citep{moroshan2026tempopfn}
on the full archive. On FEV-Bench, we use the published TabPFN-TS run with
known-future covariates and fill its missing task with Seasonal Naive. We
convert TempoPFN's published FEV-Bench skill scores to error ratios.

For the adapted models (Sec.~\ref{sec:adaptation}), we use the extended
quantile heads (Sec.~\ref{sec:extended-quantile-heads}) for comparison
with public benchmark scores. Both backbones use up to 32K history steps.
On FEV-Bench, the adapter uses
up to 8K steps on the 42 tasks with dynamic covariates.

\phantomsection
\label{sec:development-sets}
\paragraph{Development and Synthetic Sets.}
For each benchmark, we allocate a budget of 32K archived forecast windows
as evenly as possible across tasks. If a task has fewer windows than its
allocation, we redistribute
the unused slots among tasks with windows remaining. We sample uniformly
without replacement within each task and reuse the same selected windows
across evaluations. We denote the resulting subsets by \textbf{FEV-Bench-S}
and \textbf{GIFT-Eval-S}. Ablation scores use only the eligible windows
(Appendix~\ref{sec:ablations}). For architecture selection, we use a held-out
\textbf{synthetic validation bank} of 10K clean series drawn from the
nonseasonal-dynamics preset (Appendix~\ref{sec:preset-search}). The
\textbf{final-mixture bank} contains 16K fresh series from the final
counts-and-artifacts preset, including its 5\% canonical share, and
supports comparisons across training stages.

\phantomsection
\label{sec:evaluation-metrics}
\paragraph{Metrics.}
For window $i$ in task $d$ with seasonal period $m_d$, the scale $s_i$ is
the mean of $|y_{i,t}-y_{i,t-m_d}|$ over finite historical pairs. We score
finite targets in windows with finite positive scale and retain missing
positions in the time index. Using forecasts in target units, we compute
mean absolute scaled error (MASE) as the median's mean absolute error
divided by $s_i$, and scaled quantile loss (SQL) as twice the mean pinball
loss (Sec.~\ref{sec:problem-formulation}) over scored quantiles and finite
targets divided by $s_i$. We average window scores within tasks and take
an equal-weight geometric mean across tasks. On FEV-Bench-S and GIFT-Eval-S,
we aggregate the scaled task scores directly, without dividing by their
Seasonal Naive counterparts. These development scores use quantiles
$\{0.1,0.5,0.9\}$ and typically compute the scale on the visible history.
Paired context-cap evaluations use fixed reference scales
(Appendix~\ref{sec:context-caps}). The full-benchmark results use a
different evaluation population and normalize task scores against Seasonal
Naive. Full-benchmark synthetic-only comparisons compute the scale from
uncropped histories.
On GIFT-Eval, we report MASE and the decile approximation to the
continuous ranked probability score (CRPS)~\citep{aksu2024gift}, dividing
each task score by its Seasonal Naive counterpart before aggregation.
Task CRPS sums doubled pinball losses over targets, averages across
deciles, and divides by the total absolute target magnitude. In the
synthetic-only comparison on GIFT-Eval, we interpolate deciles from the
three training quantiles with GluonTS~\citep{alexandrov2020gluonts}.
On FEV-Bench, we report official MASE and nine-decile SQL as
ratios to Seasonal Naive for synthetic-only models and as skill scores
for adapted models~\citep{shchur2025fev}. \textsc{Tafsut}-Synth's FEV-Bench SQL
uses an extended quantile head trained on synthetic data.
Skill is $100(1-g)$, where $g$ is the geometric mean of task error ratios
to Seasonal Naive after clipping each ratio to $[0.01,100]$. The pairwise
win rate is the fraction of tasks with lower error, with half credit for ties
(Appendix~\ref{app:fev-pairwise}).

\begin{figure}[!htbp]
\centering
\includegraphics[width=\linewidth]{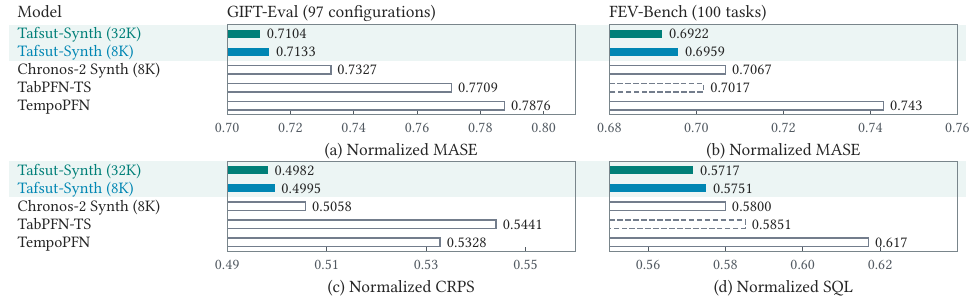}
\caption{Target-only forecasting on full GIFT-Eval and FEV-Bench.
Panels (a,b) show normalized MASE, (c) normalized CRPS, and (d) normalized
SQL. Scores are geometric means of task error ratios to Seasonal Naive.
Lower scores are better, and axes are cropped. Chronos-2 Synth uses
native task horizons with cross-learning disabled. The dashed
TabPFN-TS outlines on FEV-Bench show its
\href{https://github.com/autogluon/fev/blob/main/benchmarks/fev_bench/results/tabpfn-ts.csv}{published run with known-future covariates},
with Seasonal Naive filling one missing task. TempoPFN's FEV-Bench ratios
come from its \href{https://arxiv.org/abs/2510.25502}{published skill scores}.}
\label{fig:synthetic-normalized-mase-8k}
\end{figure}

\phantomsection
\label{sec:main-results}
\paragraph{Synthetic-Only Results.}
Figure~\ref{fig:synthetic-normalized-mase-8k} compares the
\textsc{Tafsut}-Synth backbone with other models trained on synthetic
data. At a shared 8K history cap, \textsc{Tafsut} reduces normalized MASE relative
to Chronos-2 Synth by 2.65\% on GIFT-Eval and 1.53\% on FEV-Bench.
Increasing the \textsc{Tafsut} context cap to 32K improves both point and
probabilistic scores on both benchmarks.

\begin{figure}[!htbp]
\centering
\captionsetup{skip=4pt}
\includegraphics[width=1\linewidth]{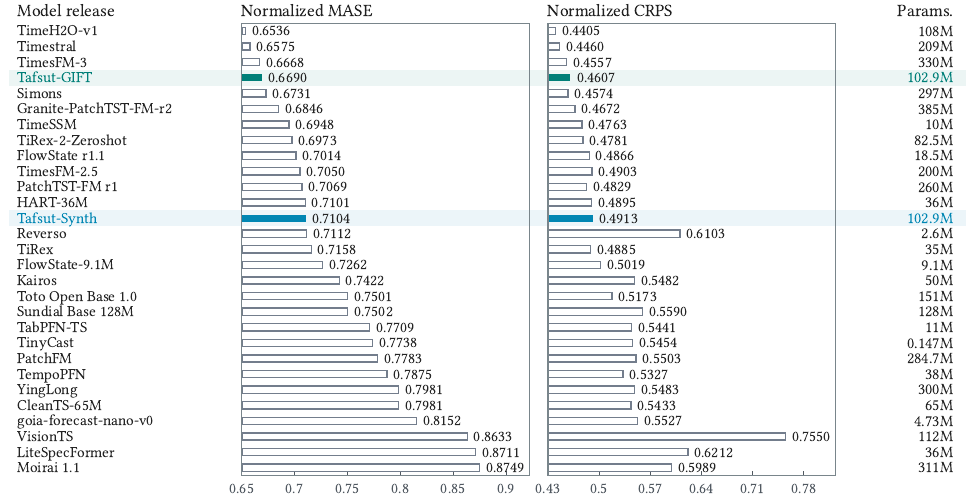}
\caption{GIFT-Eval comparison of 28 open-weight zero-shot releases,
including \textsc{Tafsut}-GIFT, across 97 tasks. The additional \textsc{Tafsut}-Synth row
uses its extended quantile head. Models are ordered by MASE.
Scores are geometric means of task errors relative to Seasonal Naive.
Lower is better. The axes start at 0.65 for MASE and 0.43 for CRPS.}
\label{fig:gift-native-benchmark}
\end{figure}

\phantomsection
\label{sec:native-benchmarks}
\paragraph{Adapted Results.}
On GIFT-Eval, \textsc{Tafsut} achieves normalized MASE of 0.6690 and normalized
CRPS of 0.4607 across 97 tasks and 371,330 windows
(Figure~\ref{fig:gift-native-benchmark}). Relative to interpolation on
the same backbone, the extended quantile head reduces CRPS by 1.02\% and
improves 72 of 97 tasks. Both methods share the median and agree on MASE
to six decimal places.

\begin{figure}[!htbp]
\centering
\captionsetup{skip=4pt}
\includegraphics[width=1\linewidth]{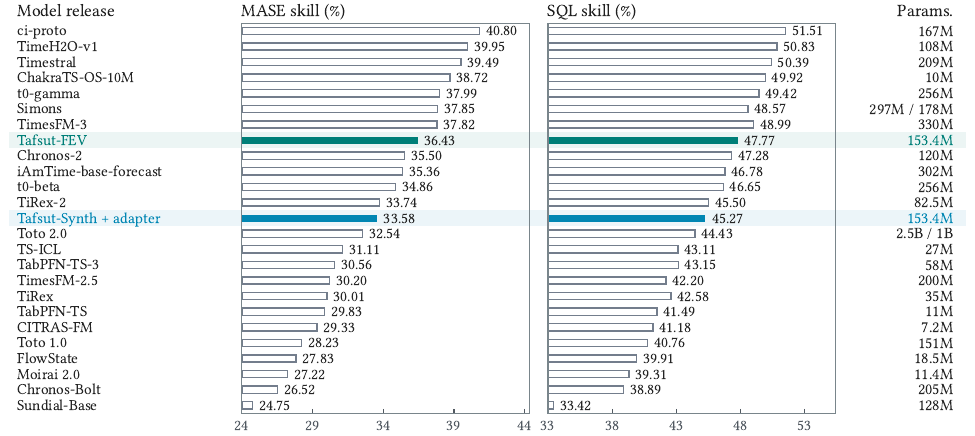}
\caption{FEV-Bench comparison of 24 open-weight zero-shot releases
across 100 tasks, ordered by MASE skill. The additional \textsc{Tafsut}-Synth row
uses its synthetic covariate adapter and extended quantile heads.
Higher is better.
The axes start at 24\% for MASE skill and 33\% for SQL skill.
We select each release's best-scoring checkpoint separately for MASE
and SQL. When these checkpoints differ in size, the two parameter counts
appear in MASE/SQL order.
Both \textsc{Tafsut} rows show 153.4M parameters, the rounded mean of their totals
with the adapters.}
\label{fig:fev-native-benchmark}
\end{figure}

On FEV-Bench, \textsc{Tafsut} achieves 36.43\% MASE skill and 47.77\% SQL skill over
100 tasks and 235,039 windows (Figure~\ref{fig:fev-native-benchmark}).
Relative to interpolation with the same  backbone, geometric
task SQL falls from 0.8226 to 0.8158 and geometric MASE rises from 1.0013
to 1.0019. SQL improves on 84 tasks, including 30 of the 42 adapter tasks.
This comparison reflects changes from both the extended quantile heads
and the covariate route.

\paragraph{Open-Weight Zero-Shot Rankings.}
Figures~\ref{fig:gift-native-benchmark} and~\ref{fig:fev-native-benchmark}
show the open-weight zero-shot subset of the published results. We
exclude API-only entries and entries without released weights. Each
release contributes its best aggregate score per metric across eligible
variants (Appendix~\ref{sec:benchmarks}). Among 28 releases on GIFT-Eval,
\textsc{Tafsut} ranks fourth by normalized MASE and fifth by normalized CRPS.
The published GIFT-Eval \textsc{Tafsut} submission uses a separate checkpoint.
Among 24 releases on FEV-Bench, \textsc{Tafsut} ranks eighth on both MASE and
SQL skill.

\begin{figure}[!htbp]
\centering
\captionsetup{skip=3pt}
\includegraphics[width=\linewidth]{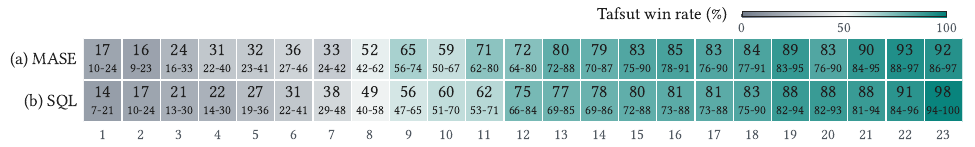}
\caption{\textsc{Tafsut}'s pairwise win rates by MASE (a) and SQL (b) against the
23 open-weight foundation baselines on 100 FEV-Bench tasks. Each cell shows
the win rate above its 95\% task-bootstrap interval. Ties count as half a win.
{In increasing SQL win-rate order, the columns are:
TimeH2O-v1~(1), ci-proto~(2), Timestral~(3),
ChakraTS-OS-10M~(4), t0-gamma~(5), TimesFM-3~(6),
Simons~(7), iAmTime-base-forecast~(8), Chronos-2~(9),
Toto~2.0~(10), t0-beta~(11), TiRex-2~(12),
TimesFM-2.5~(13), TabPFN-TS-3~(14), TS-ICL~(15),
TiRex~(16), Toto~1.0~(17), TabPFN-TS~(18),
CITRAS-FM~(19), FlowState~(20), Moirai~2.0~(21),
Chronos-Bolt~(22), and Sundial-Base~(23).}}
\label{fig:fev-tafsut-win-rates}
\end{figure}

\paragraph{Pairwise Win Rates.}
Figure~\ref{fig:fev-tafsut-win-rates} compares \textsc{Tafsut} with the 23 other
open-weight foundation models in Figure~\ref{fig:fev-native-benchmark}.
Its win rate exceeds 50\% against 16 baselines by MASE and 15 by SQL.
Its MASE and SQL win rates are 65\% and 56\% against Chronos-2 and
36\% and 31\% against TimesFM-3. Appendix~\ref{app:fev-pairwise} gives
the full pairwise matrices, including task-specific and statistical baselines.

\begin{figure}[!htbp]
\centering
\captionsetup{skip=3pt}
\includegraphics[width=.9\linewidth]{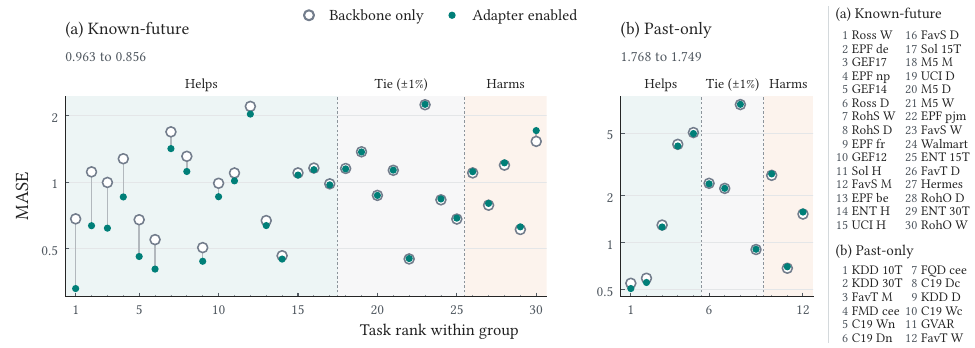}
\caption{FEV-Bench MASE for the FEV univariate checkpoint alone and with its
frozen adapter on matched rolling windows. Panels rank tasks by decreasing
relative MASE reduction. The side key abbreviates task names, and
Table~\ref{tab:fev-covariate-ranks} gives exact benchmark IDs. The suffixes
\texttt{n} and \texttt{c} denote new and cumulative COVID-19 cases. Values above
each panel are geometric means. Bands mark improvements above 1\%, ties
within $\pm 1\%$, and degradations above 1\% relative to backbone-only
MASE. Each panel uses a log axis fitted to its observed range.}
\label{fig:fev-covariate-effect}
\end{figure}

\paragraph{Covariate Effect.}
On the 42 FEV-Bench tasks with dynamic covariates, enabling the
covariate adapter on the univariate backbone lowers geometric mean MASE
by 11.2\% for tasks with known-future covariates and by 1.1\% for tasks
with past covariates only (Figure~\ref{fig:fev-covariate-effect}).
Appendix~\ref{app:fev-covariate-representations} relates these task-level
MASE changes to shifts in the adapter's representations.

\begin{figure}[!htbp]
\centering
\captionsetup{skip=4pt}
\includegraphics[width=\linewidth]{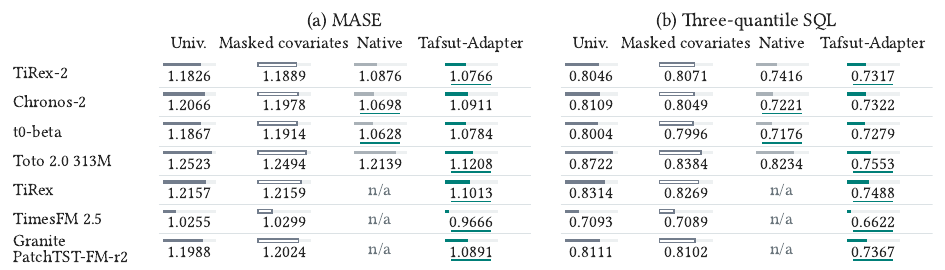}
\caption{Transfer of the frozen \textsc{Tafsut} covariate adapter across
foundation models on 42 FEV-Bench tasks with dynamic covariates.
Masked covariates and \textsc{Tafsut}-Adapter denote the same frozen
adapter with covariates masked and available, respectively. Both runs use a
1K-step univariate forecast, except for TiRex-2, which supplies a
task-horizon forecast with last-value padding. Chronos-2 and t0-beta use
64 adapter queries, and the other models use $\lceil h/16\rceil$ for task
horizon $h$. The univariate reference uses the task-horizon forecast for
TiRex-2, Chronos-2, and t0-beta and the same 1K-step forecast as the adapter
for the other backbones. Scores are equal-task geometric means over the
first $h$ steps. SQL uses quantiles $0.1$, $0.5$, and $0.9$.
Underlines mark the lowest error for each backbone and metric.
Mini-bars use scales of 0.94--1.28 for MASE and 0.64--0.895 for SQL.}
\label{fig:fev-adapter-transfer}
\end{figure}

\paragraph{Covariate Adapter Transfer.}
We test transfer on the 42 FEV-Bench tasks with dynamic covariates by applying
the frozen \textsc{Tafsut}-FEV adapter to TiRex-2, Chronos-2, t0-beta,
Toto~2.0 313M, TiRex, TimesFM~2.5, and Granite PatchTST-FM-r2. The adapter
was trained only with forecasts from the \textsc{Tafsut}-FEV backbone.
It receives each model's univariate forecast quantiles, target history,
and available covariates and uses the same weights across all seven
models. To isolate the effect of covariate access at a fixed parameter
count, we compare paired adapter runs with the same forecast inputs
and the covariates available or masked. Providing covariates reduces
MASE by 6.1--10.3\% and SQL by 6.6--9.9\% across the seven backbones
(Figure~\ref{fig:fev-adapter-transfer}). With covariates masked, MASE
stays within 0.8\% of the univariate baseline for every backbone. The
covariate-enabled adapter also improves both metrics over native
conditioning for TiRex-2 and Toto~2.0 313M, while Chronos-2 and t0-beta
score lowest with native covariates.

\begin{figure}[p]
\centering
\setlength{\parskip}{0pt}
\captionsetup{skip=3pt}
\captionsetup[subfigure]{font=footnotesize,skip=2pt}
\newcommand{\fevexamplespdf}{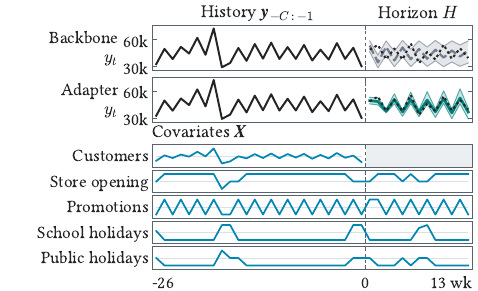}
\newcommand{\fevexamplepanel}[2]{%
  \begin{subfigure}[t]{0.49\linewidth}
    \centering
    \vspace{0pt}%
    \includegraphics[page=#1,width=\linewidth]{\fevexamplespdf}
    \caption{#2}
  \end{subfigure}}

\includegraphics[page=5,width=0.7\linewidth]{\fevexamplespdf}\par
\begin{minipage}{\linewidth}
\captionsetup{type=figure}
\centering
\fevexamplepanel{1}{{Rossmann (weekly)}:
Store sales}\hfill
\fevexamplepanel{2}{{German electricity prices (hourly)}:
Day-ahead price}
\caption{FEV-Bench windows with known-future covariates. Each pair shows
$C=2H$ history steps, median lines, and bands spanning the 0.1 and 0.9
quantiles on a common target scale. The dashed line marks the forecast
origin. Shading marks unavailable future covariates.}
\label{fig:fev-adapter-examples-known-future}
\end{minipage}
\par\smallskip
\begin{minipage}{\linewidth}
\captionsetup{type=figure}
\centering
\fevexamplepanel{3}{{Favorita (monthly)}:
Store transactions}\hfill
\fevexamplepanel{4}{{GVAR (quarterly)}: Log real equity prices in Argentina}
\caption{FEV-Bench windows with past-only covariates. The shaded regions
span the forecast horizon. Figure~\ref{fig:fev-adapter-examples-known-future}
defines the forecast bands.}
\label{fig:fev-adapter-examples-past-only}
\end{minipage}
\end{figure}

\paragraph{Qualitative Examples from Domain Case Studies.}
Figures~\ref{fig:fev-adapter-examples-known-future}
and~\ref{fig:fev-adapter-examples-past-only} show how covariates affect
forecast quantiles for retail activity, electricity prices, and
macroeconomic series. We use the \textsc{Tafsut}-FEV backbone and adapter
from the covariate-effect comparison. We rank tasks by relative MASE reduction
within each covariate setting and show two distinct datasets per group.
Each past-only task has fewer than four covariates. The adapter applies
a recurring correction to all three displayed quantiles for weekly
Rossmann sales. The selected GVAR window has lower median forecast error
with the adapter. Across the GVAR task, the adapter raises aggregate MASE.

\section{Ablations and Design Choices}
\label{sec:design-analysis}
We compare each intervention with its matched control using shorter
training and forecast windows than the final system.
The tables report means and sample standard deviations across the final
five evaluated checkpoints of each 400-epoch run.
Appendix~\ref{sec:ablations} details the evaluation protocol and
reports the remaining ablations.

\begin{figure}[!htbp]
\centering
\captionsetup{skip=3pt}
\captionsetup[subfigure]{font=footnotesize,skip=3pt,justification=centering}
\begin{subfigure}[b]{0.535\linewidth}
\centering
\begingroup\scriptsize
\setlength{\tabcolsep}{1.25pt}
\reporttablestyle[3]
\resizebox{\linewidth}{!}{%
\begin{tabular}{@{}lrrrrrr@{}}
\toprule
& \multicolumn{2}{c}{\reporttablehead{FEV-Bench-S}} &
\multicolumn{2}{c}{\reporttablehead{GIFT-Eval-S}} &
\multicolumn{2}{c}{\reporttablehead{Synthetic bank}} \\
\cmidrule(lr){2-3}\cmidrule(lr){4-5}\cmidrule(l){6-7}
\reporttablehead{Preset} &
\reporttablehead{SQL $\downarrow$} & \reporttablehead{MASE $\downarrow$} &
\reporttablehead{SQL $\downarrow$} & \reporttablehead{MASE $\downarrow$} &
\reporttablehead{SQL $\downarrow$} & \reporttablehead{MASE $\downarrow$} \\
\midrule
P01 & \shortstack[r]{$0.833\!\pm\!.004$\\[-1pt]{\textcolor{ablationworse}{$+0.2\%$}}} & \shortstack[r]{$1.211\!\pm\!.005$\\[-1pt]{\textcolor{ablationworse}{$+0.2\%$}}} & \shortstack[r]{$0.719\!\pm\!.002$\\[-1pt]{\textcolor{ablationworse}{$+0.7\%$}}} & \shortstack[r]{$1.052\!\pm\!.002$\\[-1pt]{\textcolor{ablationworse}{$+0.6\%$}}} & \shortstack[r]{$1.193\!\pm\!.006$\\[-1pt]{\textcolor{ablationbetter}{$-0.5\%$}}} & \shortstack[r]{$\mathbf{1.780}\!\pm\!.010$\\[-1pt]{\textcolor{ablationbetter}{$-0.6\%$}}} \\
P02 & \shortstack[r]{$0.831\!\pm\!.004$\\[-1pt]{\textcolor{ablationneutral}{$0.0\%$}}} & \shortstack[r]{$1.208\!\pm\!.006$\\[-1pt]{\textcolor{ablationneutral}{$0.0\%$}}} & \shortstack[r]{$\mathbf{0.714}\!\pm\!.001$\\[-1pt]{\textcolor{ablationneutral}{$0.0\%$}}} & \shortstack[r]{$1.046\!\pm\!.002$\\[-1pt]{\textcolor{ablationneutral}{$0.0\%$}}} & \shortstack[r]{$1.199\!\pm\!.006$\\[-1pt]{\textcolor{ablationneutral}{$0.0\%$}}} & \shortstack[r]{$1.791\!\pm\!.012$\\[-1pt]{\textcolor{ablationneutral}{$0.0\%$}}} \\
P03 & \shortstack[r]{$\mathbf{0.826}\!\pm\!.002$\\[-1pt]{\textcolor{ablationbetter}{$-0.6\%$}}} & \shortstack[r]{$1.204\!\pm\!.004$\\[-1pt]{\textcolor{ablationbetter}{$-0.3\%$}}} & \shortstack[r]{$0.715\!\pm\!.002$\\[-1pt]{\textcolor{ablationworse}{$+0.1\%$}}} & \shortstack[r]{$\mathbf{1.045}\!\pm\!.002$\\[-1pt]{\textcolor{ablationbetter}{$-0.1\%$}}} & \shortstack[r]{$1.198\!\pm\!.006$\\[-1pt]{\textcolor{ablationbetter}{$-0.1\%$}}} & \shortstack[r]{$1.792\!\pm\!.012$\\[-1pt]{\textcolor{ablationworse}{$+0.1\%$}}} \\
P04 & \shortstack[r]{$0.830\!\pm\!.004$\\[-1pt]{\textcolor{ablationbetter}{$-0.2\%$}}} & \shortstack[r]{$1.207\!\pm\!.007$\\[-1pt]{\textcolor{ablationbetter}{$-0.1\%$}}} & \shortstack[r]{$0.719\!\pm\!.002$\\[-1pt]{\textcolor{ablationworse}{$+0.6\%$}}} & \shortstack[r]{$1.048\!\pm\!.002$\\[-1pt]{\textcolor{ablationworse}{$+0.2\%$}}} & \shortstack[r]{$1.198\!\pm\!.009$\\[-1pt]{\textcolor{ablationbetter}{$-0.1\%$}}} & \shortstack[r]{$1.786\!\pm\!.016$\\[-1pt]{\textcolor{ablationbetter}{$-0.3\%$}}} \\
P05 & \shortstack[r]{$0.834\!\pm\!.003$\\[-1pt]{\textcolor{ablationworse}{$+0.3\%$}}} & \shortstack[r]{$1.212\!\pm\!.004$\\[-1pt]{\textcolor{ablationworse}{$+0.3\%$}}} & \shortstack[r]{$0.724\!\pm\!.002$\\[-1pt]{\textcolor{ablationworse}{$+1.4\%$}}} & \shortstack[r]{$1.060\!\pm\!.003$\\[-1pt]{\textcolor{ablationworse}{$+1.3\%$}}} & \shortstack[r]{$1.201\!\pm\!.003$\\[-1pt]{\textcolor{ablationworse}{$+0.2\%$}}} & \shortstack[r]{$1.791\!\pm\!.004$\\[-1pt]{\textcolor{ablationneutral}{$0.0\%$}}} \\
P06 & \shortstack[r]{$0.836\!\pm\!.013$\\[-1pt]{\textcolor{ablationworse}{$+0.6\%$}}} & \shortstack[r]{$1.208\!\pm\!.005$\\[-1pt]{\textcolor{ablationneutral}{$0.0\%$}}} & \shortstack[r]{$0.719\!\pm\!.003$\\[-1pt]{\textcolor{ablationworse}{$+0.6\%$}}} & \shortstack[r]{$1.051\!\pm\!.005$\\[-1pt]{\textcolor{ablationworse}{$+0.5\%$}}} & \shortstack[r]{$\mathbf{1.190}\!\pm\!.005$\\[-1pt]{\textcolor{ablationbetter}{$-0.8\%$}}} & \shortstack[r]{$1.781\!\pm\!.009$\\[-1pt]{\textcolor{ablationbetter}{$-0.6\%$}}} \\
P07 & \shortstack[r]{$0.829\!\pm\!.004$\\[-1pt]{\textcolor{ablationbetter}{$-0.2\%$}}} & \shortstack[r]{$1.206\!\pm\!.006$\\[-1pt]{\textcolor{ablationbetter}{$-0.2\%$}}} & \shortstack[r]{$0.720\!\pm\!.003$\\[-1pt]{\textcolor{ablationworse}{$+0.9\%$}}} & \shortstack[r]{$1.056\!\pm\!.004$\\[-1pt]{\textcolor{ablationworse}{$+1.0\%$}}} & \shortstack[r]{$1.198\!\pm\!.005$\\[-1pt]{\textcolor{ablationbetter}{$-0.1\%$}}} & \shortstack[r]{$1.790\!\pm\!.009$\\[-1pt]{\textcolor{ablationneutral}{$0.0\%$}}} \\
P08 & \shortstack[r]{$0.828\!\pm\!.002$\\[-1pt]{\textcolor{ablationbetter}{$-0.4\%$}}} & \shortstack[r]{$1.207\!\pm\!.003$\\[-1pt]{\textcolor{ablationbetter}{$-0.1\%$}}} & \shortstack[r]{$0.721\!\pm\!.002$\\[-1pt]{\textcolor{ablationworse}{$+0.9\%$}}} & \shortstack[r]{$1.057\!\pm\!.002$\\[-1pt]{\textcolor{ablationworse}{$+1.1\%$}}} & \shortstack[r]{$1.207\!\pm\!.008$\\[-1pt]{\textcolor{ablationworse}{$+0.7\%$}}} & \shortstack[r]{$1.800\!\pm\!.014$\\[-1pt]{\textcolor{ablationworse}{$+0.5\%$}}} \\
P09 & \shortstack[r]{$0.827\!\pm\!.004$\\[-1pt]{\textcolor{ablationbetter}{$-0.6\%$}}} & \shortstack[r]{$\mathbf{1.202}\!\pm\!.006$\\[-1pt]{\textcolor{ablationbetter}{$-0.5\%$}}} & \shortstack[r]{$0.717\!\pm\!.003$\\[-1pt]{\textcolor{ablationworse}{$+0.4\%$}}} & \shortstack[r]{$1.048\!\pm\!.004$\\[-1pt]{\textcolor{ablationworse}{$+0.2\%$}}} & \shortstack[r]{$1.202\!\pm\!.007$\\[-1pt]{\textcolor{ablationworse}{$+0.2\%$}}} & \shortstack[r]{$1.794\!\pm\!.009$\\[-1pt]{\textcolor{ablationworse}{$+0.2\%$}}} \\
P10 & \shortstack[r]{$0.835\!\pm\!.005$\\[-1pt]{\textcolor{ablationworse}{$+0.5\%$}}} & \shortstack[r]{$1.215\!\pm\!.011$\\[-1pt]{\textcolor{ablationworse}{$+0.6\%$}}} & \shortstack[r]{$0.722\!\pm\!.002$\\[-1pt]{\textcolor{ablationworse}{$+1.1\%$}}} & \shortstack[r]{$1.058\!\pm\!.002$\\[-1pt]{\textcolor{ablationworse}{$+1.2\%$}}} & \shortstack[r]{$1.194\!\pm\!.009$\\[-1pt]{\textcolor{ablationbetter}{$-0.4\%$}}} & \shortstack[r]{$1.785\!\pm\!.014$\\[-1pt]{\textcolor{ablationbetter}{$-0.3\%$}}} \\
P11 & \shortstack[r]{$0.827\!\pm\!.003$\\[-1pt]{\textcolor{ablationbetter}{$-0.6\%$}}} & \shortstack[r]{$1.204\!\pm\!.005$\\[-1pt]{\textcolor{ablationbetter}{$-0.4\%$}}} & \shortstack[r]{$0.720\!\pm\!.003$\\[-1pt]{\textcolor{ablationworse}{$+0.8\%$}}} & \shortstack[r]{$1.055\!\pm\!.003$\\[-1pt]{\textcolor{ablationworse}{$+0.9\%$}}} & \shortstack[r]{$1.195\!\pm\!.005$\\[-1pt]{\textcolor{ablationbetter}{$-0.4\%$}}} & \shortstack[r]{$1.788\!\pm\!.009$\\[-1pt]{\textcolor{ablationbetter}{$-0.2\%$}}} \\
\bottomrule
\end{tabular}}
\par\endgroup
\caption{Preset ablation scores.}
\label{fig:preset-transfer-scores}
\end{subfigure}\hfill
\begin{subfigure}[b]{0.445\linewidth}
\centering
\includegraphics[width=\linewidth]{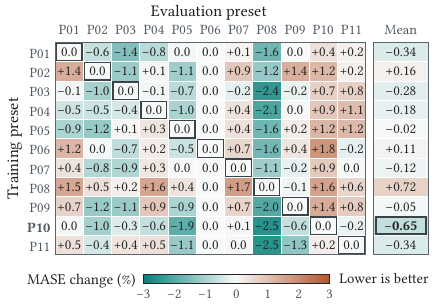}
\caption{Cross-preset MASE.}
\label{fig:preset-transfer-heatmap}
\end{subfigure}
\caption{Generator preset comparison. Panel (a) reports checkpoint
means and standard deviations with percentage changes from the reference
preset P02. Appendix~\ref{sec:preset-search} defines the presets.
Panel (b) shows percentage changes in five-checkpoint
mean MASE relative to each outlined diagonal score. Each evaluation
preset uses 4K shared synthetic series. The Mean column weights all
11 evaluation presets equally. }
\label{fig:preset-transfer-mase}
\end{figure}

\paragraph{Generator Composition.}
We explore eleven presets of the synthetic generator (P01--P11). The variants
change the mixture weights of process families, the distribution of seasonal
component counts, or the probabilities of observation profiles. For each
preset, we train a separate instance of the same forecasting backbone
under a common training protocol. Figure~\ref{fig:preset-transfer-mase}
pairs the scores on the development subsets and synthetic validation bank
with the cross-preset MASE comparison. P10, which omits the structural branch,
achieves the lowest mean relative MASE.

\paragraph{Patch Length and Stride.}
Relative to length and stride 32, setting both to 16 lowers all four
real-data errors by 0.87--1.70\%. This change slightly reduces the
parameter count and increases the token count from 37 to 73 at the
tested context and horizon. Setting both to 64 raises all four errors
(Table~\ref{tab:ablation-patching}).

\begin{table}[!htbp]
\centering
\caption{Patch length and stride. Changes are relative to the
length-32, stride-32 control.}
\label{tab:ablation-patching}
\scriptsize
\setlength{\tabcolsep}{0.9pt}
\reporttablestyle[3]
\begin{tabular}{@{}lrrrrrr@{}}
\toprule
& \multicolumn{2}{c}{\reporttablehead{FEV-Bench-S}} &
\multicolumn{2}{c}{\reporttablehead{GIFT-Eval-S}} &
\multicolumn{2}{c}{\reporttablehead{Synthetic bank}} \\
\cmidrule(lr){2-3}\cmidrule(lr){4-5}\cmidrule(l){6-7}
\reporttablehead{Length / stride} &
\reporttablehead{SQL $\downarrow$} & \reporttablehead{MASE $\downarrow$} &
\reporttablehead{SQL $\downarrow$} & \reporttablehead{MASE $\downarrow$} &
\reporttablehead{SQL $\downarrow$} & \reporttablehead{MASE $\downarrow$} \\
\midrule
16 / 16 & \shortstack[r]{$\mathbf{0.8206}\pm.0013$\\[-1pt]{\scriptsize\textcolor{ablationbetter}{$\Delta$ -1.39\%}}} & \shortstack[r]{$\mathbf{1.1928}\pm.0019$\\[-1pt]{\scriptsize\textcolor{ablationbetter}{$\Delta$ -0.87\%}}} & \shortstack[r]{$\mathbf{0.7094}\pm.0023$\\[-1pt]{\scriptsize\textcolor{ablationbetter}{$\Delta$ -1.70\%}}} & \shortstack[r]{$\mathbf{1.0425}\pm.0028$\\[-1pt]{\scriptsize\textcolor{ablationbetter}{$\Delta$ -1.44\%}}} & \shortstack[r]{$\mathbf{1.1246}\pm.0053$\\[-1pt]{\scriptsize\textcolor{ablationbetter}{$\Delta$ -6.12\%}}} & \shortstack[r]{$\mathbf{1.6878}\pm.0040$\\[-1pt]{\scriptsize\textcolor{ablationbetter}{$\Delta$ -5.51\%}}}  \\
32 / 32 & \shortstack[r]{$0.8322\pm.0058$\\[-1pt]{\scriptsize\textcolor{ablationneutral}{$\Delta$ +0.00\%}}} & \shortstack[r]{$1.2034\pm.0044$\\[-1pt]{\scriptsize\textcolor{ablationneutral}{$\Delta$ +0.00\%}}} & \shortstack[r]{$0.7217\pm.0014$\\[-1pt]{\scriptsize\textcolor{ablationneutral}{$\Delta$ +0.00\%}}} & \shortstack[r]{$1.0577\pm.0019$\\[-1pt]{\scriptsize\textcolor{ablationneutral}{$\Delta$ +0.00\%}}} & \shortstack[r]{$1.1979\pm.0057$\\[-1pt]{\scriptsize\textcolor{ablationneutral}{$\Delta$ +0.00\%}}} & \shortstack[r]{$1.7862\pm.0084$\\[-1pt]{\scriptsize\textcolor{ablationneutral}{$\Delta$ +0.00\%}}}  \\
64 / 64 & \shortstack[r]{$0.8792\pm.0771$\\[-1pt]{\scriptsize\textcolor{ablationworse}{$\Delta$ +5.66\%}}} & \shortstack[r]{$1.2327\pm.0091$\\[-1pt]{\scriptsize\textcolor{ablationworse}{$\Delta$ +2.44\%}}} & \shortstack[r]{$0.7508\pm.0214$\\[-1pt]{\scriptsize\textcolor{ablationworse}{$\Delta$ +4.03\%}}} & \shortstack[r]{$1.0861\pm.0051$\\[-1pt]{\scriptsize\textcolor{ablationworse}{$\Delta$ +2.68\%}}} & \shortstack[r]{$1.3170\pm.0094$\\[-1pt]{\scriptsize\textcolor{ablationworse}{$\Delta$ +9.94\%}}} & \shortstack[r]{$1.9644\pm.0151$\\[-1pt]{\scriptsize\textcolor{ablationworse}{$\Delta$ +9.98\%}}}  \\
\bottomrule
\end{tabular}
\end{table}

\begin{figure}[!htbp]
\centering
\captionsetup{skip=3pt}
\includegraphics[width=\linewidth]{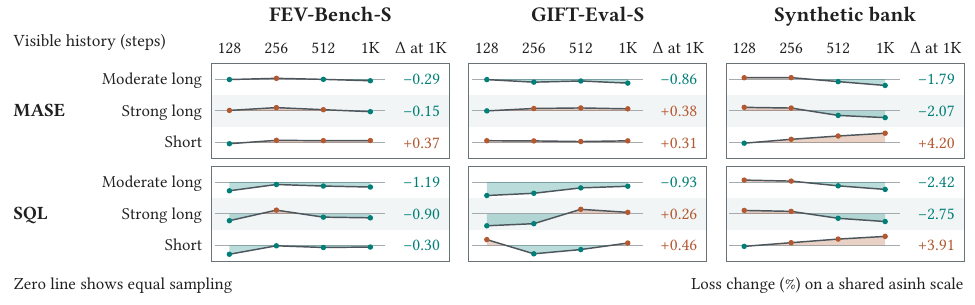}
\caption{Context-cap ablation. The values show percentage changes in
five-checkpoint mean MASE and SQL relative to equal sampling at the
same cap. Negative values indicate lower loss, and the rightmost column
of each panel gives the 1K change. Windows, targets, and full-history
scoring scales are fixed across caps (Appendix~\ref{sec:context-caps}).}
\label{fig:context-caps}
\end{figure}

\paragraph{Context Sampling.}
Surprisingly, a moderate skew toward longer training histories
lowers MASE and SQL on GIFT-Eval-S and FEV-Bench-S compared with uniform weights across the four context
buckets even at shorter contexts. A stronger skew raises both errors on GIFT-Eval-S relative
to uniform weights (Table~\ref{tab:ablation-context}). We use the moderate
skew for base pretraining. We reserve 1\% of the sampling mass for a
16--64-step bucket and distribute the remaining 99\% among the four
original buckets in proportions $(0.10,0.20,0.30,0.40)$ from shortest to
longest. Rankings vary with visible history when targets and scoring scales
are held fixed (Figure~\ref{fig:context-caps}).

\paragraph{Augmentation and Architecture.}
Removing mixup raises all six errors, and removing point masking gives
the lowest MASE and SQL on FEV-Bench-S in the augmentation study
(Table~\ref{tab:ablation-training}). Training with masking can help on
artificial gaps, with effects varying by benchmark and corruption pattern
(Figure~\ref{fig:missing-context}). The preferred mixup strength varies
across metrics (Table~\ref{tab:ablation-mixing-strength}). Value
residuals give the lowest errors on GIFT-Eval-S, and scalar residuals
give the lowest errors on FEV-Bench-S, with gains below 0.3\% against
the control. SwiGLU with width 2K lowers both losses on the synthetic
validation bank by
more than 2\%
(Table~\ref{tab:architecture-ablation}).

\section{Discussion and Conclusion}
\label{sec:discussion}
\label{sec:conclusion}
\textsc{Tafsut} combines a univariate backbone pretrained exclusively on
series from our compositional generator (open-sourced) with a separate
adapter that refines the frozen backbone's forecast quantiles using
target history and available covariates. \textsc{Tafsut}-Synth achieves the
lowest point and probabilistic errors among the fully synthetic models
compared in target-only evaluations on GIFT-Eval and FEV-Bench
(Figure~\ref{fig:synthetic-normalized-mase-8k}). After real-data
adaptation, \textsc{Tafsut} is competitive with leading open-weight zero-shot
forecasters on both benchmarks. On FEV-Bench, the adapter improves
aggregate point-forecast accuracy on tasks with covariates, with larger
gains when future covariate values are known.

We document transparently the training recipe, data sources, filtering, and overlap
audits for the separate adaptation corpora. The benchmark comparisons evaluates the zero shot performance of our model on diverse and large scale real world datasets. Controlled ablations inform the choice of generator mixture,
patch length, context sampling, and augmentation
(Section~\ref{sec:design-analysis}). The preferred generator mixture
varies across development sets and metrics. The FEV-Bench bootstrap measures variation across
tasks, and the quantile scores assess marginal forecasts.

\paragraph{Future Work.}
It remains to carefully  study the impact of preset weights, observation severity, and cross-series
dependence on which synthetic data properties improve transfer
to held-out real domains. Further work can explore the applicability of the learned representations across different timeseries tasks.   The  UCR classification study
(Appendix~\ref{app:ucr-classification}) provides a starting point for
testing whether task-specific presets improve representations across
additional classification datasets. Testing trained adapters across compatible backbones and changing
covariate availability would clarify when the quantile interface
supports reuse.

\FloatBarrier
\bibliographystyle{tmlr}
\bibliography{references}

\appendix
\section{Related Work}
\label{sec:related-work}

\paragraph{Classical Forecasting.}
Statistical forecasting models capture temporal patterns through explicit
modeling assumptions. ARIMA models use past values and forecast errors,
with differencing and seasonal terms. Exponential smoothing
updates estimates of level, trend, and seasonality \citep{hyndman2021fpp}.
These models are typically fitted separately to each time series and serve as interpretable
baselines due to their simplicity.

\paragraph{Deep Learning and Shared Representations.}
Neural methods learn temporal features and nonlinear predictive mappings
from data. Fully connected residual models such as N-BEATS operate on
lookback windows~\citep{oreshkin2020nbeats}. Recurrent models such as
DeepAR share a probabilistic forecaster across related
series~\citep{salinas2020deepar}. Temporal convolutions provide another
sequence-modeling mechanism~\citep{bai2018tcn}, while recent transformer-based architecture learn interactions through attention; this
includes temporal tokenization and patching, channel dependence, and the
organization of temporal and cross-variate processing
\citep{nie2023patchtst,zhang2023crossformer}. This setting shares representations across
series within a training collection to transfer to unseen test split.

\paragraph{Foundation Models and Cross-Dataset Transfer.}
Time-series foundation models extend shared training to heterogeneous
pretraining collections, seeking forecasts on new datasets without task-specific
weight updates. Early examples include TimeGPT and Lag-Llama
\citep{garza2023timegpt,rasul2024lagllama}, tokenized forecasting in
Chronos~\citep{ansari2024chronos}, patched decoding in
TimesFM~\citep{das2024timesfm}, and universal forecasting in
Moirai~\citep{woo2024moirai}. Subsequent releases include Chronos-Bolt and
Chronos-2~\citep{amazon2024chronosbolt,ansari2025chronos2}, TimesFM 2.0,
2.5, and 3.0~\citep{google2026timesfmversions,google2026timesfm3}, and
Moirai-MoE and Moirai 2.0~\citep{liu2024moiraimoe,liu2025moirai2}.
The Timer family studies long-context and sparse-expert forecasting
\citep{liu2024timer,liu2025timerxl,liu2026timers1}. Time-MoE also scales
forecasting with sparse experts, and Xihe uses hierarchical attention to
model multiple temporal scales~\citep{shi2025timemoe,sun2025xihe}.
MOMENT pretrains representations for forecasting and other time-series
tasks~\citep{goswami2024moment}.
\textsc{Tafsut} learns a forecasting prior from a compositional synthetic distribution,
with real-data and multivariate adaptation as subsequent stages.

\paragraph{Synthetic Pretraining and Augmentation.}
Synthetic data can supplement real data or provide all the data for
pretraining. Chronos augments real series with samples from Gaussian
processes~\citep{ansari2024chronos}, and ForecastPFN learns from a
synthetic prior~\citep{dooley2023forecastpfn}. TimePFN generates
multivariate series from kernel-based sources using linear
coregionalization~\citep{taga2025timepfn}. ApolloPFN generates synthetic
series with exogenous dependencies~\citep{potapczynski2026apollopfn}, and
Tabby combines real series with kernel-based and structural causal
generators~\citep{xie2026tabby}. Other approaches construct cross-series
dependencies from univariate sources, making the types of relationships
in the training data a design choice
\citep{ansari2025chronos2,podest2026tirex2,meyer2026t0}. \textsc{Tafsut} separates
univariate dynamics from covariate generation. A compositional generator
produces the base series, and separate covariate mechanisms introduce
shared sources, delayed effects, nonlinear relationships, and modulation.
In the first stage, we train a standalone model exclusively on synthetic
data and compare it with Chronos-2 Synth~\citep{chronos2synth}, TabPFN-TS,
which applies a synthetically pretrained tabular backbone to temporal
features~\citep{hoo2026tabpfnts}, and TempoPFN, a linear recurrent forecaster
pretrained on augmented synthetic time series~\citep{moroshan2026tempopfn}
(Figure~\ref{fig:synthetic-normalized-mase-8k}). TabPFN-TS-3 further
fine-tunes its tabular backbone on synthetic time series
\citep{grinsztajn2026tabpfn3}.

\paragraph{Temporal Representations and Probabilistic Outputs.}
Forecasting architectures differ in how they represent history, exchange
temporal information, and parameterize uncertainty. Patches shorten the
token sequence while retaining local temporal structure, as in PatchTST,
Granite PatchTST-FM, and TimesFM~\citep{nie2023patchtst,ibm2026patchtstfm,das2024timesfm}.
TS-ICL models timestamped observations and Falcon-X models interactions
in a shared latent space~\citep{lenaour2026tsicl,liu2026falconx}.
TiRex and TiRex-2 use xLSTM blocks for temporal mixing
\citep{auer2025tirex,podest2026tirex2}. FlowState uses a state-space
encoder, and HART combines recurrence with attention
\citep{graf2026flowstate,lin2026hart}. Tiny Time Mixers, Kairos, and
Reverso explore parameter-efficient forecasting
\citep{ekambaram2024ttm,feng2026kairos,fu2026reverso}.
Toto parameterizes a predictive distribution for sampling, and Sundial
generates probabilistic forecasts with flow matching
\citep{cohen2024toto,liu2025sundial}. Direct marginal quantiles are used
by Chronos-2, Toto 2.0, and TimesFM 3.0
\citep{ansari2025chronos2,khwaja2026toto2,google2026timesfmversions}.
\textsc{Tafsut} combines patch representations with joint historical and
future-query encoding and direct quantile prediction. Its outputs describe
uncertainty at each horizon step. Assessing calibration and dependence
across future steps requires additional evaluation.

\paragraph{Multivariate and Covariate-Conditioned Forecasting.}
Sharing a forecaster across channels differs from conditioning a target
on other channels. Joint models must represent both within-series temporal
structure and cross-series dependence, while distinguishing historical
covariates from information available over the forecast horizon. Existing
designs include attention over flattened variate--time tokens and
factorized temporal and variate processing
\citep{woo2024moirai,zhang2023crossformer,ansari2025chronos2,podest2026tirex2}.
TimesFM 3.0, $t_0$, Timestral, TimeH2O, and CITRAS-FM also support
covariate-informed forecasting
\citep{google2026timesfm3,meyer2026t0,petralia2026timestral,artley2026timeh2o,yamaguchi2026citrasfm}.
ChronosX and CoRA add covariate modules to pretrained backbones
\citep{arango2025chronosx,qin2025cora}. In-context approaches include
COSMIC, iAmTime, and in-context fine-tuning with demonstration series
\citep{auer2025cosmic,saha2026iamtime,faw2025incontext}.
\textsc{Tafsut} separates covariate conditioning into a trainable adapter that
refines the forecast quantiles of a frozen univariate backbone using
target history and available covariates. The quantile interface allows
the same adapter architecture to be paired with different backbones
that provide compatible forecasts. On FEV-Bench, adding the adapter
reduces geometric mean MASE by 11.2\% on tasks with known-future covariates
and by 1.1\% on tasks with past covariates only
(Figure~\ref{fig:fev-covariate-effect}).

\paragraph{Evaluation of Transfer.}
GIFT-Eval and FEV-Bench evaluate transfer across heterogeneous forecasting
tasks~\citep{aksu2024gift,shchur2025fev}. Their task definitions, overlap
policies, and aggregation rules differ. For each benchmark, we specify the
evaluated tasks, their use in model selection, the visible context, and how
we compute and aggregate scores.

\section{Synthetic Generator Details}
\label{app:generator-details}
The component constructions below expand the synthetic mixture in
Section~\ref{sec:synthetic-generator}. Each preset activates a compatible
subset of the mechanisms in Table~\ref{tab:generator-phenomena}.

\begin{table}[!htbp]
\centering
\caption{Phenomena induced by the generator. Rows describe modules and their variants, not mutually exclusive source components or additive terms. Activation and parameter priors depend on the preset.}
\label{tab:generator-phenomena}
\begingroup\small
\setlength{\tabcolsep}{4pt}
\renewcommand{\arraystretch}{1.12}
\reporttablestyle
\begin{tabular}{@{}p{.17\linewidth}p{.25\linewidth}p{\dimexpr.58\linewidth-16pt\relax}@{}}
\toprule
\raggedright \reporttablehead{Module} &
\raggedright \reporttablehead{Induced phenomena} &
\raggedright \reporttablehead{Construction and variants}\tabularnewline
\midrule
\raggedright Atomic functions and motifs & \raggedright Smooth ramps, plateaus, sharp resets, repeated local patterns & \raggedright Bounded anchors; linear, PCHIP, constant, or interval-wise interpolation; optional jumps and endpoint bridging. Symbol templates use alternation, blocks, clustered or bursty runs, rotations, palindromes, and random assignments.\tabularnewline
\raggedright Recurrence & \raggedright Single or multiple scales, harmonics, beats, evolving cycles & \raggedright Canonical sine, square, triangular, and sawtooth waves; slot levels/activity, piecewise blocks, cyclic pulses, and richer motifs. Period and amplitude priors control scale balance; cross-scale modulation and evolving shape mixtures relax exact repetition.\tabularnewline
\raggedright Latent calendars & \raggedright Repeated or nested schedules; recurring inactivity & \raggedright Phase-based templates with coupled periods; periodic, smooth, or cycle-wise basis mixing; additive, multiplicative, or suppressive carrier application.\tabularnewline
\raggedright Envelopes, clocks, and trends & \raggedright Changing magnitude or timing; drift, saturation, level revisits & \raggedright Positive amplitude envelopes and integrated local clocks; linear, bounded-smooth, random-walk, piecewise-linear, step, finite-level, logistic, exponential, mean-reverting, and damped-slope trends.\tabularnewline
\raggedright Events and breaks & \raggedright Recurring peaks, isolated shocks, lasting structural changes & \raggedright Gaussian, triangular, half-sinusoidal, or asymmetric exponential pulses; impulses with rise and decay; hazard-driven changes in level, slope, oscillation amplitude, or observation variance.\tabularnewline
\raggedright Stochastic dynamics & \raggedright Serial dependence, volatility clustering, finite-range memory, delayed responses & \raggedright ARMA and seasonal recursion; constant, smooth, or conditional innovation variance, optionally asymmetric; truncated fractional filters; direct and lagged effects of hidden structured drivers.\tabularnewline
\raggedright Activity and arrivals & \raggedright Predictable sparse support, intermittent demand, clustered or regulated events & \raggedright Periodic, smooth, or semi-Markov gates; exact or measured zeros; Bernoulli--Gamma demand; self-exciting arrivals with decaying marked responses or self-correcting arrivals.\tabularnewline
\raggedright Counts and regimes & \raggedright Discrete overdispersion, excess zeros, persistent operating conditions & \raggedright Poisson, negative-binomial, and zero-inflated counts; Markov or explicit-duration semi-Markov states controlling level, amplitude, and residual dynamics.\tabularnewline
\raggedright Observation & \raggedright Scale diversity, measurement uncertainty, recording artifacts & \raggedright Support-compatible affine transforms; Gaussian, Student, Laplace, skew-normal, correlated, or heteroscedastic noise; bias, baseline wander, gain drift, spikes, saturation, quantization, censoring, stuck values, timing perturbations, and coarse rounding.\tabularnewline
\raggedright Canonical bank & \raggedright Pulses, counters, capacity limits, hysteresis, refractory recurrence & \raggedright Five direct process laws with fixed total source share; optional seasonal, trend, residual, intermittency, affine, and observation overlays disabled.\tabularnewline
\bottomrule
\end{tabular}
\endgroup
\end{table}

\begin{figure}[t]
\centering
\subcaptionbox{Bounded anchors and feasible values\label{fig:atomic-anchors}}[.32\linewidth]{%
\includegraphics[width=\linewidth,height=3.1cm,keepaspectratio]{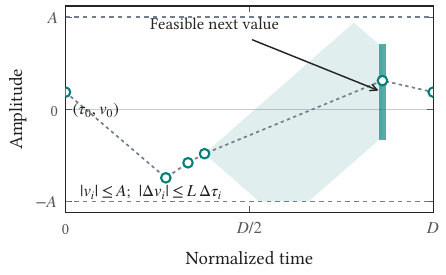}}\hfill
\subcaptionbox{Interpolation through the same anchors\label{fig:atomic-interpolation}}[.32\linewidth]{%
\includegraphics[width=\linewidth,height=3.1cm,keepaspectratio]{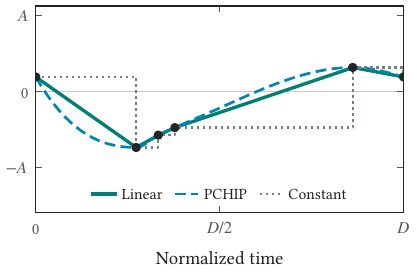}}\hfill
\subcaptionbox{Bridged and unbridged repetition\label{fig:atomic-bridging}}[.32\linewidth]{%
\includegraphics[width=\linewidth,height=3.1cm,keepaspectratio]{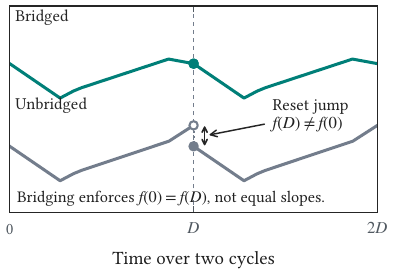}}
\caption{Atomic functions from bounded anchors. The curves in (b) use the
same anchors with different interpolation rules. Matching endpoint values
in (c) removes the reset jump. Traces in (c) are vertically offset.
Shapes are shown before normalization and rescaling.}
\label{fig:generator-atomic}
\end{figure}

\paragraph{Atomic Function Sampling.}
\label{sec:atomic-shapes}
For the continuous construction of Section~\ref{sec:synthetic-generator},
we sample the number, positions, and values of anchors to vary shape
complexity at fixed duration and amplitude bounds. With a prescribed
endpoint $v_m$, forward and endpoint reachability constrain each
intermediate value $v_i$ to the interval
\begin{equation}
 \begin{split}
 \mathcal I_i={}&[-A,A]
 \cap[v_{i-1}-L(u_i-u_{i-1}),\,
    v_{i-1}+L(u_i-u_{i-1})] \cap[v_m-L(D-u_i),\,v_m+L(D-u_i)].
 \end{split}
 \label{eq:bridge-feasibility}
\end{equation}
An unconstrained endpoint uses the first two intervals. For PCHIP, the
bound $L$ applies to anchor secants. Piecewise-constant and interval-wise
interpolation can introduce discontinuities, and an explicit jump can
bypass the secant bound. Endpoint bridging enforces equality of values
while permitting unequal endpoint slopes. The reset of an unbridged
function is $\varphi(0)-\varphi(D)$. Optional centering, standardization,
and rescaling act after the bounds on the original primitive have been applied
(Figure~\ref{fig:generator-atomic}). Table~\ref{tab:generator-defaults}
lists the standalone sampler defaults.

\begin{table}[!htbp]
\centering
\caption{Univariate generator settings. Active components draw atomic
parameters from their preset. The canonical weights are shares of the
full mixture. Renewal ranges are given before rounding and flooring.}
\label{tab:generator-defaults}
\begingroup\footnotesize
\setlength{\tabcolsep}{4pt}
\renewcommand{\arraystretch}{1.0}
\reporttablestyle
\begin{tabularx}{\linewidth}{@{}l>{\raggedright\arraybackslash}Xr@{}}
\toprule
\reporttablehead{Symbol} & \reporttablehead{Setting} & \reporttablehead{Default or range}\\
\midrule
\reporttablegroup{3}{Standalone atomic sampler}\\
$(m+1,A,L)$ & Anchor count, amplitude bound, and secant-slope bound & $(5,10,1)$\\
$\varphi$ & Interpolation, endpoint bridging, and jumps & Linear, bridged, jump-free\\
\midrule
\reporttablegroup{3}{Time-varying controls and observation}\\
$[\lambda_{\min},\lambda_{\max}]$ & Clipping range for the evolution rate & $[10^{-5},0.25]$\\
$c_\lambda$ & Rate multiplier in the persistence & $10$\\
$\varepsilon$ & Stabilizer for calendar suppression and signal power & $10^{-12}$\\
\midrule
\reporttablegroup{3}{Final mixture and canonical process bank}\\
$B$ & Active components & $24$\\
$\vect{\pi}_{\mathrm{can}}$ & Pulse, counter, stock, hysteresis, and renewal weights &
$(2.0,1.0,0.8,0.7,0.5)\%$\\
$j_{\max}$ & Maximum pulse timing jitter (steps) & $1$\\
$J$ & Counter jump size support & $\{1,2,3,5\}$\\
$p_{\mathrm{reset}}$ & Probability of periodic counter resets & $0.35$\\
$\theta_{\mathrm{hys}}$ & Magnitude of the hysteresis thresholds & $0.35$\\
$(a_S,a_d)$ & Gains on the hysteresis state and driver & $(2,0.08)$\\
$d_{\min}$ & Minimum renewal gap & $\operatorname{Uniform}(C/64,C/16)$\\
$\mu$ & Mean extra renewal wait & $\operatorname{Uniform}(C/32,C/8)$\\
$d_{\mathrm{floor}}$ & Lower bound on both renewal parameters (steps) & $2$\\
\bottomrule
\end{tabularx}
\endgroup
\end{table}

\paragraph{Symbolic Motifs.}
\label{sec:motif-generation}
A finite alphabet assigns a normalized shape $f_k:[0,1]\to\mathbb R$ to each symbol. A template selects symbols $z_j$, scales $a_j$, and offsets $c_j$ on phase intervals: $\mathcal T(\phi)=a_jf_{z_j}((\phi-u_j)/(u_{j+1}-u_j))+c_j$ for $\phi\in[u_j,u_{j+1})$. Equal-width slots $u_j=j/J$ with integer sample width $d$ give period $Q=Jd$. Symbols retain morphology; arrangements and proportions vary, including dominant motifs with rare alternatives. Shape comparisons discourage redundant symbols.

\paragraph{Multiscale and Evolving Recurrence.}
Periods $P_j$, offsets $\delta_j$, and amplitudes $a_j$ define an additive carrier, $s_t=\sum_{j=1}^K a_jf_j(\{(t+\delta_j)/P_j\})$, where braces denote fractional part. Absolute and context-relative period priors, including logarithmic sampling, span orders of magnitude; $C/P_j$ measures visible cycles, including partially observed ones. Component counts and geometric-decay, simplex, or dominant-component amplitude allocations vary spectral concentration. Harmonic periods, nearby-period beats, and cross-scale modulation add dependent scales. Independently sampled shapes sharing a period can evolve through $f_t(\phi)=\sum_{r=1}^R\alpha_{r,t}f^{(r)}(\phi)$, with $\alpha_{r,t}\geq0$ and $\sum_r\alpha_{r,t}=1$. Slowly varying bounded states drive the weights, changing peaks, asymmetry, and slopes while retaining periodic organization.

\paragraph{Latent Calendars and Mixing.}
\label{sec:latent-calendar}
Calendar bases use latent phase rather than observed dates or named covariates: $h_{k,t}=b_k\mathcal T_k(\{(t+\xi_k)/Q_k\})$. Periods $Q_k$ can couple to seasonal scales through integer, rational, or continuous ratios, producing nested subcycles and supercycles. Strengths $b_k$ and templates $\mathcal T_k$ set each schedule; several bases combine as $h_t=\sum_{k=1}^M w_{k,t}h_{k,t}$, with $w_{k,t}\geq0$ and $\sum_kw_{k,t}=1$. For two bases, periodic weights are $w_{1,t}=r_t$, $w_{2,t}=1-r_t$. Period $R$, duty fraction $q$, phase center $c$, and sharpness $\beta$ define a softened phase box:
\begin{equation}
 d_t=\left|\left(\left\{t/R\right\}-c+\tfrac12\right)\bmod 1-\tfrac12\right|,
 \qquad r_t=\operatorname{sigmoid}\!\left[\beta(q/2-d_t)\right].
 \label{eq:periodic-calendar-mixer}
\end{equation}
The weights repeat after $R$, which need not equal a basis period; more bases use normalized softened periodic windows. Smooth mixing applies a temperature-controlled softmax to correlated evolving states. Cycle-wise mixing holds their logits fixed for each $\lfloor(t+\xi_{\rm ref})/Q_{\rm ref}\rfloor$, using a randomly chosen reference basis; other bases need not align. One reference cycle triggers smooth mixing; one basis has unit weight. A separate persistent cycle-state process can change templates over groups of cycles.

\begin{figure}[t]
\centering
\subcaptionbox{Motifs from atomic anchors\label{fig:calendar-alphabet}}[.485\linewidth]{%
\includegraphics[width=\linewidth]{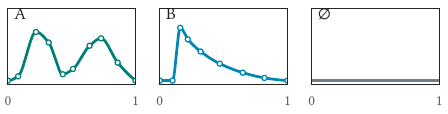}}\hfill
\subcaptionbox{A repeating symbol template\label{fig:calendar-template}}[.485\linewidth]{%
\includegraphics[width=\linewidth]{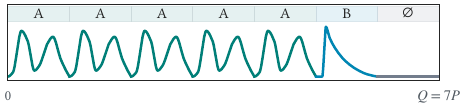}}\par\smallskip
\subcaptionbox{Periodic schedule mixing\label{fig:calendar-periodic-mixer}}[.485\linewidth]{%
\includegraphics[width=\linewidth]{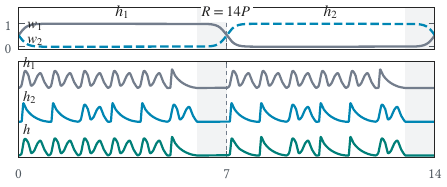}}\hfill
\subcaptionbox{Carrier and calendar applications\label{fig:calendar-application}}[.485\linewidth]{%
\includegraphics[width=\linewidth]{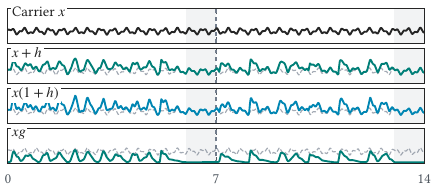}}
\caption{Motifs and latent schedules. Open circles mark atomic anchors; curves use PCHIP. Periodic weights $w_1=r(t)$, $w_2=1-r(t)$ alternate two seven-slot bases, with mixer period $R=2Q=14P$. The mixed field is $h=w_1h_1+w_2h_2$. Panel (d) shows the same carrier before and after additive, multiplicative, and suppressive application; dashed curves repeat the carrier. Panels (c,d) show one full mixer cycle ($0\leq t/P\leq14$); shading marks shared inactivity. These are illustrative, date-free constructions; schedule traces are vertically offset.}
\label{fig:generator-calendar}
\end{figure}

\paragraph{Carrier Application.}
The mixed schedule acts additively, multiplicatively, or suppressively as $\widetilde s_t=s_t+h_t$, $s_t(1+h_t)$, or $s_tg_t$, respectively. Suppression uses generated-span extrema, $g_t=(h_t-\min_u h_u)/(\max_u h_u-\min_u h_u+\varepsilon)$, with the stabilizer $\varepsilon$ given in Table~\ref{tab:generator-defaults}. Mixing determines the schedule; application controls its effect. Additive schedules can act on a zero carrier; multiplication and suppression require a nonzero one, enforced by compatibility sampling. Figure~\ref{fig:generator-calendar} illustrates seven-slot cycles, a two-cycle mixer, and shared final inactivity; these are explanatory choices. Inactive symbols may occur anywhere, recur, or be absent.

\paragraph{Amplitude and Timing.}
Let $\lambda$ denote the evolution rate clipped to
$[\lambda_{\min},\lambda_{\max}]$. The rate multiplier $c_\lambda$
sets the persistence $\rho=\exp(-c_\lambda\lambda)$.
Table~\ref{tab:generator-defaults} gives the clipping bounds and multiplier.
Given independent standard normal draws $z_0,\epsilon_1,\epsilon_2,\ldots$, we
generate $z_t=\rho z_{t-1}+\sqrt{1-\rho^2}\epsilon_t$ and
$u_t=\tanh z_t$. A smaller $\lambda$ makes $u_t$ more persistent.
A positive envelope gives
$a_t=a_{\min}+(a_{\max}-a_{\min})(u_t+1)/2$ and $\widetilde s_t=a_ts_t$.
Integrating a positive local clock rate deforms the timing. Interpolation
stretches or delays a seasonal component or composite, with source
coordinates at or before the current time.

\paragraph{Trends and Structural Changes.}
Trends follow amplitude modulation, allowing independent scales. The
bounded-state trend scales $u_t$ by a configured amplitude.
Mean reversion and damped slopes use $\ell_t=\ell_{t-1}+\kappa(\mu-\ell_{t-1})+\sigma\epsilon_t$ and $v_t=\varphi v_{t-1}+\sigma_v\zeta_t$, $\ell_t=\ell_{t-1}+v_t$, with innovations $\epsilon_t,\zeta_t$. Changepoints can jointly or separately change level, slope, oscillation amplitude, or observation variance; transition widths interpolate between abrupt and gradual changes. A forward hazard samples locations without deliberate alignment to the context--forecast boundary. Level/slope changes affect the latent signal; variance changes propagate to observation.

\paragraph{Localized Events.}
Kernels with sampled amplitude, sign, width $d_j$, and location $\tau_j$ give $e_t=\sum_{j=1}^J a_jk_j((t-\tau_j)/d_j)$. Sampled scales span small perturbations to rare large peaks. Placement distinguishes recurring motif/calendar pulses (clustered or separated), one-time latent impulses, and measurement spikes: similar shapes can encode recurrence, shocks, or corruption.

\paragraph{Residuals and Volatility.}
ARMA residuals follow $r_t=\sum_{i=1}^p\phi_ir_{t-i}+\epsilon_t+\sum_{j=1}^q\theta_j\epsilon_{t-j}$. Orders and coefficients vary; bounded partial-autocorrelation sampling stabilizes the AR dynamics. Optional seasonal recursion adds dependence at a visible lag. Residual scale uses latent-context calibration with a positive floor for weak carriers. The simple-null control disables the structured carrier and sets $p=q=0$ with constant innovation scale, before observation effects. Innovation variance is constant, smooth, or conditional:
\begin{align}
 \epsilon_t&=\sqrt{q_t}\,z_t,\\
 q_t&=\omega+\alpha\epsilon_{t-1}^2+\beta q_{t-1}
    +\gamma\epsilon_{t-1}^2
          \ind\{\epsilon_{t-1}<0\}.
 \label{eq:conditional-volatility}
\end{align}
For standardized Gaussian $z_t$, the optional asymmetric term distinguishes innovation signs; persistence constraints prevent explosive variance. This endogenous volatility differs from measurement corruption.

\paragraph{Memory and Delayed Response.}
A truncated fractional filter gives $m_t=\sum_{j=0}^{J-1}c_j\epsilon_{t-j}$, with $c_0=1$ and $c_j=c_{j-1}(j-1+d)/j$, where $d$ controls persistence or antipersistence over the finite lag range; this approximates, rather than exactly implements, infinite memory. Context-calibrated contributions affect latent level, measurement-noise scale, or both. A hidden canonical or motif driver adds $x'_t=x_t+a_0d_t+\sum_{j=1}^J a_jd_{t-\lambda_j}$, with $\lambda_j>0$; the model infers this unobserved driver from the univariate series.

\paragraph{Sparse Activity.}
Periodic, smooth, or explicit-duration semi-Markov gates create recurring windows, gradual onsets, or persistent states through $x'_t=g_tx_t+(1-g_t)\nu_t$, $0\leq g_t\leq1$. Here $\nu_t$ is an off-state noise floor; varying duty cycle controls sparsity. Exact-zero constructions restore inactive zeros after measurement transforms. Measured-zero constructions allow noise around latent inactivity. Activity schedules can make the support predictable. The resulting zeros are valid observations, and missing entries are encoded as $\infty$.

\paragraph{Counts and Demand.}
Nonnegative latent intensity drives Poisson or overdispersed negative-binomial counts; for $r>0$, $\mathbb E[N_t\mid\lambda_t]=\lambda_t$ and $\operatorname{Var}(N_t\mid\lambda_t)=\lambda_t+\lambda_t^2/r$. Zero inflation adds an inactive state beyond zeros explained by low intensity. Intermittent demand separates arrival and magnitude: $D_t=B_tV_t$, with $B_t\sim\operatorname{Bernoulli}(p_t)$ and $V_t\sim\operatorname{Gamma}(a,b)$. Gamma parameters $a,b$ are shape and scale; magnitudes need not be integers. Latent structure controls $p_t$, optionally suppressed by a periodic gate. Self-exciting arrivals raise subsequent intensity with decaying excitation and accumulated positive marked responses; self-correcting arrivals lower propensity after events and increase it with elapsed time. Bounded discrete-time intensities control both constructions.

\paragraph{Persistent Regimes.}
Markov states favor self-transitions; semi-Markov states sample explicit durations. States control level, amplitude, and local residual dynamics through a transformed carrier plus state-dependent noise, and can revisit earlier conditions. This differs from lasting changepoints and phase-associated calendar variation.

\paragraph{Scale and Observation.}
Support-compatible affine transforms vary raw magnitude and location.
Continuous signals allow signs and offsets. Nonnegative and count
constructions retain their support, and prescribed exact zeros exclude
incompatible transforms. Clean, noisy, and artifact-rich profiles sample
measurement families and recording effects from Table~\ref{tab:generator-phenomena}.
For transformed latent context $\widetilde{\vect{x}}$, the context signal
power is $P_C=C^{-1}\sum_{t=1}^C\widetilde x_t^2+\varepsilon$.
The SNR $q$ in decibels sets nominal noise variance $\sigma^2=P_C10^{-q/10}$.
Smooth envelopes and latent variance multipliers vary uncertainty over time.
Conditional activation and effect ordering distinguish recording artifacts
from latent changes. Compatible transforms, projection, or rounding preserve
support. High-dynamic-range signals use observation transforms suited to
their scale. Clean profiles retain small noise but suppress most structured
artifacts. Artifact-rich profiles can retain a high SNR.

\paragraph{Canonical Process Bank.}
\label{sec:generator-canonical-bank}
The final mixture assigns source weights $\vect{\pi}_{\mathrm{can}}$ to
five direct laws (Table~\ref{tab:generator-defaults}). Their timing
scales with $C$, and they bypass seasonal, trend, residual,
intermittency, affine, and observation overlays.

\paragraph{Pulses and Counters.}
Pulse trains repeat needle, triangular, plateau, or shoulder kernels, with
occasional omissions, timing jitter of at most $j_{\max}$ steps, or height
perturbations. Counters accumulate jumps as $X_t=\sum_{u=b(t)}^t J_u$,
where $J_u$ is the total of the jump sizes $J$ at time $u$.
Events are sampled on the generated grid and may coincide.
With probability $p_{\mathrm{reset}}$, counters reset periodically and
$b(t)$ is the most recent reset time. Before the first reset, or if resets
are disabled, $b(t)$ is the trajectory origin.

\paragraph{Stocks and Hysteresis.}
Stocks hold between signed updates projected onto the interval defined by
the sampled capacity $K$, as in
$X_{\tau_j}=\operatorname{clip}(X_{\tau_j^-}+\Delta_j,0,K)$.
Hysteresis uses a sinusoidal driver $d_t$ perturbed by a bounded state.
The state $S_t$ switches to $1$ at $d_t\geq\theta_{\mathrm{hys}}$ and to
$0$ at $d_t\leq-\theta_{\mathrm{hys}}$. Between thresholds it retains
$S_{t-1}$. The output is $X_t=a_S S_t+a_d d_t$.

\paragraph{Refractory Renewal.}
Variable positive pulse heights and widths accompany gaps
$\tau_{j+1}-\tau_j=d_{\min}+\lfloor E_j\rfloor$, with
$E_j\sim\operatorname{Exponential}(\text{mean }\mu)$.
The generator draws $d_{\min}$ and $\mu$ from context-scaled ranges, rounds
them to integers, and enforces a lower bound $d_{\mathrm{floor}}$
(Table~\ref{tab:generator-defaults}). This enforces a refractory interval
without exact periodicity.

\paragraph{Composition and Sampling.}
The ordered program is: (1) initialize the carrier, canonical law, or zero signal and apply any early warp; (2) apply calendars, delayed drivers, the amplitude envelope, and any pre-trend warp; (3) add trends/impulses, any pre-changepoint warp, and structural changes; (4) add ARMA/volatility and fractional-memory level terms, retaining structural/memory variance multipliers; (5) apply intermittency, count, event, or regime transforms and latent checks; (6) apply compatible affine/observation transforms, restore prescribed exact zeros, and validate. Disabled modules are identities; enabled modules may have internal activation probabilities, and canonical-bank laws bypass overlays. Order distinguishes gating from addition and count transforms from discrete residuals. Recurrence-rich, stochastic, and sparse components emphasize phase/morphology, dependence/volatility, and support/timing, respectively.

The generator screen changes the prevalence of sparse events, shared
dynamics, and structural components, the number of seasonal
components, and the observation-profile probabilities
(Figure~\ref{fig:preset-transfer-scores}). The later curriculum mixes
structure-focused and counts-and-artifacts components with the
fixed canonical share (Appendix~\ref{app:training-lineage}).

Validation checks finiteness, numerical scale, and nondegenerate structure where required. Enabled checks precede measurement corruption so noise cannot hide a degenerate latent construction. 

\paragraph{Multivariate Generator.}
\label{sec:adapter-covariate-families}
We train the adapter on a mixture of eight synthetic relationship families
that compose baseline dynamics, channel dependence, and modulation.
We generate complete trajectories before applying observation
transformations and marking unavailable entries with $\infty$.

\begin{figure}[tbp]
\centering
\captionsetup[subfigure]{skip=3pt,justification=centering}
\begin{subfigure}[t]{0.350\linewidth}
\centering\vspace{0pt}
\includegraphics[width=\linewidth]{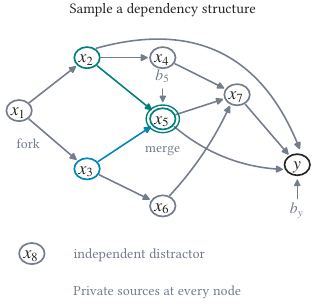}
\caption{Choose relationships}\label{fig:multivariate-generator-topology}
\end{subfigure}\hfill
\begin{subfigure}[t]{0.279\linewidth}
\centering\vspace{0pt}
\includegraphics[width=\linewidth]{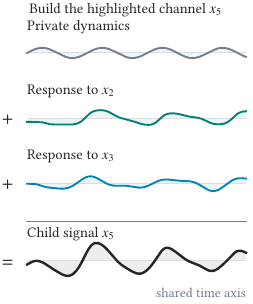}
\caption{Compose signals}\label{fig:multivariate-generator-composition}
\end{subfigure}\hfill
\begin{subfigure}[t]{0.346\linewidth}
\centering\vspace{0pt}
\includegraphics[width=\linewidth]{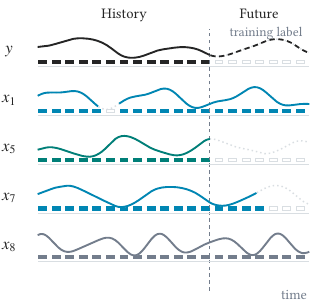}
\caption{Reveal inputs, retain labels}\label{fig:multivariate-generator-observation}
\end{subfigure}
\caption{Constructing a synthetic multivariate forecasting task (DAG example).
(a) Branching, converging, and shortcut dependencies connect seven covariates
to the target; $x_8$ is an independent distractor. Colored edges identify the
parents of ringed $x_5$. Private sources at every node are omitted except
$b_5$ and $b_y$. (b) Private dynamics and signed, delayed parent responses add
pointwise to $x_5$. Rows share time and amplitude scales, with vertical display
offsets; optional gating is omitted. (c) Complete trajectories are generated
before unavailable entries are replaced by $\infty$. Filled
rails mark available inputs, and hollow rails mark missing entries. Faint
traces illustrate the hidden values. The dashed future target is a training label.
Arrows denote generation dependencies, not attention weights.
All trajectories are constructed schematics.}
\label{fig:multivariate-generator}
\end{figure}

Figure~\ref{fig:multivariate-generator} illustrates this construction for the
graph-based families. A covariate can affect the target directly or through
other channels, while an independent distractor has no generating effect.
The highlighted child combines private dynamics with its parents' responses
and can itself drive downstream channels. Its structural role is distinct
from whether its values are available to the forecaster.

\paragraph{Covariate Family.}
Before imposing availability, let $v_{j,t}\in\reals$ denote the value of
channel $j$ at time $t$. The first $K$ channels are the covariates and
channel $K+1$ is the target. We write
\begin{equation}
v_{j,t}=g_{j,t}\,\phi_j\!\left(
 b_{j,t}+\sum_{i\in\operatorname{pa}(j)}
 \sum_{\ell=0}^{L_{\max}}h_{ji,\ell}
 f_{ji}(v_{i,t-\ell})\right)+\eta_{j,t}.
\label{eq:adapter-unified-generator}
\end{equation}
The baseline processes $b_{j,t}$ contain private dynamics or combinations of
shared sources. An acyclic parent graph specifies explicit channel effects.
Finite-support kernels $h_{ji,\ell}$ and value maps $f_{ji}$ determine their
temporal and nonlinear structure. Channel transformations $\phi_j$ define
different views of the resulting signal, while multipliers $g_{j,t}$ represent
activity gates or smooth amplitude modulation. The generation prefix of
$L_{\max}$ steps bounds the total lag along any graph path.
Additive perturbations $\eta_{j,t}$ act after modulation.

The parent-effect sum is only one source of dependence. Channels can also
share baseline sources or depend on common or channel-derived modulation
signals, even when this sum is absent.

\paragraph{Baseline Sources.}
Let $\mathcal{D}_{\mathrm{base}}$ denote the adapter recipe's fixed
$B_{\mathrm{base}}$-component instance of the univariate mixture in
Section~\ref{sec:synthetic-generator}.
Each source draw selects a component and its parameters, then generates a
fresh trajectory over the complete window, including the delay prefix.
Controls retain this draw in its original units. In other source-based families,
the draw is standardized using historical statistics and clipped to
$[-b_{\mathrm{clip}},b_{\mathrm{clip}}]$ before dependency construction.
It supplies the private target source in the
DAG, event-gating, and amplitude-modulation families, or the common source in
functional-view episodes.

Auxiliary sources are sampled from continuous processes, persistent levels,
binary events, counts, periodic signals, and categorical indicators. Each
eligible driver is replaced by an independent draw from
$\mathcal{D}_{\mathrm{base}}$ with probability $p_{\mathrm{base}}$.
Dedicated event drivers and stationary-isotropic latents retain their
specialized distributions.
Latent-linear baselines combine these shared and private drivers.
Cointegrated baselines use shared random-walk trends and stationary
residuals. Thus, $b_{j,t}$ can be a source trajectory, a combination of sources, or a
structurally generated trend process, according to the selected family.

\paragraph{Family Specializations.}
\emph{Control} episodes use independent baselines, identity channel maps,
unit multipliers, and no parent effects. With probability $p_{\mathrm{empty}}$,
they omit covariates. The remaining controls supply independent auxiliary
trajectories. \emph{Latent-linear} episodes
instead form baselines from linear combinations of shared and private sources,
varying dense, sparse, and block loading structures. \emph{Functional-view}
episodes apply different $\phi_j$ to a common baseline, with independent
noise on auxiliary observations. \emph{Cointegrated} episodes construct
baselines from shared low-rank random-walk trends and stationary
autoregressive residuals.

\emph{Delayed-linear} episodes use linear parent effects with sampled lags
and averaging kernels. \emph{Nonlinear structural} episodes additionally
sample nonlinear edge maps and optional parent-dependent activity gates.
\emph{Event-gating} episodes multiply a background target by up to
$M_{\mathrm{evt}}$ binary gates. With probability $p_{\mathrm{flag}}$,
covariates expose the realized gates, and otherwise they expose their
Bernoulli probabilities. Gates can be
contemporaneous or delayed, with optional post-gate noise allowing inexact
closures. \emph{Amplitude-modulation} episodes use a smooth positive multiplier,
\begin{equation}
g_{K+1,t}=\exp\!\left(\gamma\sum_{j=1}^{M}w_j\tanh(v_{j,t})\right),
\qquad \sum_{j=1}^{M}|w_j|=1,
\label{eq:adapter-amplitude}
\end{equation}
where $M$ is the nominal number of coupled covariates and $\gamma$ sets
the modulation strength. Table~\ref{tab:adapter-covariate-families} gives
the settings and the weights $\vect{\pi}_{\mathrm{rel}}$ in the order
control, latent-linear, delayed-linear, nonlinear structural, functional view,
cointegration, event gating, and amplitude modulation.

\paragraph{Graph and Observation Sampling.}
The delayed-linear and nonlinear structural families instantiate the
parent-effect sum with a topologically ordered graph, sampled edge
delays, and finite averaging kernels. The nonlinear family can add
monotone atomic edge maps and parent-dependent gates. The exact
sampling distributions and generation-prefix constraints appear in
Table~\ref{tab:adapter-dag-configuration}.

\paragraph{Observation Process.}
Remaining channels are independent distractors. Covariate order is randomly
permuted, and non-event channels receive random signs, scales, and offsets.
Event flags and probabilities retain their literal units. The observation
regimes in Table~\ref{tab:adapter-observation-configuration} are sampled after
the complete trajectories have been constructed. Sample-and-hold and
discretization modify covariate measurements. The availability regime
determines which entries of $\mat{X}$ are set to $\infty$.

Episodes are generated online from deterministic seed, training-step, and
sample identifiers, without persistent trajectory replay. The synthetic adapter
is trained for 250,000 updates on this distribution.

\paragraph{Adapter Graph and Observation Configuration.}
Table~\ref{tab:adapter-dag-configuration} gives the source mixture, graph
sampling laws, and observation settings. Historical gaps and unavailable
future covariate entries are encoded as $\infty$.

\begin{table}[!htbp]
\centering
\caption{Multivariate generator settings. Integer ranges are sampled uniformly.
Proposed averaging widths are truncated to respect context and prefix limits.
Availability and measurement probabilities follow the listed order.}
\label{tab:adapter-covariate-families}
\label{tab:adapter-dag-configuration}
\label{tab:adapter-observation-configuration}
\begingroup\footnotesize
\setlength{\tabcolsep}{4pt}
\renewcommand{\arraystretch}{1.0}
\reporttablestyle
\begin{tabularx}{\linewidth}{@{}l>{\raggedright\arraybackslash}Xr@{}}
\toprule
\reporttablehead{Symbol} & \reporttablehead{Setting} & \reporttablehead{Default or sampling law}\\
\midrule
\reporttablegroup{3}{Sources and relationships}\\
$B_{\mathrm{base}}$ & Univariate components used for baseline sources & $12$\\
$\vect{\pi}_{\mathrm{rel}}$ & Relationship weights (text order) &
$(0.10,0.20,0.15,0.15,0.10,0.10,0.12,0.08)$\\
$K$ & Positive covariate count & $\{1,2,4,8,16\}$ with weights $(3,3,4,4,5)/19$\\
$p_{\mathrm{empty}}$ & Probability of no covariates in controls & $0.5$\\
$M$ & Coupled covariate count outside controls &
$1{:}\min(4,K)$ with probability $0.8$, else $1{:}K$\\
$b_{\mathrm{clip}}$ & Standardized source clipping bound & $8$\\
$p_{\mathrm{base}}$ & Probability of replacing an eligible driver & $0.10$\\
$M_{\mathrm{evt}}$ & Maximum number of event gates & $3$\\
$p_{\mathrm{flag}}$ & Probability of exposing realized gates & $0.7$\\
$\gamma$ & Amplitude modulation strength & $\operatorname{Uniform}(0.2,\log 3)$\\
\midrule
\reporttablegroup{3}{Graph-based dependencies}\\
$|\operatorname{pa}(j)|$ & Parent count with $n_j$ eligible predecessors & $1{:}\min(3,n_j)$\\
$w_{ji}$ & Edge weight with independent equiprobable sign & Magnitudes from $\operatorname{Dirichlet}(1,\ldots,1)$\\
$L_{\max}$ & Generation prefix (steps) & $144$\\
$\ell_{ji}$ & Edge delay & $0{:}\min(128,\lfloor C/4\rfloor,L_{\max}-S_i)$\\
$L_{ji}$ & Proposed averaging width & $\{1,4,8,16\}$ with weights $(0.6,0.2,0.1,0.1)$\\
$\alpha_j$ & Coupling strength, intermediate and target &
$\operatorname{Uniform}(0.2,0.9)$, $\operatorname{Uniform}(0.15,0.95)$\\
$p_{\mathrm{atom}}$ & Probability of an atomic map in nonlinear episodes & $0.6$\\
$p_{\mathrm{gate}}$ & Activity gate probability per nonlinear non-root channel & $0.4$\\
\midrule
\reporttablegroup{3}{Observation process}\\
$\vect{p}_{\mathrm{avail}}$ & Known-future, mixed, past-only, absent & $(0.55,0.25,0.15,0.05)$\\
$(p_{\mathrm{keep}},p_{\mathrm{prefix}})$ & Keep a mixed future, then limit it to a prefix & $(0.5,0.25)$\\
$p_{\mathrm{gap}}$ & Historical gap probability per channel & $0.20$\\
$\vect{p}_{\mathrm{meas}}$ & Sample-and-hold, discretized, unchanged & $(0.1,0.1,0.8)$\\
\bottomrule
\end{tabularx}
\endgroup
\end{table}

\paragraph{DAG Sampling.}
For delayed-linear and nonlinear structural episodes, the parent graph spans
$M$ nominally coupled covariates and the target, ordered topologically with
the target last. For a non-root channel with private source $\bar b_{j,t}$,
Eq.~\eqref{eq:adapter-unified-generator} specializes to
$b_{j,t}=(1-\alpha_j)\bar b_{j,t}$, identity $\phi_j$, and kernels
\begin{equation}
h_{ji,\ell}=\frac{\alpha_jw_{ji}}{L_{ji}}\,
\ind\!\left\{\ell_{ji}\leq\ell<\ell_{ji}+L_{ji}\right\}.
\label{eq:adapter-graph-family}
\end{equation}
Linear episodes use identity $f_{ji}$. Nonlinear episodes choose atomic maps
with probability $p_{\mathrm{atom}}$ and elementary maps otherwise.
Each non-root channel can apply a sigmoid or threshold gate with probability
$p_{\mathrm{gate}}$. Its threshold and scale are fitted to the history of one
of the channel's parents.
The reported configuration constrains sampled atomic maps to be monotone;
signed effects and activity gates need not produce an overall monotone
relationship. Table~\ref{tab:adapter-dag-configuration} specifies the DAG
sampling distributions and the observation process.

Let $S_i$ denote the accumulated history requirement of parent $i$, with
$S_i=0$ for roots. Averaging widths are truncated to respect the context and
the generation prefix of $L_{\max}$ steps. The child requirement is
\begin{equation}
S_j=\max_{i\in\operatorname{pa}(j)}
\bigl(S_i+\ell_{ji}+L_{ji}-1\bigr).
\label{eq:adapter-support}
\end{equation}
The coupled-channel count is a structural budget: graph ancestry, coupling
strength, and observation availability determine the information ultimately
available for forecasting.

\section{Synthetic Pretraining and Producer Performance}
\label{app:training-lineage}
\label{sec:synthetic-training-lineage}

\begin{figure}[t]
\centering
\captionsetup[subfigure]{skip=3pt,justification=centering}
\begin{subfigure}[t]{.485\linewidth}
\centering
\includegraphics[width=\linewidth]{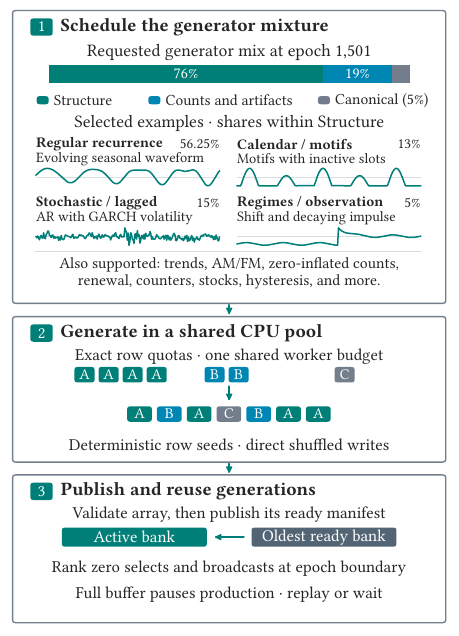}
\caption{Preset scheduling and online production.}
\label{fig:synthetic-pipeline-production}
\end{subfigure}\hfill
\begin{subfigure}[t]{.485\linewidth}
\centering
\includegraphics[width=\linewidth]{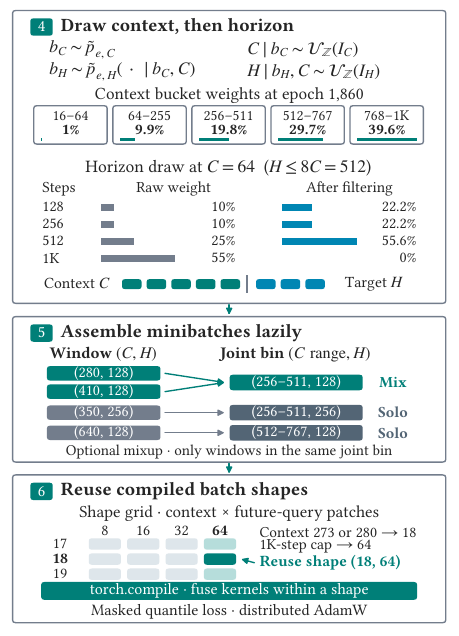}
\caption{Task sampling and distributed training.}
\label{fig:synthetic-pipeline-consumption}
\end{subfigure}
\caption{Online synthetic pretraining. (a) The requested mix at the
start of the horizon extension gives the preset shares for generated
rows. Four schematic examples show selected
Structure mechanisms, with recipe shares grouped by family.
Additional mechanisms appear in
Table~\ref{tab:generator-phenomena}. Largest-remainder allocation
sets exact component quotas. A shared CPU pool generates chunks and writes
rows directly into shuffled positions. (b) The trainer draws context
bucket $b_C$ and length $C$, then draws horizon bucket $b_H$ and length
$H$. At epoch $e$, $\tilde p_{e,C}$ and $\tilde p_{e,H}$ renormalize
the respective curriculum weights over feasible buckets. The horizon
weights come from the row for $b_C$ and are filtered using $C$.
$\mathcal U_{\mathbb Z}$ is uniform over the feasible integers in the
selected bucket interval $I_C$ or $I_H$. The worked example uses the
configured priors at cumulative epoch
1860. At $C=64$, the ratio constraint excludes $H=1\mathrm{K}$. The remaining
weights are divided by their total mass of 45\%. Feasible ranges also
respect the geometry caps and available span. The matching arrows show
which windows share a context and horizon bin. The 280- and 410-step
contexts can mix at $H=128$. The other two windows pass through as
singletons. Each cell of the shape grid represents context and
future-query patch counts at a fixed batch size. With 16-step patches,
273 and 280 context steps both round to 18 patches. The full-query
policy fixes the query count at 64, so both lengths can use the
highlighted shape $(18,64)$. Labels outside the sampled $H$ steps carry
$\infty$ and are excluded from the loss.
Gray columns illustrate shorter query widths. TorchInductor can fuse
eligible operations within a compiled graph and reuse it for repeated
shapes. Component tiles and example windows are schematic.}
\label{fig:synthetic-data-pipeline-v3}
\end{figure}

\paragraph{Generating and Sampling Training Windows.}
The producer allocates trajectories to mixture components by largest
remainder. Row and component seeds keep the output reproducible across
CPU workers. Workers write the trajectories to shuffled positions in a
memory-mapped bank while the GPU trains
(Figure~\ref{fig:synthetic-data-pipeline-v3}). At each epoch boundary,
the trainer loads the oldest ready replacement or, when replay is
enabled, keeps the active bank. New epoch seeds change the sampled
windows and augmentations. The loader draws context and horizon
lengths jointly under the feasibility constraints and trims padding
to the patch grid before transfer to the GPUs.

\begin{figure}[!htbp]
\centering
\includegraphics[width=.81\linewidth]{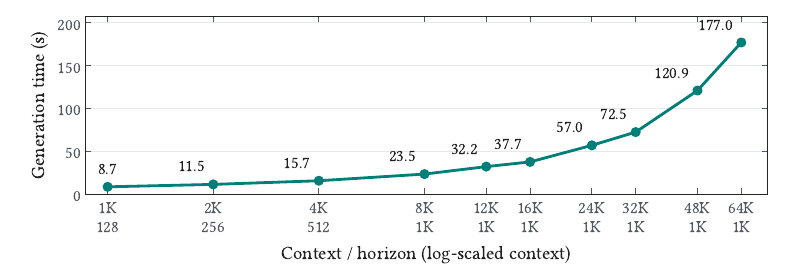}
\caption{Wall-clock generation time for a 100,000-series synthetic
bank. The 1K context with horizon 128 shows the median of three trials,
and the 64K context with horizon 1K shows the median of two. Each
intermediate configuration has one timed trial.}
\label{fig:synthetic-generation-time}
\end{figure}

\paragraph{Generation Time.}
We benchmark synthetic data generation from the final 24-component
mixture (Figure~\ref{fig:synthetic-generation-time}). After warm-up, we measure
how long it takes to generate, shuffle, and validate the data on a
single machine using 32 logical CPUs. With a 64K context and a 1K
forecast horizon, the 100,000-row bank occupies 25~GiB.

\begin{figure}[t]
\centering
\includegraphics[width=\linewidth]{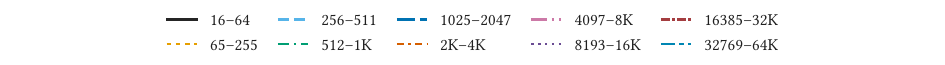}\par
\subcaptionbox{Learning rate\label{fig:curriculum-lr}}[.485\linewidth]{%
\includegraphics[width=\linewidth,trim=0 12bp 0 0,clip]{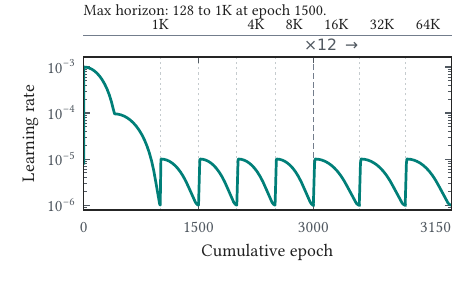}}\hfill
\subcaptionbox{Context probability\label{fig:curriculum-context}}[.485\linewidth]{%
\includegraphics[width=\linewidth,trim=0 12bp 0 0,clip]{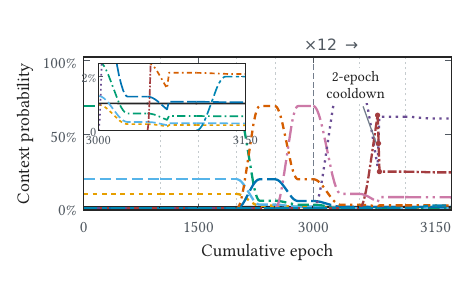}}\par\smallskip
\subcaptionbox{Supervised-horizon probability\label{fig:curriculum-horizon}}[.485\linewidth]{%
\includegraphics[width=\linewidth]{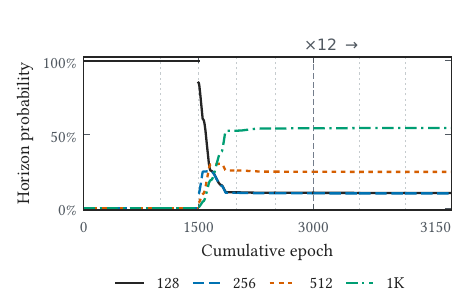}}\hfill
\subcaptionbox{Preset probability\label{fig:curriculum-preset}}[.485\linewidth]{%
\includegraphics[width=\linewidth]{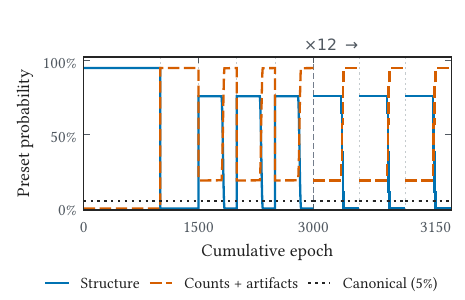}}
\caption{Configured optimization and sampling curriculum. Axes show actual cumulative epochs, with spacing enlarged $\times12$ after epoch 3000; the final three stages each span 50 epochs. Brackets in (a) show maximum context. (b) Context masses use common absolute length bins. Shape-preserving cubic curves interpolate the configured continuation-epoch values without changing those values. The inset magnifies continuation probabilities below 2.5\%, including the 64K stage's 2\% mass above 32K. (c) Horizon probabilities account for $H\leq8C$. (d) Nominal generator mixture, including the fixed 5\% canonical bank. These are configured priors before mixup, rather than empirical sampling frequencies.}
\label{fig:synthetic-training-curriculum}
\end{figure}

\paragraph{Backbone Curriculum.}
We start with a 1K-step context cap and a 128-step horizon. The first
1,000 epochs emphasize temporal structure, and the next 500 emphasize
counts and observation artifacts. With the horizon cap set to 1K, we
then run three 500-epoch stages at context caps of 1K, 4K, and 8K,
followed by three 50-epoch stages at 16K, 32K, and 64K. During the
extensions, the sampler retains short contexts and constrains supervised
horizons to $H\leq 8C$. We use AdamW~\citep{loshchilov2019adamw} with
stage-specific learning rates. Figure~\ref{fig:synthetic-training-curriculum}
shows the learning-rate, context, horizon, and mixture schedules.
The synthetic-only benchmark comparison in
Figure~\ref{fig:synthetic-normalized-mase-8k} uses the \textsc{Tafsut}-Synth checkpoint used for the synthetic-only benchmark comparison

\paragraph{Learning-Rate Schedule.}
After warmup, the base learning rate decays from $10^{-3}$ to $10^{-6}$.
Later stages ramp from $10^{-6}$ to $10^{-5}$, then decay back to $10^{-6}$
with a cosine schedule. The ramp lasts 10 epochs for the 500-epoch
adaptation, 20 for each 500-epoch extension, and 2 for each 50-epoch
continuation. The learning-rate curve shows all 123 optimizer updates per
epoch. Both figures expand the epoch axis by a factor of 12 after epoch
3000, marked by a dashed line and $\times12$.

\paragraph{Context and Supervised-Horizon Sampling.}
The initial context distribution remains fixed during the horizon extension. The 4K and 8K context extensions gradually introduce longer contexts over 300 epochs using a half-cosine transition, while the 16K and 32K extensions use a 30-epoch transition. The shortest context range (16--64) retains a sampling probability of 1\% in the original schedule. The final tiny, local, mid, long, and global bucket probabilities are 1\%, 4.95\%, 9.9\%, 29.7\%, and 54.45\%, respectively. After epoch 3070, the 32K extension returns to the context mixture of epoch 3061 over two epochs and retains it through epoch 3100. The 64K extension introduces 2\% sampling probability for contexts above 32K over 30 epochs, while preserving the 1\% tiny component. The figures aggregate context probabilities into common, non-overlapping absolute ranges.

Base training supervises 128 future steps. During the horizon extension, the sampling weights gradually reach 10\%, 10\%, 25\%, and 55\% for horizons of 128, 256, 512, and 1K, respectively, by local epoch 360. These weights remain fixed throughout all context extensions. The effective probabilities account for the constraint $H\leq 8C$ by excluding infeasible horizons and normalizing the remaining weights. The Transformer receives all 1K future query positions throughout the six extensions, with loss targets outside the sampled horizon set to $\infty$.

\paragraph{Generator Preset Curriculum.}
The canonical bank contributes 5\% throughout. Of the remaining 95\%, base
training uses the structure-focused preset and the initial 500-epoch adaptation
uses the counts-and-artifacts preset. Each subsequent extension begins with an
80:20 structure/adaptation mixture within that noncanonical share, corresponding
to 76\% structure, 19\% counts and artifacts, and 5\% canonical overall.
Local epochs 300--340 smoothly shift the noncanonical share to the adaptation
preset, ending at 95\% adaptation and 5\% canonical. These curves describe the
requested generator mixture. Asynchronous production and replay can delay
when a changed preset reaches the active training buffer.
In each 50-epoch continuation the corresponding preset transition occurs
over local epochs 30--34.

\paragraph{Adapter Pretraining.}
We freeze the backbone from the 32K-step context stage and train the
adapter for 250,000 updates on the synthetic covariate mixture.
Section~\ref{sec:adaptation} describes the subsequent FEV-Bench
adaptation.

\paragraph{Reported Validation Scores.}
\label{sec:synthetic-stage-performance}
\label{sec:synthetic-id-validation}

Table~\ref{tab:synthetic-id-validation} compares stage checkpoints with
Chronos-2 Synth on FEV-Bench-S, GIFT-Eval-S, and the final-mixture bank.
Epoch 1000 marks the end of base training. Each row reports a single
checkpoint, with stars marking the lower score before rounding.

The final checkpoint uses a maximum context length of 32K across all evaluations, while earlier checkpoints use their respective stage limits. Differences in training budget, generator composition, context length, and prediction horizon prevent isolating individual effects.

The final-mixture bank contains 16K fresh series from the 24 active components of the final counts-and-artifacts preset, including its 5\% canonical share and native observation effects. All models are evaluated on identical series with 64K context and 1K target positions, using the same past-only lag-one scales computed over the full context. Losses are averaged within each component, and component log scores are weighted by the final generator probabilities. This evaluates the final generator distribution without reproducing the online training and sampling mechanisms.

\begin{table}[!htbp]
\centering
\caption{Stage-checkpoint comparison on FEV-Bench-S, GIFT-Eval-S,
and the final-mixture bank. Trained $C$ is the \textsc{Tafsut} checkpoint's
training context cap. Visible $C/H$ gives the evaluated context and horizon.
The 64K-trained checkpoint uses 32K input. FEV-Bench-S and GIFT-Eval-S use stage-specific seasonal geometric scores
through epoch 3100. The epoch-3150 32K crop retains its paired full-64K
development population and past-only seasonal scales. GIFT-Eval-S MASE
is reported before normalization to Seasonal Naive.
Final-mixture scores use 16K paired series across 24 components, native observation
effects, fixed full-64K past-only lag-one scales and final-mixture geometric
aggregation. Rows with a 128-step horizon score 128 synthetic steps; the other rows score 1K.
Chronos-2 Synth uses at most 8K context; its measured 8K references repeat
for longer-context stages. Both models use matched synthetic series,
targets, scales, and eligibility. Stars mark the lower score within each
row pair.}
\label{tab:synthetic-id-validation}\footnotesize
\setlength{\tabcolsep}{1pt}
\setlength{\aboverulesep}{0.3ex}
\setlength{\belowrulesep}{0.3ex}
\reporttablestyle[3]
\begin{tabular}{@{}rrrl*{6}{r}@{}}
\toprule
& & & & \multicolumn{2}{c}{\reporttablehead{FEV-Bench-S}} &
\multicolumn{2}{c}{\reporttablehead{GIFT-Eval-S}} &
\multicolumn{2}{c}{\reporttablehead{\shortstack{Final-mixture\\bank}}} \\
\cmidrule(lr){5-6}\cmidrule(lr){7-8}\cmidrule(l){9-10}
\reporttablehead{\shortstack{Trained\\$C$}} &
\reporttablehead{\shortstack{Visible\\$C/H$}} &
\reporttablehead{Epoch} & \reporttablehead{Model} &
\reporttablehead{MASE} & \reporttablehead{SQL} &
\reporttablehead{MASE} & \reporttablehead{SQL} &
\reporttablehead{MASE} & \reporttablehead{SQL} \\
\midrule
1K & 1K/128 & 1000 & \textsc{Tafsut} & \textbf{$\star$ 1.1863} & 0.8172 & 1.0240 & 0.6967 & 6.4366 & 4.7547 \\
-- & 1K/128 & -- & Chronos-2 Synth & 1.1908 & \textbf{$\star$ 0.8148} & \textbf{$\star$ 1.0167} & \textbf{$\star$ 0.6947} & \textbf{$\star$ 6.3102} & \textbf{$\star$ 4.6668} \\
\midrule
1K & 1K/128 & 1500 & \textsc{Tafsut} & \textbf{$\star$ 1.1763} & \textbf{$\star$ 0.8077} & 1.0167 & \textbf{$\star$ 0.6914} & 6.3406 & 4.7174 \\
-- & 1K/128 & -- & Chronos-2 Synth & 1.1908 & 0.8148 & \textbf{$\star$ 1.0167} & 0.6947 & \textbf{$\star$ 6.3102} & \textbf{$\star$ 4.6668} \\
\midrule
1K & 1K/1K & 2000 & \textsc{Tafsut} & \textbf{$\star$ 1.2186} & \textbf{$\star$ 0.8404} & 1.1817 & \textbf{$\star$ 0.8138} & \textbf{$\star$ 17.5860} & \textbf{$\star$ 12.8571} \\
-- & 1K/1K & -- & Chronos-2 Synth & 1.2273 & 0.8440 & \textbf{$\star$ 1.1747} & 0.8180 & 17.6738 & 15.1535 \\
\midrule
4K & 4K/1K & 2500 & \textsc{Tafsut} & \textbf{$\star$ 1.1231} & \textbf{$\star$ 0.7714} & 1.0928 & \textbf{$\star$ 0.7469} & \textbf{$\star$ 15.1083} & \textbf{$\star$ 11.2238} \\
-- & 4K/1K & -- & Chronos-2 Synth & 1.1284 & 0.7722 & \textbf{$\star$ 1.0879} & 0.7485 & 16.7204 & 12.7797 \\
\midrule
8K & 8K/1K & 3000 & \textsc{Tafsut} & \textbf{$\star$ 1.1021} & \textbf{$\star$ 0.7565} & \textbf{$\star$ 1.0261} & \textbf{$\star$ 0.6974} & \textbf{$\star$ 13.0929} & \textbf{$\star$ 9.8649} \\
-- & 8K/1K & -- & Chronos-2 Synth & 1.1125 & 0.7615 & 1.0374 & 0.7121 & 16.1718 & 12.0925 \\
\midrule
16K & 16K/1K & 3050 & \textsc{Tafsut} & \textbf{$\star$ 1.0967} & \textbf{$\star$ 0.7525} & \textbf{$\star$ 1.0082} & \textbf{$\star$ 0.6873} & \textbf{$\star$ 11.8404} & \textbf{$\star$ 8.9705} \\
-- & 8K/1K & -- & Chronos-2 Synth & 1.1125 & 0.7615 & 1.0374 & 0.7121 & 16.1718 & 12.0925 \\
\midrule
32K & 32K/1K & 3100 & \textsc{Tafsut} & \textbf{$\star$ 1.0980} & \textbf{$\star$ 0.7537} & \textbf{$\star$ 1.0035} & \textbf{$\star$ 0.6836} & \textbf{$\star$ 10.9908} & \textbf{$\star$ 8.6131} \\
-- & 8K/1K & -- & Chronos-2 Synth & 1.1125 & 0.7615 & 1.0374 & 0.7121 & 16.1718 & 12.0925 \\
\midrule
\textbf{64K} & \textbf{32K/1K} & 3150 & \textsc{Tafsut} (capped at 32K) & \textbf{$\star$ 1.0961} & \textbf{$\star$ 0.7522} & \textbf{$\star$ 0.9978} & \textbf{$\star$ 0.6789} & \textbf{$\star$ 10.0367} & \textbf{$\star$ 7.8515} \\
-- & 8K/1K & -- & Chronos-2 Synth & 1.1125 & 0.7615 & 1.0374 & 0.7121 & 16.1718 & 12.0925 \\
\bottomrule
\end{tabular}
\end{table}

\paragraph{Observed Producer-Buffer Freshness.}
\label{sec:synthetic-buffer-freshness}

\begin{figure}[t]
\centering
\subcaptionbox{New-generation epoch fraction\label{fig:freshness-fraction}}[.485\linewidth]{%
\includegraphics[width=\linewidth]{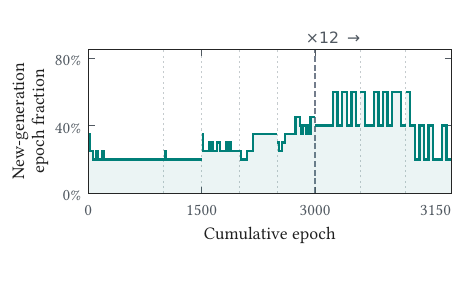}}\hfill
\subcaptionbox{Consecutive replay epochs\label{fig:freshness-replay}}[.485\linewidth]{%
\includegraphics[width=\linewidth]{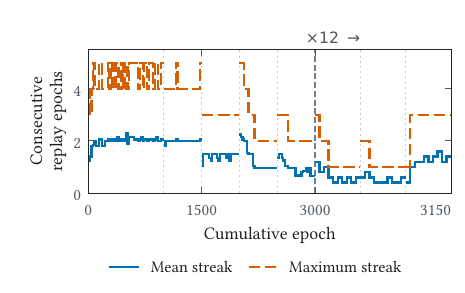}}
\caption{Observed buffer freshness across the synthetic training lineage. (a) Fraction of epochs loading a new one-million-series generation. (b) Mean and maximum consecutive replay counts, with zero on fresh epochs. Values summarize complete non-overlapping windows within each stage: 20 epochs before epoch 3000 and five epochs thereafter. The continuous axis enlarges epoch spacing $\times12$ after 3000; all three continuation stages contain 50 completed epochs, through 3150. Logged initialization and resume flags are retained. Solid and dashed lines distinguish the replay summaries in grayscale.}
\label{fig:synthetic-buffer-freshness}
\end{figure}

Each epoch samples one million synthetic series from the active generation. At epoch boundaries, the oldest unseen ready generation replaces the active one if available; otherwise, the active generation is replayed. The buffer holds one active generation and one ready replacement.

Freshness measures how often the active generation is replaced, rather than the age or novelty of individual series. Initial loads after resuming training are counted as neither fresh nor replayed, and only the last occurrence of duplicate epoch records is retained.

The observed fresh fractions are 20.7\%, 20.2\%, 26.6\%, 30.6\%, and 36.8\% for base training, adaptation, horizon extension, and the 4K and 8K context extensions, respectively. The 16K, 32K, and 64K continuations achieve fresh fractions of 46.0\%, 50.0\%, and 32.0\%, respectively. The maximum consecutive replay count ranges from two to five epochs across all stages.

Freshness depends on both generation throughput and training consumption time. It does not measure the delay between curriculum changes and their appearance in the active generation. Figure~\ref{fig:synthetic-training-curriculum} therefore describes the configured curriculum, while Figure~\ref{fig:synthetic-buffer-freshness} shows the observed generation replacement frequency.

\section{Representation Analyses}
\label{app:representation-analyses}

\begin{figure}[!htbp]
\centering
\captionsetup{skip=3pt}
\includegraphics[width=.90\linewidth]{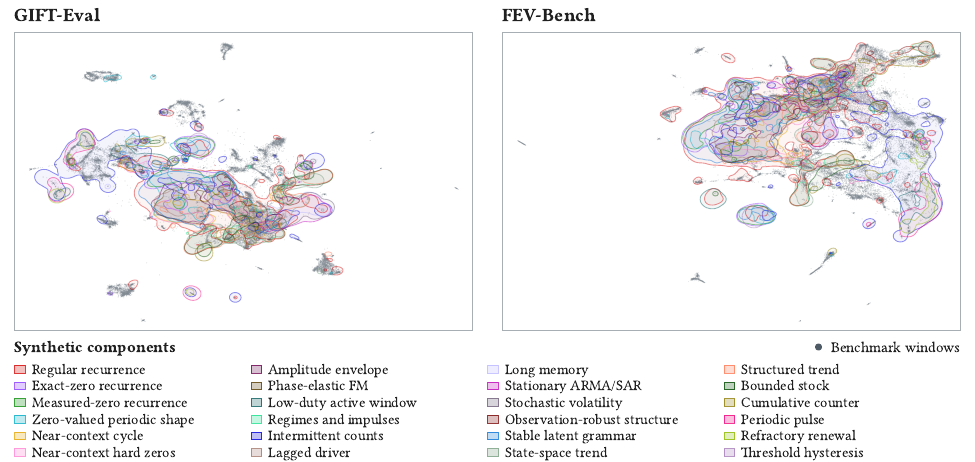}
\caption{Representation coverage on GIFT-Eval-S and FEV-Bench-S.
Dark points mark benchmark windows. Colored contours show the smoothed
densities of the 24 synthetic generator components at a threshold of 2.5
kernel-weighted points. Each panel uses a separate UMAP projection.}
\label{fig:generator-benchmark-coverage}
\end{figure}

\paragraph{Synthetic Generator Coverage.}
Figure~\ref{fig:generator-benchmark-coverage} compares the final generator
mixture with 32,685 windows from GIFT-Eval-S and 32,714 from FEV-Bench-S.
For each benchmark, we draw an equally sized synthetic sample with a
matched distribution of visible history spans. We average the frozen
\textsc{Tafsut-Synth} encoder states over valid historical patches, with the
context capped at 8K steps. Each benchmark uses a joint UMAP of its
real and synthetic representations after $\ell_2$ normalization and
48-component PCA. Synthetic density overlaps many benchmark regions in
the projection, and some benchmark clusters extend beyond the contours.

\begin{figure}[!htbp]
\centering
\captionsetup{skip=3pt}
\includegraphics[width=0.54\linewidth]{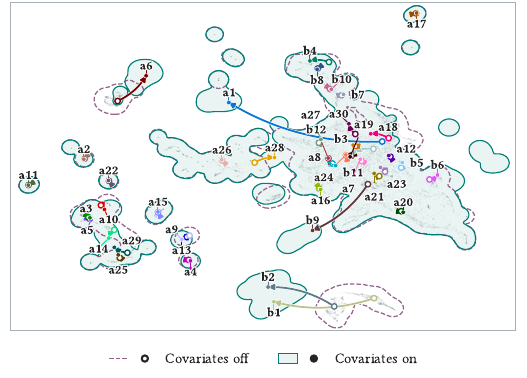}
\caption{Paired UMAP of adapter states on FEV-Bench-S without (hollow endpoints, dashed support)
 and with (filled endpoints, shaded
support) historical covariates. Each curved arrow joins one observed window pair nearest its
task's mean in the joint off/on projection. Labels give the panel and rank
in Figure~\ref{fig:fev-covariate-effect}, with exact task IDs in
Table~\ref{tab:fev-covariate-ranks}.}
\label{fig:fev-covariate-umap}
\end{figure}

\paragraph{Covariate Sensitivity on FEV-Bench-S.}
\phantomsection
\label{app:fev-covariate-representations}
Figure~\ref{fig:fev-covariate-umap} compares adapter representations with
historical covariates masked and available on 10,482 paired windows from
FEV-Bench-S, covering the 42 tasks in Figure~\ref{fig:fev-covariate-effect}.
We fix the last synthetic checkpoint, the covariate adapter, and the
target history, and mask future covariates in both passes. The readout is
the mean adapter forecast-query state before
the residual head. We fit one UMAP to the paired states after $\ell_2$
normalization and 48-component PCA. This probes historical-covariate
sensitivity under a different checkpoint and evaluation population from
the \textsc{Tafsut}-FEV covariate comparison in Figure~\ref{fig:fev-covariate-effect}.

A large projected shift can accompany a large forecasting gain. For weekly
Rossmann (\texttt{a1}), the arrow spans distinct regions and the adapter
reduces MASE by 51.6\% in Figure~\ref{fig:fev-covariate-effect}. We could observe the same for past-only covariates like KDD 10T (\texttt{b1}) and KDD 30T (\texttt{b2}).

\section{Adaptation Configurations and Source Inventories}
\label{app:adaptation-corpora}

\paragraph{GIFT-Eval Source Inventory.}
The selection in Table~\ref{tab:adaptation-source-selection} records
retained, filtered, and excluded GIFT-Eval source configurations.

\begin{table}[!htbp]
\centering
\captionsetup{font=footnotesize,skip=3pt}
\caption{GIFT-Eval-targeted real-data corpus selection. $+$ retained; $\triangle$ retained after filtering; $\times$ excluded entirely. Datasets remain under their original collection even when replaced: $\rightarrow$ selects the corresponding copy from C (Chronos pretrain) or P (GIFT-Eval-Pretrain), at the indicated frequency unless noted. D/W/H/M/Q/Y denote daily/weekly/hourly/monthly/quarterly/yearly; named year ranges group variants.}
\label{tab:adaptation-source-selection}
\label{tab:adaptation-replacements}
\begingroup
\definecolor{adteal}{HTML}{007F78}
\definecolor{adcyan}{HTML}{0086B3}
\definecolor{adslate}{HTML}{737D8C}
\newcommand{\adkeep}{\textcolor{adteal}{$\boldsymbol{+}$}}
\newcommand{\adreplace}{\textcolor{adcyan}{$\boldsymbol{\rightarrow}$}}
\newcommand{\adexclude}{\textcolor{adslate}{$\boldsymbol{\times}$}}
\newcommand{\adfilter}{\textcolor{adteal}{$\boldsymbol{\triangle}$}}
\fontsize{8}{9.4}\selectfont
\setlength{\tabcolsep}{4pt}
\renewcommand{\arraystretch}{1.0}
\reporttablestyle
\begin{tabularx}{\linewidth}{@{}*{4}{>{\raggedright\arraybackslash}X}@{}}
\toprule
\reporttablegroup{4}{GIFT-Eval-Pretrain: original datasets, including replaced copies} \\
\midrule
\adkeep\enspace Beijing Subway & \adkeep\enspace HZMetro & \adkeep\enspace SHMetro & \adkeep\enspace LOS\_LOOP \\
\adkeep\enspace PEMS03 & \adkeep\enspace PEMS04 & \adkeep\enspace PEMS07 & \adkeep\enspace PEMS08 \\
\adkeep\enspace PEMS-BAY & \adkeep\enspace Q-Traffic & \adkeep\enspace LargeST 2017--21 & \adkeep\enspace Alibaba 2018 \\
\adkeep\enspace Azure VM 2017 & \adkeep\enspace Borg 2011 & \adkeep\enspace BDG-2 bear & \adkeep\enspace BDG-2 fox \\
\adkeep\enspace BDG-2 panther & \adkeep\enspace BDG-2 rat & \adkeep\enspace Buildings900K & \adkeep\enspace Borealis \\
\adkeep\enspace Bull & \adkeep\enspace Cockatoo & \adkeep\enspace Hog & \adkeep\enspace Sceaux \\
\adkeep\enspace Beijing air quality & \adkeep\enspace China air quality & \adkeep\enspace Oikolab & \adkeep\enspace Subseasonal temperature \\
\adkeep\enspace Subseasonal precipitation & \adkeep\enspace CMIP6 (1850--2010) & \adkeep\enspace ERA5 (1989--2018) & \adkeep\enspace CDC ILINet \\
\adkeep\enspace CDC NREVSS & \adkeep\enspace Project Tycho & \adkeep\enspace COVID mobility & \adkeep\enspace COVID energy \\
\adkeep\enspace Favorita sales & \adkeep\enspace Favorita transactions & \adkeep\enspace GoDaddy & \adkeep\enspace Bitcoin \\
\adkeep\enspace GEFCom 2012 & \adkeep\enspace GEFCom 2014 & \adkeep\enspace GEFCom 2017 & \adkeep\enspace ELF \\
\adkeep\enspace IDEAL & \adkeep\enspace LCL & \adkeep\enspace PDB & \adkeep\enspace SMART \\
\adkeep\enspace Spain & \adkeep\enspace Residential load & \adkeep\enspace Residential PV & \adkeep\enspace Solar power \\
\adkeep\enspace wind power & \adkeep\enspace Wind Farms & \adkeep\enspace KDD2022 & \adkeep\enspace Extended web traffic \\
\adkeep\enspace Wiki-Rolling & \adkeep\enspace vehicle trips & \adkeep\enspace sunspots & \adkeep\enspace M3 quarterly \\
\adkeep\enspace M3 yearly & \adkeep\enspace M3 other & \adkeep\enspace mixed weekly traffic &  \\
\addlinespace[3pt]
Australian elec. \adreplace\ C\textsuperscript{g} & ElecDemand \adreplace\ C\textsuperscript{g} & CIF 2016 (6/12) \adreplace\ C & M1 (M/Q/Y) \adreplace\ C \\
M3 monthly \adreplace\ C & FRED-MD \adreplace\ C & London meters \adreplace\ C & Pedestrian \adreplace\ C \\
M5 \adreplace\ C\textsuperscript{b} & NN5 (D/W) \adreplace\ C & Rideshare \adreplace\ C & Taxi 30-min \adreplace\ C \\
Uber (D/H) \adreplace\ C & Tourism (M/Q/Y) \adreplace\ C\textsuperscript{e} & Traffic hourly \adreplace\ C & Web weekly \adreplace\ P\textsuperscript{a} \\
Weather \adreplace\ C/P\textsuperscript{h} &  &  &  \\
\midrule
\reporttablegroup{4}{Additional Chronos sources (including the companion collection)} \\
\midrule
\adkeep\enspace Taxi hourly & \adkeep\enspace USHCN daily & \adkeep\enspace WeatherBench daily & \adkeep\enspace WeatherBench weekly \\
\adkeep\enspace WeatherBench hourly\textsuperscript{c} & \adkeep\enspace Dominick & \adkeep\enspace ERCOT & \adkeep\enspace Exchange rate \\
\adkeep\enspace Mexico City bikes & \adkeep\enspace Wiki daily 100K & \adkeep\enspace Wind Farms daily & \adkeep\enspace Wind Farms hourly \\
\adfilter\enspace Solar native\textsuperscript{d} & \adfilter\enspace Solar hourly\textsuperscript{d} & \multicolumn{2}{l@{}}{\adkeep\enspace Spanish energy and weather\textsuperscript{f}} \\
\multicolumn{2}{@{}l}{\adkeep\enspace Brazilian cities temperature\textsuperscript{f}} & & \\
\bottomrule
\end{tabularx}
\par\vspace{2pt}
\reporttablestyle
\begin{tabularx}{\linewidth}{@{}*{4}{>{\raggedright\arraybackslash}X}@{}}
\toprule
\reporttablegroup{4}{Excluded entirely: no GIFT-Pretrain or Chronos copy is used} \\
\midrule
\adexclude\enspace Car parts & \adexclude\enspace COVID deaths & \adexclude\enspace M4 (all frequencies) & \adexclude\enspace Saugeen \\
\adexclude\enspace US births & \adexclude\enspace UCI electricity & \adexclude\enspace Hospital & \adexclude\enspace KDD Cup 2018 \\
\adexclude\enspace Temperature/rain & & & \\
\bottomrule
\end{tabularx}
\par\vspace{3pt}
\begin{minipage}{\linewidth}
\textsuperscript{a}Web weekly maps to P/\texttt{traffic\_weekly}, retaining the web series plus 862 road series.
\textsuperscript{b}Chronos M5 adds 56 days.
\textsuperscript{c}WeatherBench hourly groups the 14 stored variables; CMIP6 spans 1850--2010 in 5-year steps, ERA5 spans 1989--2018 annually, and LargeST spans 2017--2021.
\textsuperscript{d}Remove 137 Alabama series; retain 5,029 per resolution.
\textsuperscript{e}The historical corpus includes Chronos monthly tourism; it shares histories with GIFT-Eval M4 monthly. Checked shared intervals do not intersect forecast targets.
\textsuperscript{f}From the Chronos companion collection. \textsuperscript{g}Australian electricity and ElecDemand both map to C/\texttt{monash\_australian\_electricity}. \textsuperscript{h}Weather maps to C/\texttt{monash\_weather} (daily) and P/\texttt{oikolab\_weather} (hourly).
\end{minipage}
\endgroup
\end{table}

\paragraph{FEV-Bench Source Inventory.}
We construct a separate real-data selection for FEV-Bench. Under its
zero-shot protocol~\citep{shchur2025fev}, exposure to any part of a benchmark dataset at the same frequency counts as overlap; truncating the training history does not remove that overlap. Resampled variants at different frequencies remain eligible. Following the benchmark-specific treatment in
Appendix~E of \citet{podest2026tirex2}, we retain the original subdaily solar datasets, rather than the Alabama-filtered GIFT-Eval copies.

The reviewed selection in Table~\ref{tab:fev-adaptation-source-selection} contains 2,980 dataset/member links: 138 GIFT-Eval-Pretrain, 48 Chronos, two Chronos companion sources, and 2,792 BOOM members. The 50 omitted source copies comprise 28 overlap/source exclusions and 22 alternate representations. Local full-target comparisons identify ERCOT, M5, hospital, ETTh, ETTm, and 15 BOOM members as matching FEV sources; this member-level BOOM list is our verified mapping, rather than a published TiRex-2 filter. One Oikolab copy is retained. Saugeen uses corrected daily timestamps with unchanged values, and COVID deaths retains all 266 original series, including \texttt{T000256}. The FEV-Bench backbone uses eligible subsets of this source inventory, with separate sampling rules (Section~\ref{sec:adaptation}).

\begin{table}[!htbp]
\centering
\captionsetup{font=footnotesize,skip=3pt}
\caption{Prepared FEV-Bench-targeted source selection (1 November 2026), using the compact format of Table~\ref{tab:adaptation-source-selection}. $+$ retained; $\rightarrow$ unused copy replaced by the corresponding copy in the indicated collection; $\times$ excluded; $\diamond$ timestamp/schema repair with unchanged target values. P = GIFT-Eval-Pretrain, C = Chronos pretrain, X = Chronos companion collection. Arrows preserve frequency unless noted. Source-link counts include year and variable configurations; they are not series counts.}
\label{tab:fev-adaptation-source-selection}
\begingroup
\definecolor{fevteal}{HTML}{007F78}
\definecolor{fevcyan}{HTML}{0086B3}
\definecolor{fevslate}{HTML}{737D8C}
\newcommand{\fevkeep}{\textcolor{fevteal}{$\boldsymbol{+}$}}
\newcommand{\fevarrow}{\textcolor{fevcyan}{$\boldsymbol{\rightarrow}$}}
\newcommand{\fevexclude}{\textcolor{fevslate}{$\boldsymbol{\times}$}}
\newcommand{\fevrepair}{\textcolor{fevteal}{$\boldsymbol{\diamond}$}}
\fontsize{7.5}{8.8}\selectfont
\setlength{\tabcolsep}{3pt}
\renewcommand{\arraystretch}{1.0}
\reporttablestyle
\begin{tabularx}{\linewidth}{@{}*{4}{>{\raggedright\arraybackslash}X}@{}}
\toprule
\reporttablegroup{4}{GIFT-Eval-Pretrain (P): 138 retained source links} \\
\midrule
\fevkeep\enspace Beijing Subway & \fevkeep\enspace HZMETRO & \fevkeep\enspace LOS\_LOOP & \fevkeep\enspace PEMS03 \\
\fevkeep\enspace PEMS04 & \fevkeep\enspace PEMS07 & \fevkeep\enspace PEMS08 & \fevkeep\enspace PEMS-BAY \\
\fevkeep\enspace Q-Traffic & \fevkeep\enspace SHMETRO & \fevkeep\enspace Alibaba 2018 & \fevkeep\enspace Australian elec. \\
\fevkeep\enspace Azure VM 2017 & \fevkeep\enspace BDG-2 bear & \fevkeep\enspace BDG-2 fox & \fevkeep\enspace BDG-2 panther \\
\fevkeep\enspace BDG-2 rat & \fevkeep\enspace Beijing air quality & \fevkeep\enspace Bitcoin & \fevkeep\enspace Borealis \\
\fevkeep\enspace Borg 2011 & \fevkeep\enspace Buildings900K & \fevkeep\enspace Bull & \fevkeep\enspace CDC ILINet \\
\fevkeep\enspace CDC NREVSS & \fevkeep\enspace China air quality & \fevkeep\enspace CIF 2016 (12) & \fevkeep\enspace CIF 2016 (6) \\
\fevkeep\enspace CMIP6 (1850--2010) & \fevkeep\enspace Cockatoo & \fevkeep\enspace COVID energy & \fevkeep\enspace COVID mobility \\
\fevkeep\enspace ElecDemand & \fevkeep\enspace ELF & \fevkeep\enspace ERA5 (1989--2018) & \fevkeep\enspace Extended web traffic \\
\fevkeep\enspace Favorita sales & \fevkeep\enspace GoDaddy & \fevkeep\enspace Hog & \fevkeep\enspace IDEAL \\
\fevkeep\enspace Web traffic weekly & \fevkeep\enspace LargeST (2017--21) & \fevkeep\enspace LCL & London meters\,\fevarrow\,C \\
\fevkeep\enspace M1 monthly & \fevkeep\enspace M1 quarterly & \fevkeep\enspace M1 yearly & \fevkeep\enspace M3 monthly \\
\fevkeep\enspace M3 other & \fevkeep\enspace M3 quarterly & \fevkeep\enspace M3 yearly & \fevkeep\enspace NN5 daily \\
\fevkeep\enspace NN5 weekly & \fevkeep\enspace Oikolab\textsuperscript{a} & \fevkeep\enspace PDB & Pedestrian\,\fevarrow\,C \\
\fevkeep\enspace Project Tycho & \fevkeep\enspace Residential load & \fevkeep\enspace Residential PV & \fevkeep\enspace Rideshare \\
\fevkeep\enspace Sceaux & \fevkeep\enspace SMART & \fevkeep\enspace Solar power & \fevkeep\enspace Spain \\
\fevkeep\enspace Subseasonal temp. & \fevkeep\enspace Subseasonal precip. & \fevkeep\enspace Sunspots & Taxi 30-min\,\fevarrow\,C \\
\fevkeep\enspace Tourism monthly & \fevkeep\enspace Tourism quarterly & \fevkeep\enspace Tourism yearly & Traffic hourly\,\fevarrow\,C \\
\fevkeep\enspace Mixed weekly traffic & Uber daily\,\fevarrow\,C & Uber hourly\,\fevarrow\,C & \fevkeep\enspace Vehicle trips \\
Weather\,\fevarrow\,C/P\textsuperscript{a} & \fevkeep\enspace Wiki-Rolling & \fevkeep\enspace Wind Farms (1-min) & \fevkeep\enspace Wind power \\
\midrule
\reporttablegroup{4}{Chronos pretrain (C): 48 retained source links} \\
\midrule
\fevkeep\enspace Dominick & \fevkeep\enspace Electricity 15-min & \fevkeep\enspace Exchange rate & \fevkeep\enspace M4 (D/H/M/Q/W/Y) \\
\fevkeep\enspace Mexico city bikes & Australian elec.\,\fevarrow\,P & \fevkeep\enspace Car parts & CIF 2016\,\fevarrow\,P(6/12) \\
\fevkeep\enspace COVID deaths\textsuperscript{b} & \fevkeep\enspace Electricity hourly & \fevkeep\enspace Electricity weekly & \fevkeep\enspace KDD Cup 2018 \\
\fevkeep\enspace London meters & M1 monthly\,\fevarrow\,P & M1 quarterly\,\fevarrow\,P & M1 yearly\,\fevarrow\,P \\
M3 monthly\,\fevarrow\,P & M3 quarterly\,\fevarrow\,P & M3 yearly\,\fevarrow\,P & NN5 weekly\,\fevarrow\,P \\
\fevkeep\enspace Pedestrian & \fevkeep\enspace Rideshare & \fevrepair\enspace Saugeen\textsuperscript{c} & \fevkeep\enspace Temperature/rain \\
Tourism monthly\,\fevarrow\,P & Tourism quarterly\,\fevarrow\,P & Tourism yearly\,\fevarrow\,P & \fevkeep\enspace Traffic hourly \\
\fevkeep\enspace Weather daily & NN5 daily\,\fevarrow\,P & \fevkeep\enspace Solar original\textsuperscript{d} & \fevkeep\enspace Solar hourly\textsuperscript{d} \\
Solar hourly safe\,\fevarrow\,C: original & Solar safe\,\fevarrow\,C: original & \fevkeep\enspace Taxi hourly & \fevkeep\enspace Taxi 30-min \\
\fevkeep\enspace Uber daily & \fevkeep\enspace Uber hourly & \fevkeep\enspace USHCN daily & \fevkeep\enspace WeatherBench daily \\
\fevkeep\enspace WeatherBench hourly\textsuperscript{e} & \fevkeep\enspace WeatherBench weekly & \fevkeep\enspace Wiki daily 100K & \fevkeep\enspace Wind Farms daily \\
\fevkeep\enspace Wind Farms hourly &  &  &  \\
\midrule
\reporttablegroup{4}{Chronos companion collection (X): 2 retained source links} \\
\midrule
\fevkeep\enspace Brazilian city temp. & \fevkeep\enspace Spanish energy/weather &  &  \\
\midrule
\reporttablegroup{4}{BOOM: 2,792 members retained; 15 verified FEV members excluded} \\
\midrule
\multicolumn{4}{@{}l}{\fevkeep\enspace All remaining BOOM members; excluded members (ID / frequency):} \\
\fevexclude\enspace 1062 / 5T & \fevexclude\enspace 1209 / 5T & \fevexclude\enspace 1225 / T & \fevexclude\enspace 1230 / 5T \\
\fevexclude\enspace 1282 / T & \fevexclude\enspace 1487 / 5T & \fevexclude\enspace 1631 / 30T & \fevexclude\enspace 1676 / 30T \\
\fevexclude\enspace 1855 / H & \fevexclude\enspace 1975 / H & \fevexclude\enspace 2187 / H & \fevexclude\enspace 285 / T \\
\fevexclude\enspace 619 / T & \fevexclude\enspace 772 / T & \fevexclude\enspace 963 / T &  \\
\midrule
\reporttablegroup{4}{Other overlap/source exclusions: each dataset family listed once} \\
\midrule
\fevexclude\enspace ERCOT (C) & \fevexclude\enspace M5 (C/P) & \fevexclude\enspace Hospital (C) & \fevexclude\enspace FRED-MD (C/P) \\
\fevexclude\enspace Favorita transactions (P) & \fevexclude\enspace GEFCom 2012 (P) & \fevexclude\enspace GEFCom 2014 (P) & \fevexclude\enspace GEFCom 2017 (P) \\
\fevexclude\enspace KDD2022 (P) & \fevexclude\enspace ETTh (X) & \fevexclude\enspace ETTm (X) &  \\
\bottomrule
\end{tabularx}
\par\vspace{3pt}
\begin{minipage}{\linewidth}
\textsuperscript{a}P/\texttt{weather} maps to C/\texttt{monash\_weather} (daily) plus the single P/\texttt{oikolab\_weather} copy (hourly).
\textsuperscript{b}All 266 original COVID-death series retained; no UK-series filter.
\textsuperscript{c}Saugeen: 23,741 unchanged values, daily timestamps restored for 1915--1979, and \texttt{T1} renamed to \texttt{target}.
\textsuperscript{d}Original subdaily solar copies retained; the local \texttt{\_safe} alternatives are unused.
\textsuperscript{e}WeatherBench hourly groups 14 variables. CMIP6 years advance in 5-year steps; ERA5 and LargeST annually. D/H/M/Q/W/Y denote daily/hourly/monthly/quarterly/weekly/yearly; BOOM T = one minute, 5T = five minutes, 30T = thirty minutes. Duplicate-copy omissions do not imply family-wide overlap. 
\end{minipage}
\endgroup
\end{table}

\section{Cross-Task Audit of In-Domain Adaptation Histories}
\label{app:gift-cross-task-audit}
\label{app:benchmark-cross-task-audit}
\begin{figure}[!htbp]
\centering
\includegraphics[width=\linewidth]{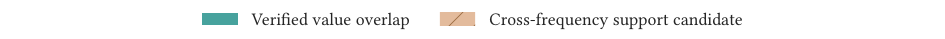}\par\vspace{2pt}
\subcaptionbox{GIFT-Eval: 136 pairs\label{fig:gift-overlap-family-flow}}[.49\linewidth]{%
\includegraphics[page=1,width=\linewidth]{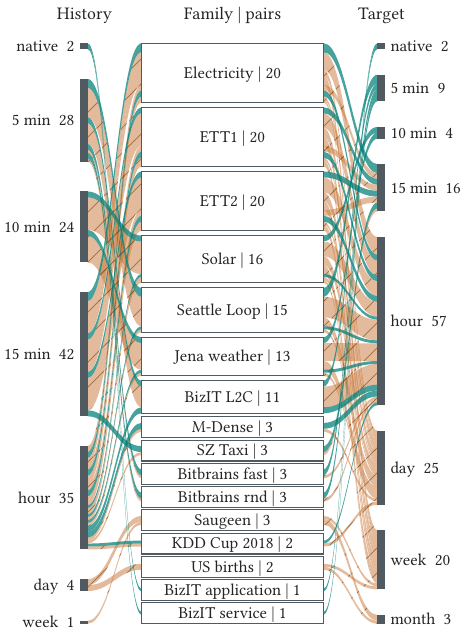}}\hfill
\subcaptionbox{FEV-Bench: 48 pairs\label{fig:fev-overlap-family-flow}}[.49\linewidth]{%
\includegraphics[page=2,width=\linewidth]{figures/benchmark_overlap/family-flow.pdf}}
\caption{Cross-task overlap traced from source-history frequency through
dataset family to target frequency. Ribbon widths scale with flagged
ordered-pair counts within each panel. The 16 affected GIFT families and
21 affected FEV families are labeled individually with their pair counts.
Dataset boxes have a minimum height for legibility. GIFT has 38 verified
same-frequency value overlaps (solid teal) and 98 cross-frequency
calendar-support candidates (hatched amber). All 48 FEV pairs are
candidates. GIFT uses official training
prefixes and FEV uses eligible first-origin contexts. Native denotes an
unsuffixed stored configuration.}
\label{fig:benchmark-overlap-family-flow}
\end{figure}

\paragraph{GIFT-Eval Histories.}
The audit catalogs all $97^2=9{,}409$ ordered task pairs, including
self-pairs. Of these, 521 share a declared dataset root; the other
8,888 lack a declared shared-series identity and are not certified
disjoint. No complete-series duplicate was found across roots under
the recorded frequency/start fingerprint, but arbitrary partial,
transformed, or renamed duplicates were not exhaustively searched.
The 292 prepared split files and all 97 training prefixes were verified
against original values. The audit checked 326,490 official rolling-test
windows and 40,670,379 target positions. These counts describe this
audit's official-window extraction, separately from the archived
evaluation population in Section~\ref{sec:experimental-setup}.

\paragraph{Verified Value Overlap and Cross-Frequency Overlap Candidates.}
Same-frequency conflicts use verified original values. Cross-frequency
candidates use intersecting calendar-period supports for aligned
root/item/channel identities with finite source observations. Different
aggregates need not have equal values, and upstream left/right aggregation
labeling was not independently reconstructed for every source. A numerical
follow-up verifies one solar case: summing six successive 10-minute
training values reconstructs 174,264 finite hourly long-horizon test
targets across 137 series within a maximum absolute discrepancy of
$3.052\times10^{-5}$. This does not certify all 98 candidates.

\paragraph{Verified Derived-History Scope.}
All 139,326 retained derived GIFT histories (160,730,282 values) match
original prefixes and pass the shared target-boundary check. This verification applies
only to the derived GIFT histories. The GIFT-Eval model reported in
Section~\ref{sec:native-benchmarks} uses the separate source corpus in
Table~\ref{tab:adaptation-source-selection}. Earlier test targets
appearing in a later rolling forecast's context are legitimate by themselves
and are not treated as training leakage.

\paragraph{Deduplicated Overlap Volume.}
The follow-up counts each series by dataset root, item identity, and
channel, deduplicating across forecast terms and frequency variants.
Target observations are keyed by stored dataset/frequency, item,
channel, and original position, with only finite values retained.
Thus the same target appearing in several task pairs is counted once;
different-frequency aggregates remain distinct observations rather than
a common-duration time grid. Verified value overlap affects 4,753 series
and 12,257,094 target observations. Cross-frequency overlap candidates
affect 4,711 series and 6,064,699 observations. Their union is 4,755
series and 13,121,651 observations; 4,709 series occur in both categories,
and 5,200,142 target observations are flagged by both. Electricity and
Seattle Loop account for 76\% of the union of affected target observations.
Candidate volume remains a temporal-support measure, not proof of
numerical identity.

\paragraph{FEV-Bench Histories.}
\label{app:fev-cross-task-audit}
The FEV audit catalogs all $100^2=10{,}000$ ordered pairs, including self-pairs.
It compares eligible contexts from each task's first official forecast origin
against the union of that task's and all other tasks' official rolling targets.
Minimum and maximum context lengths and the availability of a full target
horizon determine eligibility. Later rolling contexts may legitimately contain
earlier targets during evaluation; they are not pooled as adaptation histories.
This reference differs from GIFT's official training prefixes, so absolute pair
counts are not leakage rates for comparing the two benchmarks.

Of these cells, 250 share one of 56 declared dataset families;
9,750 cross-family cells remain identity-unresolved. An additional
48 within-family cells have no declared shared item/target identity.
Frequency variants share a family, with conservative aliases for solar,
monthly/quarterly FRED, and UK COVID representations. Matching item IDs and
target-column names establish the audit's declared alignment. Different
entity IDs, spatial aggregates, and new-versus-cumulative target names are
not silently equated. Same-frequency conflicts require equal finite float32
values at the same timestamp. Cross-frequency candidates use intersecting
calendar-period supports, following the GIFT audit convention; upstream
aggregation values and labeling are not reconstructed for these candidates.

The audit checks 235,039 eligible univariate rolling windows and
8,976,557 target positions, including missing positions in the window count.
It flags 48 distinct ordered task pairs in 21 families:
0 verified same-frequency value conflicts and 48
cross-frequency support candidates. Deduplicating repeated tasks and windows
within each stored configuration yields 1,557,899 finite affected target
observations on 37,750 declared series identities. Series are keyed by
family, item ID, and target column across frequencies; target observations
are keyed by stored configuration, item ID, target column, and raw position.
M5, Seattle Loop, and Favorita stores account for 88\% of the affected
target observations. Different frequencies are not treated as a
common-duration time grid.

\paragraph{Prepared In-Domain Histories.}
The separate corpus contains 121,335 univariate histories and 76,105,168
values, of which 75,193,885 are finite. These stored histories may represent
the same entity at different frequencies. All 100 supplied task definitions
contribute protected boundaries, even when a task has no usable adaptation
history. A family-wide cutoff avoids relying on entity or channel IDs to
protect cross-frequency variants. Its support guard extends one frequency
interval on either side of each timestamp, conservatively covering uncertain
aggregation labels; it is stricter than the calendar-period convention used
to count the candidates above. Every retained observation ends strictly
before the earliest protected support. All serialized prefix hashes and
member hashes were verified, and an independent terminal-support audit
checks all 121,335 retained histories, finding zero boundary violations.
Requiring 16 finite observations excludes 7,704 histories;
seven members remain empty.

\paragraph{Verified Source Supersets.}
Candidates from 28 existing source files cover 38 FEV configurations.
Containment requires the entire benchmark sequence to match both timestamps
and float32 target values, including missingness; IDs and hashes only propose
candidates. The search identifies 31,504 complete-series matches. A matched
source can contribute earlier history, but its protected suffix and any
later future are removed. A source matched to several tasks inherits their
earliest boundary before export. Supersets add one earlier observation to
each of 133 KDD 2022 histories; the other matches supply no larger usable
prefix. This is a candidate-source search, not an exhaustive search for all
renamed representations. Unknown cross-family partial or transformed copies
remain unresolved. These data checks neither establish an adaptation gain
nor remove possible contamination already present in a base checkpoint.


\section{Benchmarks and Task Inventory}
\label{sec:benchmarks}
Table~\ref{tab:evaluation-populations} summarizes the archived windows
and development subsets for both benchmarks.
FEV covers all rows in the archived benchmark extraction. GIFT covers 55 dataset/frequency configurations
and 21 additional medium/long pairs in the configured suite.
The ablations use FEV-Bench-S and GIFT-Eval-S
(Section~\ref{sec:development-sets}), with up to 1K context steps and
horizons capped at 128.
Section~\ref{sec:evaluation-metrics} defines the seasonal scoring and
geometric task aggregation.
The inventories below describe native archive lengths, not model scores.

\begin{table}[!htbp]
\centering
\caption{Evaluation populations. A series can contribute several forecast origins.
Section~\ref{sec:development-sets} defines the development subsets.
The number of scored windows depends on metric eligibility.}
\label{tab:evaluation-populations}\small
\setlength{\tabcolsep}{6pt}
\reporttablestyle
\begin{tabular}{@{}lrrlr@{}}
\toprule
\reporttablehead{Benchmark} & \reporttablehead{Tasks} &
\reporttablehead{Archived windows} & \reporttablehead{Development subset} &
\reporttablehead{Candidate windows}\\
\midrule
GIFT-Eval & 97 & 371,330 & GIFT-Eval-S & 32K\\
FEV-Bench & 100 & 235,039 & FEV-Bench-S & 32K\\
\bottomrule
\end{tabular}
\end{table}

\paragraph{Model Release Rankings.}
For each benchmark, we compare the evaluated \textsc{Tafsut} checkpoint with
zero-shot entries whose weights have been released for download,
including gated research releases. We exclude API-only entries and
entries without released weights. We verify weight availability from
the authors' repositories and checkpoint links for 47 of the 61 zero-shot
entries on GIFT-Eval and 34 of the 39 on FEV-Bench. The adapted GIFT-Eval
ranking also excludes Chronos-2 Synth. We retain distinct releases
such as TimesFM-2.5 and TimesFM-3 and use the best score per metric
across variants of a release. On GIFT-Eval, we group the variants of
FlowState r1.1, PatchTST-FM r1, Kairos, Moirai 1.1, Reverso, Simons,
TimeH2O-v1, TimeSSM, Timestral, and YingLong. On FEV-Bench, we group
variants of Simons, TimeH2O-v1, Timestral, and Toto 2.0. For missing
task scores we use Seasonal Naive, and for flagged training overlap
we use Chronos-Bolt, following the FEV-Bench protocol.
We retrieved the archives on 24~November~2026
(\href{https://huggingface.co/spaces/Salesforce/GIFT-Eval/tree/95e9a011172ce9a26f72716a0d82f4eb32299b42/results}{GIFT-Eval},
\href{https://github.com/autogluon/fev/tree/ecce67120aae4827e9702b978a9e9327ecde81d0/benchmarks/fev_bench}{FEV-Bench}).

\begingroup
\captionsetup{font=footnotesize,skip=3pt,hypcap=false}
\newcommand{\lengthbars}[1]{\textcolor{reportaccent}{\rule{#1\dimexpr2.5pt\relax}{3.5pt}}\textcolor{black!12}{\rule{\dimexpr12.5pt-#1\dimexpr2.5pt\relax}{3.5pt}}}
\ExplSyntaxOn
\NewDocumentCommand{\tasklabel}{mmm}{
  \group_begin:
  \tl_set:Nn \l_tmpa_tl {#2}
  \hbox_set:Nn \l_tmpa_box {#2#3}
  \dim_compare:nNnT {\box_wd:N \l_tmpa_box} > {\linewidth} {
    \bool_while_do:nn
      {\dim_compare_p:nNn {\box_wd:N \l_tmpa_box} > {\linewidth}}
      {
        \tl_set:Nx \l_tmpa_tl {\tl_range:Nnn \l_tmpa_tl {1}{-2}}
        \hbox_set:Nn \l_tmpa_box {\tl_use:N \l_tmpa_tl\ldots#3}
      }
  }
  \box_use:N \l_tmpa_box
  \group_end:
}
\ExplSyntaxOff
\fontsize{7.5}{8.8}\selectfont
\noindent\textbf{How to Read.} Read top to bottom within each block, then move right,
ordered by increasing mean context within each benchmark.
GIFT-Eval superscripts s/m/l denote short/medium/long tasks.
The $N$ column counts distinct univariate series per task (not forecast windows);
repeated forecast origins are counted once. A multivariate target contributes
one series per variate, and the same series may occur in multiple tasks.
The $m$ column gives the seasonal period in observed steps.
C/H bars encode window-mean context/horizon lengths, including missing positions.

\smallskip
\reporttablestyle
\noindent\begin{tabularx}{\linewidth}{@{}l*{5}{>{\centering\arraybackslash}X}@{}}
\toprule
\reporttablehead{Filled units} & \lengthbars{1} & \lengthbars{2} & \lengthbars{3} & \lengthbars{4} & \lengthbars{5} \\
\midrule
C (steps) & $\leq128$ & $(128,1\mathrm{K}]$ & $(1\mathrm{K},8\mathrm{K}]$ & $(8\mathrm{K},64\mathrm{K}]$ & $>64\mathrm{K}$ \\
H (steps) & $\leq16$ & $(16,64]$ & $(64,128]$ & $(128,512]$ & $>512$ \\
\bottomrule
\end{tabularx}
\par\smallskip
\noindent Exact window counts and
seasonal periods are retained in the supplementary task key; bar lengths
are discrete bins, not proportional scales.
\par\smallskip
\begin{table}[!htbp]
\centering
\fontsize{7.5}{8.8}\selectfont
\caption{FEV-Bench: all 100 archived tasks. Ellipses indicate shortened names; frequency suffixes are retained; $N$ counts distinct univariate series; $m$ is the seasonal period.}\label{tab:full-fev-tasks}
\setlength{\tabcolsep}{1pt}\renewcommand{\arraystretch}{0.90}
\reporttablestyle
\begin{tabularx}{\linewidth}{@{}*{3}{>{\raggedright\arraybackslash}Xrrcc}@{}}
\toprule
\reporttablehead{Task} & \reporttablehead{$N$} & \reporttablehead{$m$} & \reporttablehead{C} & \reporttablehead{H} &
\reporttablehead{Task} & \reporttablehead{$N$} & \reporttablehead{$m$} & \reporttablehead{C} & \reporttablehead{H} &
\reporttablehead{Task} & \reporttablehead{$N$} & \reporttablehead{$m$} & \reporttablehead{C} & \reporttablehead{H} \\
\midrule
\tasklabel{f63}{world\_tourism}{} & 178 & 1 & \lengthbars{1} & \lengthbars{1} & \tasklabel{f39}{uci\_air\_quality}{\_1D} & 4 & 7 & \lengthbars{2} & \lengthbars{2} & \tasklabel{f77}{boomlet\_1062}{} & 21 & 288 & \lengthbars{4} & \lengthbars{4} \\
\tasklabel{f62}{australian\_tourism}{} & 89 & 4 & \lengthbars{1} & \lengthbars{1} & \tasklabel{f17}{restaurant}{} & 817 & 7 & \lengthbars{2} & \lengthbars{2} & \tasklabel{f78}{boomlet\_1209}{} & 53 & 288 & \lengthbars{4} & \lengthbars{4} \\
\tasklabel{f64}{world\_co2\_emissions}{} & 191 & 1 & \lengthbars{1} & \lengthbars{1} & \tasklabel{f53}{ETT}{\_1D} & 14 & 7 & \lengthbars{2} & \lengthbars{2} & \tasklabel{f80}{boomlet\_1230}{} & 23 & 288 & \lengthbars{4} & \lengthbars{4} \\
\tasklabel{f27}{favorita\_stores}{\_1M} & 1579 & 12 & \lengthbars{1} & \lengthbars{1} & \tasklabel{f72}{uk\_covid\_nation}{\_1D} & 12 & 7 & \lengthbars{2} & \lengthbars{2} & \tasklabel{f82}{boomlet\_1487}{} & 54 & 288 & \lengthbars{4} & \lengthbars{4} \\
\tasklabel{f30}{favorita\_transactions}{\_1M} & 51 & 12 & \lengthbars{1} & \lengthbars{1} & \tasklabel{f73}{uk\_covid\_nation}{\_1D} & 12 & 1 & \lengthbars{2} & \lengthbars{2} & \tasklabel{f54}{ETT}{\_1H} & 14 & 24 & \lengthbars{4} & \lengthbars{4} \\
\tasklabel{f98}{us\_consumption}{\_1Y} & 31 & 1 & \lengthbars{1} & \lengthbars{1} & \tasklabel{f46}{SZ\_TAXI}{\_1H} & 156 & 24 & \lengthbars{2} & \lengthbars{4} & \tasklabel{f79}{boomlet\_1225}{} & 49 & 1440 & \lengthbars{4} & \lengthbars{2} \\
\tasklabel{f34}{solar}{\_1W} & 137 & 1 & \lengthbars{1} & \lengthbars{1} & \tasklabel{f70}{uk\_covid\_utla}{\_1D} & 214 & 7 & \lengthbars{2} & \lengthbars{2} & \tasklabel{f81}{boomlet\_1282}{} & 35 & 1440 & \lengthbars{4} & \lengthbars{2} \\
\tasklabel{f22}{m5}{\_1M} & 29364 & 12 & \lengthbars{1} & \lengthbars{1} & \tasklabel{f43}{M\_DENSE}{\_1D} & 30 & 7 & \lengthbars{2} & \lengthbars{2} & \tasklabel{f88}{boomlet\_285}{} & 75 & 1440 & \lengthbars{4} & \lengthbars{2} \\
\tasklabel{f65}{world\_life\_expectancy}{} & 237 & 1 & \lengthbars{1} & \lengthbars{1} & \tasklabel{f92}{fred\_md\_2025}{} & 3 & 1 & \lengthbars{2} & \lengthbars{1} & \tasklabel{f89}{boomlet\_619}{} & 52 & 1440 & \lengthbars{4} & \lengthbars{2} \\
\tasklabel{f56}{hospital}{} & 767 & 12 & \lengthbars{1} & \lengthbars{1} & \tasklabel{f94}{fred\_md\_2025}{} & 51 & 1 & \lengthbars{2} & \lengthbars{1} & \tasklabel{f90}{boomlet\_772}{} & 67 & 1440 & \lengthbars{4} & \lengthbars{2} \\
\tasklabel{f52}{ETT}{\_1W} & 14 & 1 & \lengthbars{1} & \lengthbars{1} & \tasklabel{f15}{rossmann}{\_1D} & 1115 & 7 & \lengthbars{2} & \lengthbars{2} & \tasklabel{f91}{boomlet\_963}{} & 28 & 1440 & \lengthbars{4} & \lengthbars{2} \\
\tasklabel{f71}{uk\_covid\_utla}{\_1W} & 214 & 1 & \lengthbars{1} & \lengthbars{1} & \tasklabel{f96}{us\_consumption}{\_1M} & 31 & 12 & \lengthbars{2} & \lengthbars{1} & \tasklabel{f01}{proenfo\_gfc14}{} & 1 & 24 & \lengthbars{4} & \lengthbars{4} \\
\tasklabel{f16}{rossmann}{\_1W} & 1115 & 1 & \lengthbars{1} & \lengthbars{1} & \tasklabel{f68}{ercot}{\_1W} & 8 & 1 & \lengthbars{2} & \lengthbars{1} & \tasklabel{f02}{proenfo\_gfc17}{} & 8 & 24 & \lengthbars{4} & \lengthbars{4} \\
\tasklabel{f74}{uk\_covid\_nation}{\_1W} & 12 & 1 & \lengthbars{1} & \lengthbars{1} & \tasklabel{f04}{rohlik\_orders}{\_1D} & 7 & 7 & \lengthbars{2} & \lengthbars{2} & \tasklabel{f44}{M\_DENSE}{\_1H} & 30 & 24 & \lengthbars{4} & \lengthbars{4} \\
\tasklabel{f75}{uk\_covid\_nation}{\_1W} & 12 & 1 & \lengthbars{1} & \lengthbars{1} & \tasklabel{f03}{rohlik\_sales}{\_1D} & 4116 & 7 & \lengthbars{2} & \lengthbars{1} & \tasklabel{f47}{redset}{\_5T} & 118 & 288 & \lengthbars{4} & \lengthbars{4} \\
\tasklabel{f19}{walmart}{} & 2936 & 1 & \lengthbars{1} & \lengthbars{2} & \tasklabel{f50}{hospital\_admissions}{\_1D} & 8 & 7 & \lengthbars{3} & \lengthbars{2} & \tasklabel{f61}{bizitobs\_l2c}{\_5T} & 7 & 288 & \lengthbars{4} & \lengthbars{4} \\
\tasklabel{f69}{ercot}{\_1M} & 8 & 12 & \lengthbars{1} & \lengthbars{1} & \tasklabel{f25}{favorita\_stores}{\_1D} & 1579 & 7 & \lengthbars{3} & \lengthbars{2} & \tasklabel{f35}{kdd\_cup\_2022}{\_10T} & 134 & 144 & \lengthbars{4} & \lengthbars{4} \\
\tasklabel{f76}{ecdc\_ili}{} & 25 & 1 & \lengthbars{1} & \lengthbars{1} & \tasklabel{f28}{favorita\_transactions}{\_1D} & 51 & 7 & \lengthbars{3} & \lengthbars{2} & \tasklabel{f00}{proenfo\_gfc12}{} & 11 & 24 & \lengthbars{4} & \lengthbars{4} \\
\tasklabel{f06}{rohlik\_orders}{\_1W} & 7 & 1 & \lengthbars{1} & \lengthbars{1} & \tasklabel{f20}{m5}{\_1D} & 30490 & 7 & \lengthbars{3} & \lengthbars{2} & \tasklabel{f09}{entsoe}{\_1H} & 6 & 24 & \lengthbars{4} & \lengthbars{4} \\
\tasklabel{f05}{rohlik\_sales}{\_1W} & 3942 & 1 & \lengthbars{2} & \lengthbars{1} & \tasklabel{f23}{hierarchical\_sales}{\_1D} & 118 & 7 & \lengthbars{3} & \lengthbars{2} & \tasklabel{f32}{solar\_with\_weather}{\_1H} & 1 & 24 & \lengthbars{4} & \lengthbars{2} \\
\tasklabel{f99}{gvar}{} & 198 & 4 & \lengthbars{2} & \lengthbars{1} & \tasklabel{f49}{redset}{\_1H} & 138 & 24 & \lengthbars{3} & \lengthbars{2} & \tasklabel{f57}{jena\_weather}{\_10T} & 21 & 144 & \lengthbars{4} & \lengthbars{4} \\
\tasklabel{f51}{hospital\_admissions}{\_1W} & 8 & 1 & \lengthbars{2} & \lengthbars{1} & \tasklabel{f60}{bizitobs\_l2c}{\_1H} & 7 & 24 & \lengthbars{3} & \lengthbars{2} & \tasklabel{f10}{epf\_be}{} & 1 & 24 & \lengthbars{4} & \lengthbars{2} \\
\tasklabel{f37}{kdd\_cup\_2022}{\_1D} & 134 & 1 & \lengthbars{2} & \lengthbars{1} & \tasklabel{f45}{SZ\_TAXI}{\_15T} & 156 & 96 & \lengthbars{3} & \lengthbars{3} & \tasklabel{f11}{epf\_de}{} & 1 & 24 & \lengthbars{4} & \lengthbars{2} \\
\tasklabel{f26}{favorita\_stores}{\_1W} & 1579 & 1 & \lengthbars{2} & \lengthbars{1} & \tasklabel{f85}{boomlet\_1855}{} & 52 & 24 & \lengthbars{3} & \lengthbars{2} & \tasklabel{f12}{epf\_fr}{} & 1 & 24 & \lengthbars{4} & \lengthbars{2} \\
\tasklabel{f29}{favorita\_transactions}{\_1W} & 51 & 1 & \lengthbars{2} & \lengthbars{1} & \tasklabel{f86}{boomlet\_1975}{} & 75 & 24 & \lengthbars{3} & \lengthbars{2} & \tasklabel{f13}{epf\_np}{} & 1 & 24 & \lengthbars{4} & \lengthbars{2} \\
\tasklabel{f93}{fred\_qd\_2025}{} & 3 & 1 & \lengthbars{2} & \lengthbars{1} & \tasklabel{f87}{boomlet\_2187}{} & 100 & 24 & \lengthbars{3} & \lengthbars{2} & \tasklabel{f14}{epf\_pjm}{} & 1 & 24 & \lengthbars{4} & \lengthbars{2} \\
\tasklabel{f95}{fred\_qd\_2025}{} & 51 & 1 & \lengthbars{2} & \lengthbars{1} & \tasklabel{f67}{ercot}{\_1D} & 8 & 7 & \lengthbars{3} & \lengthbars{2} & \tasklabel{f55}{ETT}{\_15T} & 14 & 96 & \lengthbars{5} & \lengthbars{3} \\
\tasklabel{f24}{hierarchical\_sales}{\_1W} & 118 & 1 & \lengthbars{2} & \lengthbars{1} & \tasklabel{f38}{uci\_air\_quality}{\_1H} & 4 & 24 & \lengthbars{3} & \lengthbars{4} & \tasklabel{f08}{entsoe}{\_30T} & 6 & 48 & \lengthbars{5} & \lengthbars{3} \\
\tasklabel{f58}{jena\_weather}{\_1D} & 21 & 1 & \lengthbars{2} & \lengthbars{2} & \tasklabel{f48}{redset}{\_15T} & 126 & 96 & \lengthbars{3} & \lengthbars{3} & \tasklabel{f40}{LOOP\_SEATTLE}{\_5T} & 323 & 288 & \lengthbars{5} & \lengthbars{4} \\
\tasklabel{f21}{m5}{\_1W} & 30490 & 1 & \lengthbars{2} & \lengthbars{1} & \tasklabel{f42}{LOOP\_SEATTLE}{\_1H} & 323 & 24 & \lengthbars{3} & \lengthbars{4} & \tasklabel{f66}{ercot}{\_1H} & 8 & 24 & \lengthbars{5} & \lengthbars{4} \\
\tasklabel{f18}{hermes}{} & 10000 & 1 & \lengthbars{2} & \lengthbars{2} & \tasklabel{f59}{jena\_weather}{\_1H} & 21 & 24 & \lengthbars{4} & \lengthbars{2} & \tasklabel{f07}{entsoe}{\_15T} & 6 & 96 & \lengthbars{5} & \lengthbars{3} \\
\tasklabel{f33}{solar}{\_1D} & 137 & 1 & \lengthbars{2} & \lengthbars{2} & \tasklabel{f83}{boomlet\_1631}{} & 40 & 48 & \lengthbars{4} & \lengthbars{3} & \tasklabel{f31}{solar\_with\_weather}{\_15T} & 1 & 96 & \lengthbars{5} & \lengthbars{3} \\
\tasklabel{f41}{LOOP\_SEATTLE}{\_1D} & 323 & 7 & \lengthbars{2} & \lengthbars{2} & \tasklabel{f84}{boomlet\_1676}{} & 100 & 48 & \lengthbars{4} & \lengthbars{3} &  & & & &  \\
\tasklabel{f97}{us\_consumption}{\_1Q} & 31 & 4 & \lengthbars{2} & \lengthbars{1} & \tasklabel{f36}{kdd\_cup\_2022}{\_30T} & 134 & 48 & \lengthbars{4} & \lengthbars{3} &  & & & &  \\
\bottomrule
\end{tabularx}
\end{table}
\par\smallskip
\begin{table}[!htbp]
\centering
\fontsize{7.5}{8.8}\selectfont
\caption{GIFT-Eval: all 97 archived tasks. Ellipses indicate shortened names; frequency suffixes are retained; $N$ counts distinct univariate series; $m$ is the seasonal period.}\label{tab:full-gift-tasks}
\setlength{\tabcolsep}{1pt}\renewcommand{\arraystretch}{0.90}
\reporttablestyle
\begin{tabularx}{\linewidth}{@{}*{3}{>{\raggedright\arraybackslash}Xrrcc}@{}}
\toprule
\reporttablehead{Task} & \reporttablehead{$N$} & \reporttablehead{$m$} & \reporttablehead{C} & \reporttablehead{H} &
\reporttablehead{Task} & \reporttablehead{$N$} & \reporttablehead{$m$} & \reporttablehead{C} & \reporttablehead{H} &
\reporttablehead{Task} & \reporttablehead{$N$} & \reporttablehead{$m$} & \reporttablehead{C} & \reporttablehead{H} \\
\midrule
\tasklabel{g01}{m4\_yearly}{\textsuperscript{s}} & 22974 & 1 & \lengthbars{1} & \lengthbars{1} & \tasklabel{g05}{m4\_daily}{\textsuperscript{s}} & 4227 & 1 & \lengthbars{3} & \lengthbars{1} & \tasklabel{g58}{ett1}{/H\textsuperscript{m}} & 7 & 24 & \lengthbars{4} & \lengthbars{4} \\
\tasklabel{g35}{car\_parts\_with\_missing}{\textsuperscript{s}} & 2674 & 12 & \lengthbars{1} & \lengthbars{1} & \tasklabel{g47}{SZ\_TAXI}{/15T\textsuperscript{m}} & 156 & 96 & \lengthbars{3} & \lengthbars{4} & \tasklabel{g66}{ett2}{/H\textsuperscript{m}} & 7 & 24 & \lengthbars{4} & \lengthbars{4} \\
\tasklabel{g22}{solar}{/W\textsuperscript{s}} & 137 & 1 & \lengthbars{1} & \lengthbars{1} & \tasklabel{g94}{bizitobs\_l2c}{/H\textsuperscript{s}} & 7 & 24 & \lengthbars{3} & \lengthbars{2} & \tasklabel{g51}{M\_DENSE}{/H\textsuperscript{m}} & 30 & 24 & \lengthbars{4} & \lengthbars{4} \\
\tasklabel{g23}{hospital}{\textsuperscript{s}} & 767 & 12 & \lengthbars{1} & \lengthbars{1} & \tasklabel{g46}{SZ\_TAXI}{/15T\textsuperscript{s}} & 156 & 96 & \lengthbars{3} & \lengthbars{2} & \tasklabel{g57}{ett1}{/H\textsuperscript{s}} & 7 & 24 & \lengthbars{4} & \lengthbars{2} \\
\tasklabel{g61}{ett1}{/W\textsuperscript{s}} & 7 & 1 & \lengthbars{1} & \lengthbars{1} & \tasklabel{g30}{saugeenday}{/W\textsuperscript{s}} & 1 & 1 & \lengthbars{3} & \lengthbars{1} & \tasklabel{g65}{ett2}{/H\textsuperscript{s}} & 7 & 24 & \lengthbars{4} & \lengthbars{2} \\
\tasklabel{g69}{ett2}{/W\textsuperscript{s}} & 7 & 1 & \lengthbars{1} & \lengthbars{1} & \tasklabel{g25}{us\_births}{/D\textsuperscript{s}} & 1 & 1 & \lengthbars{3} & \lengthbars{2} & \tasklabel{g50}{M\_DENSE}{/H\textsuperscript{s}} & 30 & 24 & \lengthbars{4} & \lengthbars{2} \\
\tasklabel{g02}{m4\_quarterly}{\textsuperscript{s}} & 24000 & 4 & \lengthbars{1} & \lengthbars{1} & \tasklabel{g79}{bitbrains\_fast\_storage}{/5T\textsuperscript{l}} & 2500 & 288 & \lengthbars{3} & \lengthbars{5} & \tasklabel{g28}{saugeenday}{/D\textsuperscript{s}} & 1 & 1 & \lengthbars{4} & \lengthbars{2} \\
\tasklabel{g24}{covid\_deaths}{\textsuperscript{s}} & 266 & 1 & \lengthbars{2} & \lengthbars{2} & \tasklabel{g83}{bitbrains\_rnd}{/5T\textsuperscript{l}} & 1000 & 288 & \lengthbars{3} & \lengthbars{5} & \tasklabel{g93}{bizitobs\_l2c}{/5T\textsuperscript{l}} & 7 & 288 & \lengthbars{4} & \lengthbars{5} \\
\tasklabel{g14}{electricity}{/W\textsuperscript{s}} & 370 & 1 & \lengthbars{2} & \lengthbars{1} & \tasklabel{g20}{solar}{/H\textsuperscript{l}} & 137 & 24 & \lengthbars{3} & \lengthbars{5} & \tasklabel{g92}{bizitobs\_l2c}{/5T\textsuperscript{m}} & 7 & 288 & \lengthbars{4} & \lengthbars{4} \\
\tasklabel{g03}{m4\_monthly}{\textsuperscript{s}} & 48000 & 12 & \lengthbars{2} & \lengthbars{2} & \tasklabel{g44}{LOOP\_SEATTLE}{/H\textsuperscript{l}} & 323 & 24 & \lengthbars{3} & \lengthbars{5} & \tasklabel{g91}{bizitobs\_l2c}{/5T\textsuperscript{s}} & 7 & 288 & \lengthbars{4} & \lengthbars{2} \\
\tasklabel{g26}{us\_births}{/M\textsuperscript{s}} & 1 & 12 & \lengthbars{2} & \lengthbars{1} & \tasklabel{g75}{jena\_weather}{/H\textsuperscript{l}} & 21 & 24 & \lengthbars{3} & \lengthbars{5} & \tasklabel{g11}{electricity}{/H\textsuperscript{m}} & 370 & 24 & \lengthbars{4} & \lengthbars{4} \\
\tasklabel{g38}{hierarchical\_sales}{/W\textsuperscript{s}} & 118 & 1 & \lengthbars{2} & \lengthbars{1} & \tasklabel{g78}{bitbrains\_fast\_storage}{/5T\textsuperscript{m}} & 2500 & 288 & \lengthbars{3} & \lengthbars{4} & \tasklabel{g12}{electricity}{/H\textsuperscript{l}} & 370 & 24 & \lengthbars{4} & \lengthbars{5} \\
\tasklabel{g21}{solar}{/D\textsuperscript{s}} & 137 & 1 & \lengthbars{2} & \lengthbars{2} & \tasklabel{g82}{bitbrains\_rnd}{/5T\textsuperscript{m}} & 1000 & 288 & \lengthbars{3} & \lengthbars{4} & \tasklabel{g10}{electricity}{/H\textsuperscript{s}} & 370 & 24 & \lengthbars{4} & \lengthbars{2} \\
\tasklabel{g45}{LOOP\_SEATTLE}{/D\textsuperscript{s}} & 323 & 1 & \lengthbars{2} & \lengthbars{2} & \tasklabel{g86}{bizitobs\_application}{\textsuperscript{m}} & 2 & 360 & \lengthbars{3} & \lengthbars{5} & \tasklabel{g17}{solar}{/10T\textsuperscript{l}} & 137 & 144 & \lengthbars{4} & \lengthbars{5} \\
\tasklabel{g76}{jena\_weather}{/D\textsuperscript{s}} & 21 & 1 & \lengthbars{2} & \lengthbars{2} & \tasklabel{g87}{bizitobs\_application}{\textsuperscript{l}} & 2 & 360 & \lengthbars{3} & \lengthbars{5} & \tasklabel{g72}{jena\_weather}{/10T\textsuperscript{l}} & 21 & 144 & \lengthbars{4} & \lengthbars{5} \\
\tasklabel{g36}{restaurant}{\textsuperscript{s}} & 807 & 1 & \lengthbars{2} & \lengthbars{2} & \tasklabel{g89}{bizitobs\_service}{\textsuperscript{m}} & 42 & 360 & \lengthbars{3} & \lengthbars{5} & \tasklabel{g16}{solar}{/10T\textsuperscript{m}} & 137 & 144 & \lengthbars{4} & \lengthbars{4} \\
\tasklabel{g34}{kdd\_cup\_2018\_with\_missing}{/D\textsuperscript{s}} & 270 & 1 & \lengthbars{2} & \lengthbars{2} & \tasklabel{g90}{bizitobs\_service}{\textsuperscript{l}} & 42 & 360 & \lengthbars{3} & \lengthbars{5} & \tasklabel{g71}{jena\_weather}{/10T\textsuperscript{m}} & 21 & 144 & \lengthbars{4} & \lengthbars{4} \\
\tasklabel{g84}{bitbrains\_rnd}{/H\textsuperscript{s}} & 1000 & 24 & \lengthbars{2} & \lengthbars{2} & \tasklabel{g19}{solar}{/H\textsuperscript{m}} & 137 & 24 & \lengthbars{3} & \lengthbars{4} & \tasklabel{g15}{solar}{/10T\textsuperscript{s}} & 137 & 144 & \lengthbars{4} & \lengthbars{2} \\
\tasklabel{g80}{bitbrains\_fast\_storage}{/H\textsuperscript{s}} & 2500 & 24 & \lengthbars{2} & \lengthbars{2} & \tasklabel{g43}{LOOP\_SEATTLE}{/H\textsuperscript{m}} & 323 & 24 & \lengthbars{3} & \lengthbars{4} & \tasklabel{g70}{jena\_weather}{/10T\textsuperscript{s}} & 21 & 144 & \lengthbars{4} & \lengthbars{2} \\
\tasklabel{g00}{temperature\_rain\_with\_missing}{\textsuperscript{s}} & 32072 & 1 & \lengthbars{2} & \lengthbars{2} & \tasklabel{g74}{jena\_weather}{/H\textsuperscript{m}} & 21 & 24 & \lengthbars{3} & \lengthbars{4} & \tasklabel{g56}{ett1}{/15T\textsuperscript{l}} & 7 & 96 & \lengthbars{5} & \lengthbars{5} \\
\tasklabel{g60}{ett1}{/D\textsuperscript{s}} & 7 & 1 & \lengthbars{2} & \lengthbars{2} & \tasklabel{g77}{bitbrains\_fast\_storage}{/5T\textsuperscript{s}} & 2500 & 288 & \lengthbars{3} & \lengthbars{2} & \tasklabel{g64}{ett2}{/15T\textsuperscript{l}} & 7 & 96 & \lengthbars{5} & \lengthbars{5} \\
\tasklabel{g68}{ett2}{/D\textsuperscript{s}} & 7 & 1 & \lengthbars{2} & \lengthbars{2} & \tasklabel{g81}{bitbrains\_rnd}{/5T\textsuperscript{s}} & 1000 & 288 & \lengthbars{3} & \lengthbars{2} & \tasklabel{g55}{ett1}{/15T\textsuperscript{m}} & 7 & 96 & \lengthbars{5} & \lengthbars{4} \\
\tasklabel{g53}{M\_DENSE}{/D\textsuperscript{s}} & 30 & 1 & \lengthbars{2} & \lengthbars{2} & \tasklabel{g18}{solar}{/H\textsuperscript{s}} & 137 & 24 & \lengthbars{4} & \lengthbars{2} & \tasklabel{g63}{ett2}{/15T\textsuperscript{m}} & 7 & 96 & \lengthbars{5} & \lengthbars{4} \\
\tasklabel{g49}{SZ\_TAXI}{/H\textsuperscript{s}} & 156 & 24 & \lengthbars{2} & \lengthbars{2} & \tasklabel{g42}{LOOP\_SEATTLE}{/H\textsuperscript{s}} & 323 & 24 & \lengthbars{4} & \lengthbars{2} & \tasklabel{g54}{ett1}{/15T\textsuperscript{s}} & 7 & 96 & \lengthbars{5} & \lengthbars{2} \\
\tasklabel{g29}{saugeenday}{/M\textsuperscript{s}} & 1 & 12 & \lengthbars{2} & \lengthbars{1} & \tasklabel{g73}{jena\_weather}{/H\textsuperscript{s}} & 21 & 24 & \lengthbars{4} & \lengthbars{2} & \tasklabel{g62}{ett2}{/15T\textsuperscript{s}} & 7 & 96 & \lengthbars{5} & \lengthbars{2} \\
\tasklabel{g06}{m4\_hourly}{\textsuperscript{s}} & 414 & 24 & \lengthbars{2} & \lengthbars{2} & \tasklabel{g85}{bizitobs\_application}{\textsuperscript{s}} & 2 & 360 & \lengthbars{4} & \lengthbars{2} & \tasklabel{g41}{LOOP\_SEATTLE}{/5T\textsuperscript{l}} & 323 & 288 & \lengthbars{5} & \lengthbars{5} \\
\tasklabel{g27}{us\_births}{/W\textsuperscript{s}} & 1 & 1 & \lengthbars{2} & \lengthbars{1} & \tasklabel{g88}{bizitobs\_service}{\textsuperscript{s}} & 42 & 360 & \lengthbars{4} & \lengthbars{2} & \tasklabel{g40}{LOOP\_SEATTLE}{/5T\textsuperscript{m}} & 323 & 288 & \lengthbars{5} & \lengthbars{4} \\
\tasklabel{g04}{m4\_weekly}{\textsuperscript{s}} & 359 & 1 & \lengthbars{2} & \lengthbars{1} & \tasklabel{g33}{kdd\_cup\_2018\_with\_missing}{/H\textsuperscript{l}} & 270 & 24 & \lengthbars{4} & \lengthbars{5} & \tasklabel{g39}{LOOP\_SEATTLE}{/5T\textsuperscript{s}} & 323 & 288 & \lengthbars{5} & \lengthbars{2} \\
\tasklabel{g13}{electricity}{/D\textsuperscript{s}} & 370 & 1 & \lengthbars{3} & \lengthbars{2} & \tasklabel{g32}{kdd\_cup\_2018\_with\_missing}{/H\textsuperscript{m}} & 270 & 24 & \lengthbars{4} & \lengthbars{4} & \tasklabel{g09}{electricity}{/15T\textsuperscript{l}} & 370 & 96 & \lengthbars{5} & \lengthbars{5} \\
\tasklabel{g37}{hierarchical\_sales}{/D\textsuperscript{s}} & 118 & 1 & \lengthbars{3} & \lengthbars{2} & \tasklabel{g31}{kdd\_cup\_2018\_with\_missing}{/H\textsuperscript{s}} & 270 & 24 & \lengthbars{4} & \lengthbars{2} & \tasklabel{g08}{electricity}{/15T\textsuperscript{m}} & 370 & 96 & \lengthbars{5} & \lengthbars{4} \\
\tasklabel{g96}{bizitobs\_l2c}{/H\textsuperscript{l}} & 7 & 24 & \lengthbars{3} & \lengthbars{5} & \tasklabel{g59}{ett1}{/H\textsuperscript{l}} & 7 & 24 & \lengthbars{4} & \lengthbars{5} & \tasklabel{g07}{electricity}{/15T\textsuperscript{s}} & 370 & 96 & \lengthbars{5} & \lengthbars{2} \\
\tasklabel{g95}{bizitobs\_l2c}{/H\textsuperscript{m}} & 7 & 24 & \lengthbars{3} & \lengthbars{4} & \tasklabel{g67}{ett2}{/H\textsuperscript{l}} & 7 & 24 & \lengthbars{4} & \lengthbars{5} &  & & & &  \\
\tasklabel{g48}{SZ\_TAXI}{/15T\textsuperscript{l}} & 156 & 96 & \lengthbars{3} & \lengthbars{5} & \tasklabel{g52}{M\_DENSE}{/H\textsuperscript{l}} & 30 & 24 & \lengthbars{4} & \lengthbars{5} &  & & & &  \\
\bottomrule
\end{tabularx}
\end{table}
\par\smallskip
\endgroup

\begin{table}[!htbp]
\centering
\footnotesize
\caption{FEV-Bench task ranks in Figure~\ref{fig:fev-covariate-effect}. Ranks restart in each panel.}
\label{tab:fev-covariate-ranks}
\setlength{\tabcolsep}{4pt}
\reporttablestyle[3]
\begin{tabularx}{\linewidth}{@{}rlrlr>{\raggedright\arraybackslash}X@{}}
\toprule
\multicolumn{4}{c}{\reporttablehead{(a) Known-future covariates}} &
\multicolumn{2}{c}{\reporttablehead{(b) Past covariates only}} \\
\cmidrule(r){1-4}\cmidrule(l){5-6}
\reporttablehead{Rank} & \reporttablehead{Task} &
\reporttablehead{Rank} & \reporttablehead{Task} &
\reporttablehead{Rank} & \reporttablehead{Task} \\
\midrule
1 & rossmann\_1W & 16 & favorita\_stores\_1D & 1 & kdd\_cup\_2022\_10T \\
2 & epf\_de & 17 & solar\_with\_weather\_15T & 2 & kdd\_cup\_2022\_30T \\
3 & proenfo\_gfc17 & 18 & m5\_1M & 3 & favorita\_transactions\_1M \\
4 & epf\_np & 19 & uci\_air\_quality\_1D & 4 & fred\_md\_2025/cee \\
5 & proenfo\_gfc14 & 20 & m5\_1D & 5 & uk\_covid\_nation\_1W/new \\
6 & rossmann\_1D & 21 & m5\_1W & 6 & uk\_covid\_nation\_1D/new \\
7 & rohlik\_sales\_1W & 22 & epf\_pjm & 7 & fred\_qd\_2025/cee \\
8 & rohlik\_sales\_1D & 23 & favorita\_stores\_1W & 8 & uk\_covid\_nation\_1D/cumulative \\
9 & epf\_fr & 24 & walmart & 9 & kdd\_cup\_2022\_1D \\
10 & proenfo\_gfc12 & 25 & entsoe\_15T & 10 & uk\_covid\_nation\_1W/cumulative \\
11 & solar\_with\_weather\_1H & 26 & favorita\_transactions\_1D & 11 & gvar \\
12 & favorita\_stores\_1M & 27 & hermes & 12 & favorita\_transactions\_1W \\
13 & epf\_be & 28 & rohlik\_orders\_1D &  &  \\
14 & entsoe\_1H & 29 & entsoe\_30T &  &  \\
15 & uci\_air\_quality\_1H & 30 & rohlik\_orders\_1W &  &  \\
\bottomrule
\end{tabularx}
\end{table}

\section{FEV-Bench Pairwise Win Rates}
\label{app:fev-pairwise}

\begin{figure}[!htbp]
\centering
\includegraphics[width=\linewidth]{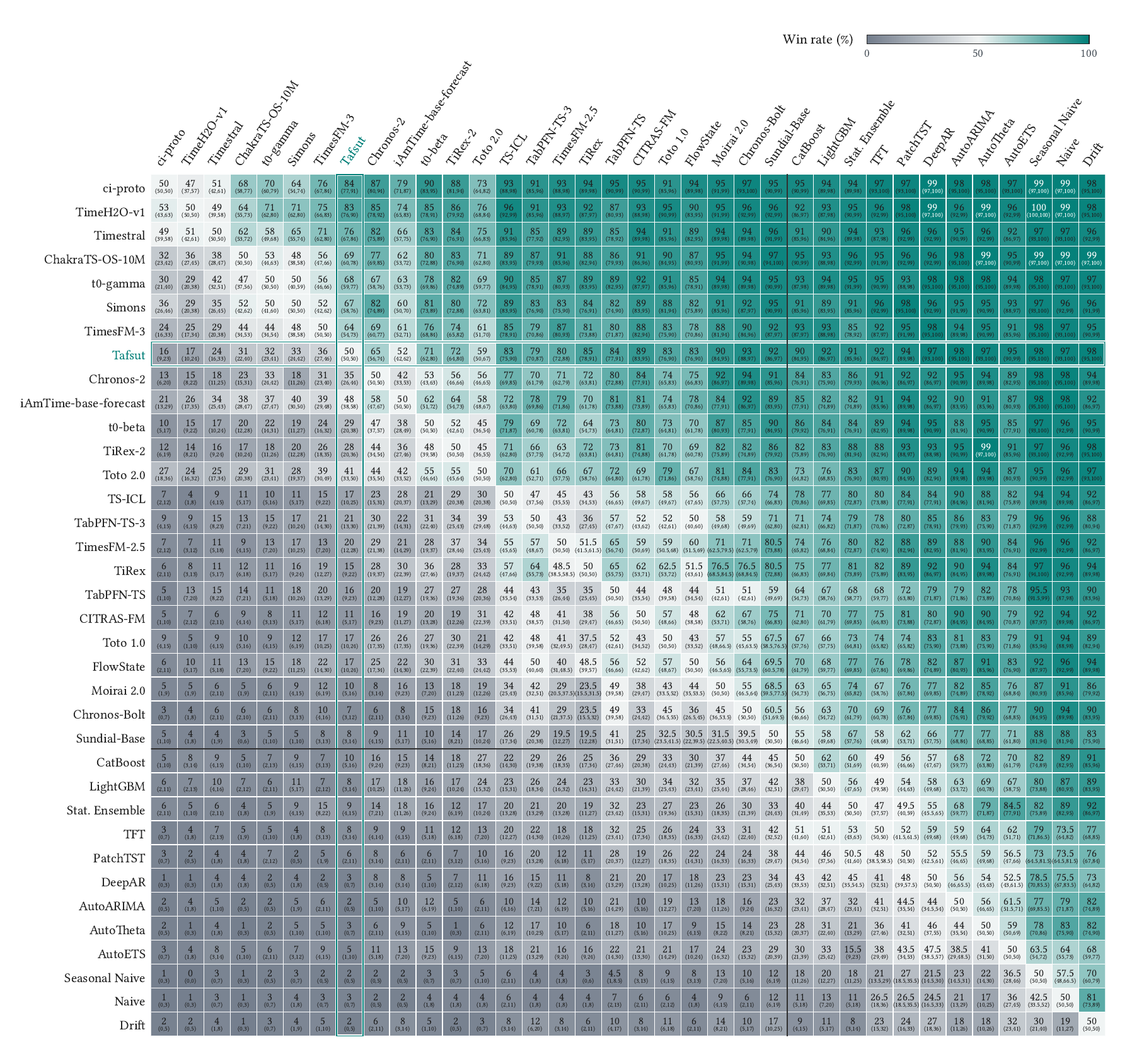}
\caption{Pairwise win rates under MASE. Each cell shows the row model's win rate in percent and its 95\% interval.
The outline identifies \textsc{Tafsut}. The divider separates foundation models
from task-specific and statistical baselines.}
\label{fig:fev-pairwise-mase}
\end{figure}

\begin{figure}[!tbp]
\centering
\includegraphics[width=\linewidth]{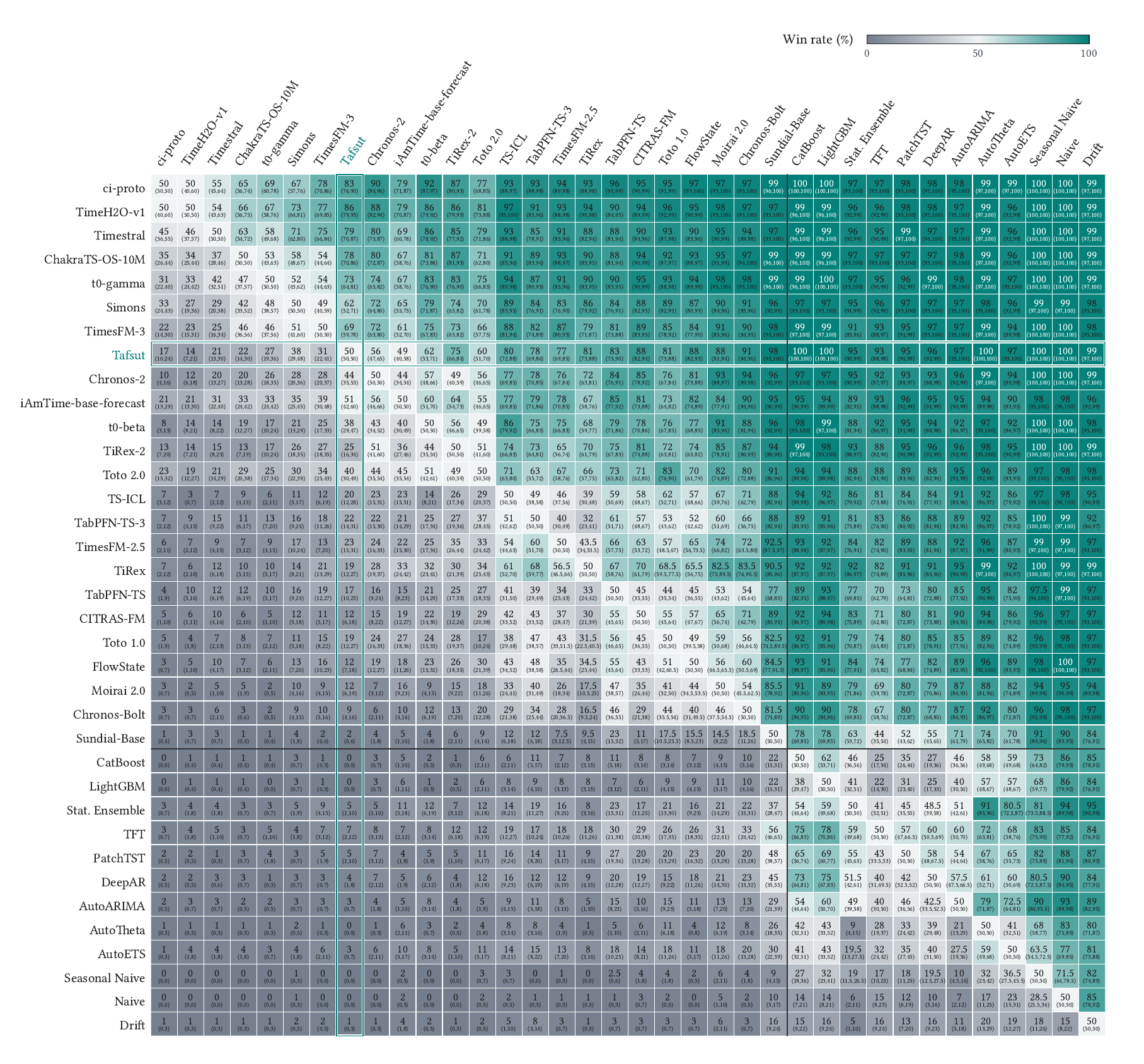}
\caption{Pairwise win rates under SQL. Model order and bootstrap procedure match Figure~\ref{fig:fev-pairwise-mase}.}
\label{fig:fev-pairwise-sql}
\end{figure}

For each metric, the win rate of row model $j$ against column model
$k$ is the fraction of the 100 tasks on which $j$ has lower error,
with ties contributing half a win. The intervals in
Figures~\ref{fig:fev-pairwise-mase} and~\ref{fig:fev-pairwise-sql}
are 95\% percentile intervals from 1,000 bootstrap draws of the
tasks with replacement. Each draw uses the same sampled tasks for
both models.

The matrices include the 24 open-weight zero-shot releases in
Figure~\ref{fig:fev-native-benchmark} and 12 task-specific or statistical
baselines. Each release uses the same variant for each metric as in the
bar plots. The public task scores come from the archived FEV-Bench
results~\citep{shchur2025fev}. For public
comparators, missing task scores take the corresponding Seasonal
Naive score, and tasks flagged for training overlap take the
Chronos-Bolt score. We use \textsc{Tafsut}'s measured scores on all 100 tasks
without these substitutions.

\section{Ablation Protocols and Additional Results}
\label{sec:ablations}

\paragraph{Evaluation and Controls.}
The ablation tables use the seasonal scoring rule in
Section~\ref{sec:evaluation-metrics} on FEV-Bench-S and GIFT-Eval-S.
The eligible sets contain 29,583 windows from 94 tasks on FEV-Bench-S
and 32,003 windows from 97 tasks on GIFT-Eval-S.
FEV excludes 3,044 windows for insufficient seasonal pairs, 59 for
a zero scale, and 82 for missing targets. The corresponding GIFT
counts are 248, 442, and 75.
For the synthetic validation bank, we use 1K history steps and 128 target
steps. Its losses use lag-one scaling and target-weighted arithmetic averaging.

Cells show the mean and sample standard deviation over epochs
360, 370, 380, 390, and 400. A signed change is
$100(S_{\mathrm{variant}}/S_{\mathrm{control}}-1)$ for mean scores $S$.
Bold indicates the lowest observed ablation mean in a column.
These checkpoint statistics measure
variation during training at one seed.

\paragraph{Generator Composition.}
\label{sec:preset-search}
The eleven candidates share a 104.7M-parameter gated-attention backbone
with patches of length and stride 32, a 1K-step context cap, a
128-step horizon, and training seed 42. Each epoch samples one million
windows, with batch size 8K. All use three-way mixup with concentration
0.15 and activation probability 0.5, point masking at 1\%, and no
PatchDrop. During the first 200 epochs each candidate uses a fresh
200,000-row source bank per epoch. The continuation restores the
optimizer and schedule through epoch 400; it can replay banks when
production falls behind.

The reference mixture is P02. The nonseasonal-dynamics preset P01, which
supplies the synthetic validation bank, reallocates five percentage points
from ordinary recurrence to
trend/state-space (1.75), stationary ARMA or SAR (1.25), volatility
(1.00), long memory (0.625), and delayed response (0.375). P03 moves
the same mass to sparse point-process events. P04 smooths the
seasonal component-count distribution at the same mean. P05 sets
the structural share to 5\%, P06 sets the modern share to 72.5\%,
and P07 sets the legacy share to 30\%. P08 and P09 apply half-strength
and full observation-profile shifts to the modern and structural families,
respectively. P10 removes the structural branch, and P11 fixes the
seasonal component count to one. Changes to branch shares rescale
the other branches.

Report labels P01--P03 correspond to archived IDs P02, P00, and P01,
respectively. Report labels P04--P11 correspond to archived IDs
P03--P10 in order.
At 200 epochs, nonseasonal dynamics gave the lowest MASE on the
GIFT-Eval development windows under the early scoring protocol
(2.1742 vs.\ 2.1868 for the reference). We used that mixture for subsequent
controlled studies.
Figure~\ref{fig:preset-transfer-scores} reports the 400-epoch
comparison under the scoring rules above.

\paragraph{Mixup and Masking.}
\label{sec:ablation-training}
Table~\ref{tab:ablation-training} compares seven settings with a
shared 104.6M-parameter backbone (14 blocks, width 768, 12 heads,
feed-forward width 3K, patch length and stride 32). Each setting
uses the same 80 cached source banks of 200,000 rows, replaying each
for five epochs and drawing one million windows per epoch.
Contexts are sampled uniformly within four equally likely buckets:
$[64,255]$, $[256,511]$, $[512,767]$, and $[768,1\mathrm{K}]$.
The horizon is 128. We use seed 42 and a global batch size of 8K.
AdamW uses $(\beta_1,\beta_2)=(0.9,0.999)$, weight decay 0.01, 123
warmup steps to $10^{-3}$, and a nominal 500-epoch cosine schedule
with minimum-to-peak ratio $10^{-4}$. Training stops at epoch 400.

Mixup~\citep{zhang2018mixup} draws a single weight vector
$\vect w\sim\operatorname{Dirichlet}(\alpha\vect{1}_K)$ for a
window with $K$ contributors. At time $t$, let $\mathcal J_t$ contain
the contributors $j$ with $x_{j,t}\in\mathbb{R}$. Mixup
combines their values in the original units as $\widetilde{x}_t =
\frac{\sum_{j\in\mathcal J_t} w_jx_{j,t}}{\sum_{j\in\mathcal J_t}w_j}$, when the denominator is positive, and sets $\widetilde{x}_t=\infty$
otherwise. The same weights apply to the history and target. The
control activates mixup with probability 0.5 per rank-local batch,
chooses two partners by batch permutations, and sets
$\alpha=0.15$. The sample-activated variants draw the activation
per window. Restricting partners to a context bucket also changes
sampling from batch permutations to per-window draws.

Point masking selects 1\% of observed historical values. PatchDrop
independently selects each eligible aligned block of 32 history steps
with probability 0.01. Both replace selected entries with
$\infty$ before mixing. Each row changes only the listed
setting relative to the control.

\begin{table}[!htbp]
\centering
\caption{Training augmentation. Changes are relative to the control with
batch-activated mixup, point masking, and PatchDrop. The bucket-restricted
setting also changes how partners are sampled.}
\label{tab:ablation-training}
\scriptsize
\setlength{\tabcolsep}{0.9pt}
\renewcommand{\arraystretch}{1.0}
\reporttablestyle[3]
\begin{tabular}{@{}l>{\raggedright\arraybackslash}p{2.1cm}rrrrrr@{}}
\toprule
& & \multicolumn{2}{c}{\reporttablehead{FEV-Bench-S}} &
\multicolumn{2}{c}{\reporttablehead{GIFT-Eval-S}} &
\multicolumn{2}{c}{\reporttablehead{Synthetic bank}} \\
\cmidrule(lr){3-4}\cmidrule(lr){5-6}\cmidrule(l){7-8}
\reporttablehead{Axis} & \reporttablehead{Configuration} &
\reporttablehead{SQL $\downarrow$} & \reporttablehead{MASE $\downarrow$} &
\reporttablehead{SQL $\downarrow$} & \reporttablehead{MASE $\downarrow$} &
\reporttablehead{SQL $\downarrow$} & \reporttablehead{MASE $\downarrow$} \\
\midrule
Control & Batch-activated mixup & \shortstack[r]{$0.8324\pm.0100$\\[-1pt]{\scriptsize\textcolor{ablationneutral}{$\Delta$ +0.00\%}}} & \shortstack[r]{$1.2048\pm.0069$\\[-1pt]{\scriptsize\textcolor{ablationneutral}{$\Delta$ +0.00\%}}} & \shortstack[r]{$0.7194\pm.0020$\\[-1pt]{\scriptsize\textcolor{ablationneutral}{$\Delta$ +0.00\%}}} & \shortstack[r]{$1.0534\pm.0033$\\[-1pt]{\scriptsize\textcolor{ablationneutral}{$\Delta$ +0.00\%}}} & \shortstack[r]{$1.2069\pm.0055$\\[-1pt]{\scriptsize\textcolor{ablationneutral}{$\Delta$ +0.00\%}}} & \shortstack[r]{$1.7952\pm.0065$\\[-1pt]{\scriptsize\textcolor{ablationneutral}{$\Delta$ +0.00\%}}}  \\
\midrule
Mixing & Sample-activated, unrestricted & \shortstack[r]{$0.8267\pm.0027$\\[-1pt]{\scriptsize\textcolor{ablationbetter}{$\Delta$ -0.69\%}}} & \shortstack[r]{$1.2010\pm.0051$\\[-1pt]{\scriptsize\textcolor{ablationbetter}{$\Delta$ -0.32\%}}} & \shortstack[r]{$\mathbf{0.7140}\pm.0012$\\[-1pt]{\scriptsize\textcolor{ablationbetter}{$\Delta$ -0.76\%}}} & \shortstack[r]{$\mathbf{1.0458}\pm.0018$\\[-1pt]{\scriptsize\textcolor{ablationbetter}{$\Delta$ -0.72\%}}} & \shortstack[r]{$1.1989\pm.0046$\\[-1pt]{\scriptsize\textcolor{ablationbetter}{$\Delta$ -0.66\%}}} & \shortstack[r]{$1.7862\pm.0073$\\[-1pt]{\scriptsize\textcolor{ablationbetter}{$\Delta$ -0.50\%}}}  \\
 & Sample-activated, bucket-restricted & \shortstack[r]{$0.8359\pm.0026$\\[-1pt]{\scriptsize\textcolor{ablationworse}{$\Delta$ +0.41\%}}} & \shortstack[r]{$1.2159\pm.0054$\\[-1pt]{\scriptsize\textcolor{ablationworse}{$\Delta$ +0.91\%}}} & \shortstack[r]{$0.7180\pm.0012$\\[-1pt]{\scriptsize\textcolor{ablationbetter}{$\Delta$ -0.21\%}}} & \shortstack[r]{$1.0520\pm.0012$\\[-1pt]{\scriptsize\textcolor{ablationbetter}{$\Delta$ -0.13\%}}} & \shortstack[r]{$1.2008\pm.0095$\\[-1pt]{\scriptsize\textcolor{ablationbetter}{$\Delta$ -0.50\%}}} & \shortstack[r]{$1.7967\pm.0170$\\[-1pt]{\scriptsize\textcolor{ablationworse}{$\Delta$ +0.08\%}}}  \\
 & No mixup & \shortstack[r]{$0.8456\pm.0046$\\[-1pt]{\scriptsize\textcolor{ablationworse}{$\Delta$ +1.58\%}}} & \shortstack[r]{$1.2227\pm.0080$\\[-1pt]{\scriptsize\textcolor{ablationworse}{$\Delta$ +1.49\%}}} & \shortstack[r]{$0.7247\pm.0012$\\[-1pt]{\scriptsize\textcolor{ablationworse}{$\Delta$ +0.73\%}}} & \shortstack[r]{$1.0596\pm.0032$\\[-1pt]{\scriptsize\textcolor{ablationworse}{$\Delta$ +0.59\%}}} & \shortstack[r]{$1.2261\pm.0155$\\[-1pt]{\scriptsize\textcolor{ablationworse}{$\Delta$ +1.59\%}}} & \shortstack[r]{$1.8361\pm.0238$\\[-1pt]{\scriptsize\textcolor{ablationworse}{$\Delta$ +2.28\%}}}  \\
\midrule
Masking & No PatchDrop & \shortstack[r]{$0.8315\pm.0023$\\[-1pt]{\scriptsize\textcolor{ablationbetter}{$\Delta$ -0.11\%}}} & \shortstack[r]{$1.2097\pm.0049$\\[-1pt]{\scriptsize\textcolor{ablationworse}{$\Delta$ +0.40\%}}} & \shortstack[r]{$0.7164\pm.0021$\\[-1pt]{\scriptsize\textcolor{ablationbetter}{$\Delta$ -0.43\%}}} & \shortstack[r]{$1.0511\pm.0021$\\[-1pt]{\scriptsize\textcolor{ablationbetter}{$\Delta$ -0.21\%}}} & \shortstack[r]{$1.1966\pm.0083$\\[-1pt]{\scriptsize\textcolor{ablationbetter}{$\Delta$ -0.85\%}}} & \shortstack[r]{$1.7873\pm.0092$\\[-1pt]{\scriptsize\textcolor{ablationbetter}{$\Delta$ -0.44\%}}}  \\
 & No point masking & \shortstack[r]{$\mathbf{0.8239}\pm.0021$\\[-1pt]{\scriptsize\textcolor{ablationbetter}{$\Delta$ -1.03\%}}} & \shortstack[r]{$\mathbf{1.2004}\pm.0041$\\[-1pt]{\scriptsize\textcolor{ablationbetter}{$\Delta$ -0.37\%}}} & \shortstack[r]{$0.7154\pm.0015$\\[-1pt]{\scriptsize\textcolor{ablationbetter}{$\Delta$ -0.56\%}}} & \shortstack[r]{$1.0497\pm.0024$\\[-1pt]{\scriptsize\textcolor{ablationbetter}{$\Delta$ -0.35\%}}} & \shortstack[r]{$1.2028\pm.0042$\\[-1pt]{\scriptsize\textcolor{ablationbetter}{$\Delta$ -0.33\%}}} & \shortstack[r]{$1.7934\pm.0042$\\[-1pt]{\scriptsize\textcolor{ablationbetter}{$\Delta$ -0.10\%}}}  \\
 & No masking & \shortstack[r]{$0.8289\pm.0035$\\[-1pt]{\scriptsize\textcolor{ablationbetter}{$\Delta$ -0.42\%}}} & \shortstack[r]{$1.2047\pm.0049$\\[-1pt]{\scriptsize\textcolor{ablationbetter}{$\Delta$ -0.01\%}}} & \shortstack[r]{$0.7162\pm.0018$\\[-1pt]{\scriptsize\textcolor{ablationbetter}{$\Delta$ -0.45\%}}} & \shortstack[r]{$1.0504\pm.0026$\\[-1pt]{\scriptsize\textcolor{ablationbetter}{$\Delta$ -0.28\%}}} & \shortstack[r]{$\mathbf{1.1832}\pm.0085$\\[-1pt]{\scriptsize\textcolor{ablationbetter}{$\Delta$ -1.96\%}}} & \shortstack[r]{$\mathbf{1.7600}\pm.0111$\\[-1pt]{\scriptsize\textcolor{ablationbetter}{$\Delta$ -1.96\%}}}  \\
\bottomrule
\end{tabular}
\end{table}

\paragraph{Sensitivity to Missing History.}
\label{sec:missing-context}
At epoch 400 we evaluate the four masking configurations with
512 or 1K history steps and the usual capped horizon.
We replace the same random points or contiguous spans with
$\infty$ across configurations at corruption levels
of 5\%, 10\%, and 20\%. Seasonal denominators and eligibility come from the
uncorrupted history at each cap. The synthetic panel uses a separate
masking bank of 1K windows from the nonseasonal-dynamics preset scored
with lag-one scales.

\begin{figure}[!htbp]
\centering
\includegraphics[width=\linewidth]{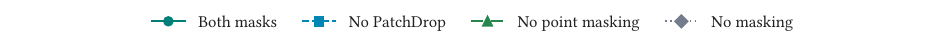}\par
\subcaptionbox{FEV-Bench-S\label{fig:missing-fev}}[0.30\linewidth]{\includegraphics[width=\linewidth]{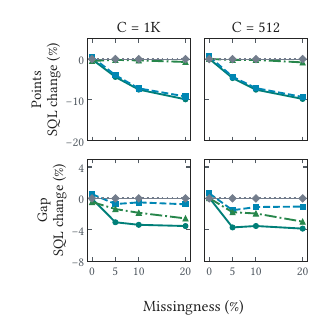}}
\hfill
\subcaptionbox{GIFT-Eval-S\label{fig:missing-gift}}[0.30\linewidth]{\includegraphics[width=\linewidth]{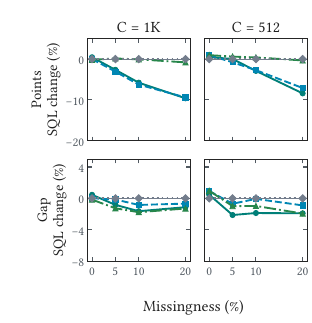}}
\hfill
\subcaptionbox{Synthetic masking bank\label{fig:missing-synthetic}}[0.30\linewidth]{\includegraphics[width=\linewidth]{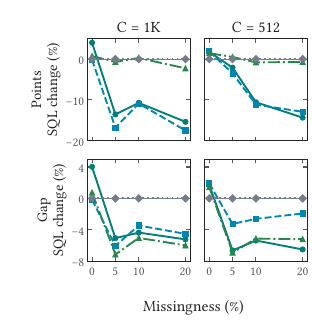}}
\par\smallskip
\caption{Sensitivity to missing history at epoch 400.
The curves compare each trained masking variant with the model trained
without masking, using the same test corruption in each comparison.
Negative values indicate lower SQL. Columns show context caps, and
rows show random points and contiguous gaps. A 0\% corruption rate
leaves the history unchanged. Seasonal scales are fixed to clean
history at each cap; the synthetic panel uses lag-one scales.}
\label{fig:missing-context}
\end{figure}

\paragraph{Mixup Strength with the Nonseasonal-Dynamics Mixture.}
\label{sec:ablation-mixing-strength}
This study uses a 104.7M-parameter patch-32 backbone with 1K
history steps and a 128-step horizon. The control uses three
contributors, $\alpha=0.15$, batch activation probability 0.50,
1\% point masking, and no PatchDrop. Changing the contributor
count at fixed $\alpha$ also changes the total concentration
$K\alpha$.

All variants use seed 42, one million windows per epoch, and the
same optimization schedule as the training augmentation study.
At epoch 200, each resumes with its optimizer and schedule and
replays the original 200 banks of 200,000 trajectories in the
same order. In total, each arm samples 400 million windows
from 40 million cached source rows. The window and augmentation
draws can change on replay. The original checkpoints did not
store the complete random-number state.

\begin{table}[!htbp]
\centering
\caption{Augmentation strength with the nonseasonal-dynamics mixture.
Each block varies one setting; changes are relative to the repeated
control in that block.}
\label{tab:ablation-mixing-strength}
\scriptsize
\setlength{\tabcolsep}{0.9pt}
\reporttablestyle[3]
\begin{tabular}{@{}lrrrrrr@{}}
\toprule
& \multicolumn{2}{c}{\reporttablehead{FEV-Bench-S}} &
\multicolumn{2}{c}{\reporttablehead{GIFT-Eval-S}} &
\multicolumn{2}{c}{\reporttablehead{Synthetic bank}} \\
\cmidrule(lr){2-3}\cmidrule(lr){4-5}\cmidrule(l){6-7}
\reporttablehead{Setting} &
\reporttablehead{SQL $\downarrow$} & \reporttablehead{MASE $\downarrow$} &
\reporttablehead{SQL $\downarrow$} & \reporttablehead{MASE $\downarrow$} &
\reporttablehead{SQL $\downarrow$} & \reporttablehead{MASE $\downarrow$} \\
\midrule
\reporttablegroup{7}{Contributor count ($k$)} \\
\addlinespace[2pt]
$k=3$ (default) & \shortstack[r]{$0.8322\pm.0058$\\[-1pt]{\scriptsize\textcolor{ablationneutral}{$\Delta$ +0.00\%}}} & \shortstack[r]{$1.2034\pm.0044$\\[-1pt]{\scriptsize\textcolor{ablationneutral}{$\Delta$ +0.00\%}}} & \shortstack[r]{$0.7217\pm.0014$\\[-1pt]{\scriptsize\textcolor{ablationneutral}{$\Delta$ +0.00\%}}} & \shortstack[r]{$1.0577\pm.0019$\\[-1pt]{\scriptsize\textcolor{ablationneutral}{$\Delta$ +0.00\%}}} & \shortstack[r]{$1.1979\pm.0057$\\[-1pt]{\scriptsize\textcolor{ablationneutral}{$\Delta$ +0.00\%}}} & \shortstack[r]{$1.7862\pm.0084$\\[-1pt]{\scriptsize\textcolor{ablationneutral}{$\Delta$ +0.00\%}}}  \\
$k=2$ & \shortstack[r]{$0.8374\pm.0242$\\[-1pt]{\scriptsize\textcolor{ablationworse}{$\Delta$ +0.63\%}}} & \shortstack[r]{$1.2047\pm.0051$\\[-1pt]{\scriptsize\textcolor{ablationworse}{$\Delta$ +0.12\%}}} & \shortstack[r]{$0.7200\pm.0017$\\[-1pt]{\scriptsize\textcolor{ablationbetter}{$\Delta$ -0.24\%}}} & \shortstack[r]{$1.0545\pm.0024$\\[-1pt]{\scriptsize\textcolor{ablationbetter}{$\Delta$ -0.30\%}}} & \shortstack[r]{$\mathbf{1.1797}\pm.0036$\\[-1pt]{\scriptsize\textcolor{ablationbetter}{$\Delta$ -1.52\%}}} & \shortstack[r]{$\mathbf{1.7635}\pm.0062$\\[-1pt]{\scriptsize\textcolor{ablationbetter}{$\Delta$ -1.27\%}}}  \\
$k=4$ & \shortstack[r]{$\mathbf{0.8255}\pm.0044$\\[-1pt]{\scriptsize\textcolor{ablationbetter}{$\Delta$ -0.80\%}}} & \shortstack[r]{$1.1994\pm.0070$\\[-1pt]{\scriptsize\textcolor{ablationbetter}{$\Delta$ -0.33\%}}} & \shortstack[r]{$0.7194\pm.0023$\\[-1pt]{\scriptsize\textcolor{ablationbetter}{$\Delta$ -0.32\%}}} & \shortstack[r]{$1.0548\pm.0031$\\[-1pt]{\scriptsize\textcolor{ablationbetter}{$\Delta$ -0.28\%}}} & \shortstack[r]{$1.1915\pm.0062$\\[-1pt]{\scriptsize\textcolor{ablationbetter}{$\Delta$ -0.54\%}}} & \shortstack[r]{$1.7740\pm.0100$\\[-1pt]{\scriptsize\textcolor{ablationbetter}{$\Delta$ -0.68\%}}}  \\
\midrule
\reporttablegroup{7}{Dirichlet concentration ($\alpha$)} \\
\addlinespace[2pt]
$\alpha=0.15$ (default) & \shortstack[r]{$0.8322\pm.0058$\\[-1pt]{\scriptsize\textcolor{ablationneutral}{$\Delta$ +0.00\%}}} & \shortstack[r]{$1.2034\pm.0044$\\[-1pt]{\scriptsize\textcolor{ablationneutral}{$\Delta$ +0.00\%}}} & \shortstack[r]{$0.7217\pm.0014$\\[-1pt]{\scriptsize\textcolor{ablationneutral}{$\Delta$ +0.00\%}}} & \shortstack[r]{$1.0577\pm.0019$\\[-1pt]{\scriptsize\textcolor{ablationneutral}{$\Delta$ +0.00\%}}} & \shortstack[r]{$1.1979\pm.0057$\\[-1pt]{\scriptsize\textcolor{ablationneutral}{$\Delta$ +0.00\%}}} & \shortstack[r]{$1.7862\pm.0084$\\[-1pt]{\scriptsize\textcolor{ablationneutral}{$\Delta$ +0.00\%}}}  \\
$\alpha=0.05$ & \shortstack[r]{$0.8281\pm.0026$\\[-1pt]{\scriptsize\textcolor{ablationbetter}{$\Delta$ -0.48\%}}} & \shortstack[r]{$1.2048\pm.0059$\\[-1pt]{\scriptsize\textcolor{ablationworse}{$\Delta$ +0.12\%}}} & \shortstack[r]{$0.7201\pm.0019$\\[-1pt]{\scriptsize\textcolor{ablationbetter}{$\Delta$ -0.23\%}}} & \shortstack[r]{$1.0551\pm.0026$\\[-1pt]{\scriptsize\textcolor{ablationbetter}{$\Delta$ -0.25\%}}} & \shortstack[r]{$1.1926\pm.0095$\\[-1pt]{\scriptsize\textcolor{ablationbetter}{$\Delta$ -0.44\%}}} & \shortstack[r]{$1.7789\pm.0145$\\[-1pt]{\scriptsize\textcolor{ablationbetter}{$\Delta$ -0.41\%}}}  \\
$\alpha=0.30$ & \shortstack[r]{$0.8271\pm.0054$\\[-1pt]{\scriptsize\textcolor{ablationbetter}{$\Delta$ -0.61\%}}} & \shortstack[r]{$1.1983\pm.0092$\\[-1pt]{\scriptsize\textcolor{ablationbetter}{$\Delta$ -0.42\%}}} & \shortstack[r]{$0.7206\pm.0022$\\[-1pt]{\scriptsize\textcolor{ablationbetter}{$\Delta$ -0.15\%}}} & \shortstack[r]{$1.0555\pm.0035$\\[-1pt]{\scriptsize\textcolor{ablationbetter}{$\Delta$ -0.21\%}}} & \shortstack[r]{$1.1994\pm.0066$\\[-1pt]{\scriptsize\textcolor{ablationworse}{$\Delta$ +0.12\%}}} & \shortstack[r]{$1.7902\pm.0113$\\[-1pt]{\scriptsize\textcolor{ablationworse}{$\Delta$ +0.22\%}}}  \\
\midrule
\reporttablegroup{7}{Mixing activation probability ($p_{\mathrm{mix}}$)} \\
\addlinespace[2pt]
$p_{\mathrm{mix}}=0.50$ (default) & \shortstack[r]{$0.8322\pm.0058$\\[-1pt]{\scriptsize\textcolor{ablationneutral}{$\Delta$ +0.00\%}}} & \shortstack[r]{$1.2034\pm.0044$\\[-1pt]{\scriptsize\textcolor{ablationneutral}{$\Delta$ +0.00\%}}} & \shortstack[r]{$0.7217\pm.0014$\\[-1pt]{\scriptsize\textcolor{ablationneutral}{$\Delta$ +0.00\%}}} & \shortstack[r]{$1.0577\pm.0019$\\[-1pt]{\scriptsize\textcolor{ablationneutral}{$\Delta$ +0.00\%}}} & \shortstack[r]{$1.1979\pm.0057$\\[-1pt]{\scriptsize\textcolor{ablationneutral}{$\Delta$ +0.00\%}}} & \shortstack[r]{$1.7862\pm.0084$\\[-1pt]{\scriptsize\textcolor{ablationneutral}{$\Delta$ +0.00\%}}}  \\
$p_{\mathrm{mix}}=0.75$ & \shortstack[r]{$0.8313\pm.0027$\\[-1pt]{\scriptsize\textcolor{ablationbetter}{$\Delta$ -0.11\%}}} & \shortstack[r]{$1.2098\pm.0040$\\[-1pt]{\scriptsize\textcolor{ablationworse}{$\Delta$ +0.53\%}}} & \shortstack[r]{$\mathbf{0.7187}\pm.0017$\\[-1pt]{\scriptsize\textcolor{ablationbetter}{$\Delta$ -0.41\%}}} & \shortstack[r]{$\mathbf{1.0515}\pm.0016$\\[-1pt]{\scriptsize\textcolor{ablationbetter}{$\Delta$ -0.59\%}}} & \shortstack[r]{$1.2092\pm.0055$\\[-1pt]{\scriptsize\textcolor{ablationworse}{$\Delta$ +0.94\%}}} & \shortstack[r]{$1.8047\pm.0092$\\[-1pt]{\scriptsize\textcolor{ablationworse}{$\Delta$ +1.04\%}}}  \\
$p_{\mathrm{mix}}=1.00$ & \shortstack[r]{$0.8446\pm.0079$\\[-1pt]{\scriptsize\textcolor{ablationworse}{$\Delta$ +1.50\%}}} & \shortstack[r]{$1.2234\pm.0100$\\[-1pt]{\scriptsize\textcolor{ablationworse}{$\Delta$ +1.66\%}}} & \shortstack[r]{$0.7256\pm.0024$\\[-1pt]{\scriptsize\textcolor{ablationworse}{$\Delta$ +0.54\%}}} & \shortstack[r]{$1.0623\pm.0028$\\[-1pt]{\scriptsize\textcolor{ablationworse}{$\Delta$ +0.43\%}}} & \shortstack[r]{$1.2598\pm.0082$\\[-1pt]{\scriptsize\textcolor{ablationworse}{$\Delta$ +5.17\%}}} & \shortstack[r]{$1.8796\pm.0113$\\[-1pt]{\scriptsize\textcolor{ablationworse}{$\Delta$ +5.23\%}}}  \\
\midrule
\reporttablegroup{7}{Point-mask probability ($p_{\mathrm{mask}}$)} \\
\addlinespace[2pt]
$p_{\mathrm{mask}}=0.01$ (default) & \shortstack[r]{$0.8322\pm.0058$\\[-1pt]{\scriptsize\textcolor{ablationneutral}{$\Delta$ +0.00\%}}} & \shortstack[r]{$1.2034\pm.0044$\\[-1pt]{\scriptsize\textcolor{ablationneutral}{$\Delta$ +0.00\%}}} & \shortstack[r]{$0.7217\pm.0014$\\[-1pt]{\scriptsize\textcolor{ablationneutral}{$\Delta$ +0.00\%}}} & \shortstack[r]{$1.0577\pm.0019$\\[-1pt]{\scriptsize\textcolor{ablationneutral}{$\Delta$ +0.00\%}}} & \shortstack[r]{$1.1979\pm.0057$\\[-1pt]{\scriptsize\textcolor{ablationneutral}{$\Delta$ +0.00\%}}} & \shortstack[r]{$1.7862\pm.0084$\\[-1pt]{\scriptsize\textcolor{ablationneutral}{$\Delta$ +0.00\%}}}  \\
$p_{\mathrm{mask}}=0.05$ & \shortstack[r]{$0.8295\pm.0032$\\[-1pt]{\scriptsize\textcolor{ablationbetter}{$\Delta$ -0.32\%}}} & \shortstack[r]{$\mathbf{1.1982}\pm.0053$\\[-1pt]{\scriptsize\textcolor{ablationbetter}{$\Delta$ -0.43\%}}} & \shortstack[r]{$0.7260\pm.0032$\\[-1pt]{\scriptsize\textcolor{ablationworse}{$\Delta$ +0.59\%}}} & \shortstack[r]{$1.0584\pm.0041$\\[-1pt]{\scriptsize\textcolor{ablationworse}{$\Delta$ +0.06\%}}} & \shortstack[r]{$1.2036\pm.0058$\\[-1pt]{\scriptsize\textcolor{ablationworse}{$\Delta$ +0.48\%}}} & \shortstack[r]{$1.7963\pm.0113$\\[-1pt]{\scriptsize\textcolor{ablationworse}{$\Delta$ +0.57\%}}}  \\
\bottomrule
\end{tabular}
\end{table}

\paragraph{Patch Length and Stride.}
\label{sec:ablation-patching}
The patch study uses the same cached continuation protocol.
At context 1K and horizon 128, the length-and-stride settings
16, 32, and 64 produce 73, 37, and 19 tokens, respectively.
Their parameter counts are 104,338,440, 104,707,176, and
105,444,648. The temporal resolution, token count, and parameter
count therefore vary together. Table~\ref{tab:ablation-patching} reports
the results.

\paragraph{Training Context Distribution.}
\label{sec:ablation-context-distribution}
The four context buckets are the same as in the mixing study.
From shortest to longest, equal sampling uses probabilities
$(0.25,0.25,0.25,0.25)$; moderate long bias uses
$(0.10,0.20,0.30,0.40)$; strong long bias uses
$(0.05,0.10,0.25,0.60)$; short bias uses
$(0.40,0.30,0.20,0.10)$. We keep the probabilities fixed
during training and retain the cached continuation protocol.

\begin{table}[!htbp]
\centering
\caption{Training context distribution. Changes are relative to equal
bucket weights. All models use up to 1K history steps at evaluation.}
\label{tab:ablation-context}
\scriptsize
\setlength{\tabcolsep}{0.9pt}
\reporttablestyle[3]
\begin{tabular}{@{}lrrrrrr@{}}
\toprule
& \multicolumn{2}{c}{\reporttablehead{FEV-Bench-S}} &
\multicolumn{2}{c}{\reporttablehead{GIFT-Eval-S}} &
\multicolumn{2}{c}{\reporttablehead{Synthetic bank}} \\
\cmidrule(lr){2-3}\cmidrule(lr){4-5}\cmidrule(l){6-7}
\reporttablehead{Setting} &
\reporttablehead{SQL $\downarrow$} & \reporttablehead{MASE $\downarrow$} &
\reporttablehead{SQL $\downarrow$} & \reporttablehead{MASE $\downarrow$} &
\reporttablehead{SQL $\downarrow$} & \reporttablehead{MASE $\downarrow$} \\
\midrule
Equal buckets & \shortstack[r]{$0.8322\pm.0058$\\[-1pt]{\scriptsize\textcolor{ablationneutral}{$\Delta$ +0.00\%}}} & \shortstack[r]{$1.2034\pm.0044$\\[-1pt]{\scriptsize\textcolor{ablationneutral}{$\Delta$ +0.00\%}}} & \shortstack[r]{$0.7217\pm.0014$\\[-1pt]{\scriptsize\textcolor{ablationneutral}{$\Delta$ +0.00\%}}} & \shortstack[r]{$1.0577\pm.0019$\\[-1pt]{\scriptsize\textcolor{ablationneutral}{$\Delta$ +0.00\%}}} & \shortstack[r]{$1.1979\pm.0057$\\[-1pt]{\scriptsize\textcolor{ablationneutral}{$\Delta$ +0.00\%}}} & \shortstack[r]{$1.7862\pm.0084$\\[-1pt]{\scriptsize\textcolor{ablationneutral}{$\Delta$ +0.00\%}}}  \\
\midrule
Moderate long bias & \shortstack[r]{$\mathbf{0.8222}\pm.0049$\\[-1pt]{\scriptsize\textcolor{ablationbetter}{$\Delta$ -1.19\%}}} & \shortstack[r]{$\mathbf{1.1999}\pm.0092$\\[-1pt]{\scriptsize\textcolor{ablationbetter}{$\Delta$ -0.29\%}}} & \shortstack[r]{$\mathbf{0.7150}\pm.0021$\\[-1pt]{\scriptsize\textcolor{ablationbetter}{$\Delta$ -0.93\%}}} & \shortstack[r]{$\mathbf{1.0487}\pm.0027$\\[-1pt]{\scriptsize\textcolor{ablationbetter}{$\Delta$ -0.86\%}}} & \shortstack[r]{$1.1688\pm.0067$\\[-1pt]{\scriptsize\textcolor{ablationbetter}{$\Delta$ -2.42\%}}} & \shortstack[r]{$1.7542\pm.0085$\\[-1pt]{\scriptsize\textcolor{ablationbetter}{$\Delta$ -1.79\%}}}  \\
Strong long bias & \shortstack[r]{$0.8247\pm.0023$\\[-1pt]{\scriptsize\textcolor{ablationbetter}{$\Delta$ -0.90\%}}} & \shortstack[r]{$1.2015\pm.0027$\\[-1pt]{\scriptsize\textcolor{ablationbetter}{$\Delta$ -0.15\%}}} & \shortstack[r]{$0.7236\pm.0018$\\[-1pt]{\scriptsize\textcolor{ablationworse}{$\Delta$ +0.26\%}}} & \shortstack[r]{$1.0617\pm.0025$\\[-1pt]{\scriptsize\textcolor{ablationworse}{$\Delta$ +0.38\%}}} & \shortstack[r]{$\mathbf{1.1649}\pm.0044$\\[-1pt]{\scriptsize\textcolor{ablationbetter}{$\Delta$ -2.75\%}}} & \shortstack[r]{$\mathbf{1.7491}\pm.0064$\\[-1pt]{\scriptsize\textcolor{ablationbetter}{$\Delta$ -2.07\%}}}  \\
Short bias & \shortstack[r]{$0.8297\pm.0012$\\[-1pt]{\scriptsize\textcolor{ablationbetter}{$\Delta$ -0.30\%}}} & \shortstack[r]{$1.2079\pm.0031$\\[-1pt]{\scriptsize\textcolor{ablationworse}{$\Delta$ +0.37\%}}} & \shortstack[r]{$0.7251\pm.0018$\\[-1pt]{\scriptsize\textcolor{ablationworse}{$\Delta$ +0.46\%}}} & \shortstack[r]{$1.0610\pm.0019$\\[-1pt]{\scriptsize\textcolor{ablationworse}{$\Delta$ +0.31\%}}} & \shortstack[r]{$1.2447\pm.0045$\\[-1pt]{\scriptsize\textcolor{ablationworse}{$\Delta$ +3.91\%}}} & \shortstack[r]{$1.8612\pm.0079$\\[-1pt]{\scriptsize\textcolor{ablationworse}{$\Delta$ +4.20\%}}}  \\
\bottomrule
\end{tabular}
\end{table}

\paragraph{Evaluation across Context Caps.}
\label{sec:context-caps}
We also score each variant using 128, 256, 512, and 1K
visible steps. All caps retain the same eligible windows, targets,
and seasonal scales computed from the full 1K-step history.
Figure~\ref{fig:context-caps} shows changes relative to equal
sampling at each cap. At 128 visible steps, moderate long bias lowers
SQL on GIFT-Eval-S by 13.30\% and MASE on FEV-Bench-S by 0.07\%.

\paragraph{Architecture Variants.}
\label{sec:architecture-ablation}
Each arm changes one operation in a common backbone. Query--key
normalization rescales attention queries and keys. Attention gating
applies a learned sigmoid gate to each head output. The value
residual mixes first-layer and current-layer values; scalar and
dynamic residuals rescale the residual updates with learned
scalars or input-dependent feature gates. SwiGLU changes the
feed-forward activation and also changes capacity through its
width and the shared embedding layers.

The architecture arms use the reference generator mixture,
seed 42, the four equally weighted context buckets, a 128-step
horizon, three-way mixup, 1\% point masking, and no PatchDrop.
They share the same 8K batch size and optimization schedule,
resuming from epoch 200 with optimizer and scheduler state
through epoch 400. Evaluation uses FP32 with TF32 disabled.

\begin{table}[!htbp]
\centering
\caption{Architecture variants. Each row adds one option to the control.
Changes are relative to that control. Parameter counts include all
model components.}
\label{tab:architecture-ablation}\scriptsize
\setlength{\tabcolsep}{0.9pt}
\renewcommand{\arraystretch}{1.0}
\reporttablestyle[3]
\begin{tabular}{@{}>{\raggedright\arraybackslash}p{2.25cm}rrrrrrr@{}}
\toprule
& & \multicolumn{2}{c}{\reporttablehead{FEV-Bench-S}} &
\multicolumn{2}{c}{\reporttablehead{GIFT-Eval-S}} &
\multicolumn{2}{c}{\reporttablehead{Synthetic bank}} \\
\cmidrule(lr){3-4}\cmidrule(lr){5-6}\cmidrule(l){7-8}
\reporttablehead{Architecture} & \reporttablehead{\shortstack{Params\\(M)}} &
\reporttablehead{SQL} & \reporttablehead{MASE} &
\reporttablehead{SQL} & \reporttablehead{MASE} &
\reporttablehead{SQL} & \reporttablehead{MASE} \\
\midrule
Control & 104.58 & \shortstack[r]{$0.8309\pm.0046$\\[-1pt]{\scriptsize\textcolor{ablationneutral}{$\Delta$ +0.00\%}}} & \shortstack[r]{$1.2066\pm.0058$\\[-1pt]{\scriptsize\textcolor{ablationneutral}{$\Delta$ +0.00\%}}} & \shortstack[r]{$0.7191\pm.0021$\\[-1pt]{\scriptsize\textcolor{ablationneutral}{$\Delta$ +0.00\%}}} & \shortstack[r]{$1.0552\pm.0027$\\[-1pt]{\scriptsize\textcolor{ablationneutral}{$\Delta$ +0.00\%}}} & \shortstack[r]{$1.2042\pm.0075$\\[-1pt]{\scriptsize\textcolor{ablationneutral}{$\Delta$ +0.00\%}}} & \shortstack[r]{$1.8034\pm.0125$\\[-1pt]{\scriptsize\textcolor{ablationneutral}{$\Delta$ +0.00\%}}}  \\
QK normalization & 104.58 & \shortstack[r]{$0.8343\pm.0038$\\[-1pt]{\scriptsize\textcolor{ablationworse}{$\Delta$ +0.41\%}}} & \shortstack[r]{$1.2133\pm.0056$\\[-1pt]{\scriptsize\textcolor{ablationworse}{$\Delta$ +0.55\%}}} & \shortstack[r]{$0.7222\pm.0025$\\[-1pt]{\scriptsize\textcolor{ablationworse}{$\Delta$ +0.43\%}}} & \shortstack[r]{$1.0584\pm.0039$\\[-1pt]{\scriptsize\textcolor{ablationworse}{$\Delta$ +0.30\%}}} & \shortstack[r]{$1.2094\pm.0121$\\[-1pt]{\scriptsize\textcolor{ablationworse}{$\Delta$ +0.43\%}}} & \shortstack[r]{$1.8014\pm.0211$\\[-1pt]{\scriptsize\textcolor{ablationbetter}{$\Delta$ -0.11\%}}}  \\
Attention gating & 104.71 & \shortstack[r]{$0.8370\pm.0115$\\[-1pt]{\scriptsize\textcolor{ablationworse}{$\Delta$ +0.73\%}}} & \shortstack[r]{$1.2083\pm.0072$\\[-1pt]{\scriptsize\textcolor{ablationworse}{$\Delta$ +0.14\%}}} & \shortstack[r]{$0.7215\pm.0016$\\[-1pt]{\scriptsize\textcolor{ablationworse}{$\Delta$ +0.34\%}}} & \shortstack[r]{$1.0581\pm.0015$\\[-1pt]{\scriptsize\textcolor{ablationworse}{$\Delta$ +0.27\%}}} & \shortstack[r]{$1.1948\pm.0112$\\[-1pt]{\scriptsize\textcolor{ablationbetter}{$\Delta$ -0.78\%}}} & \shortstack[r]{$1.7821\pm.0160$\\[-1pt]{\scriptsize\textcolor{ablationbetter}{$\Delta$ -1.18\%}}}  \\
Value residual & 104.58 & \shortstack[r]{$0.8314\pm.0085$\\[-1pt]{\scriptsize\textcolor{ablationworse}{$\Delta$ +0.05\%}}} & \shortstack[r]{$1.2074\pm.0142$\\[-1pt]{\scriptsize\textcolor{ablationworse}{$\Delta$ +0.06\%}}} & \shortstack[r]{$\mathbf{0.7182}\pm.0034$\\[-1pt]{\scriptsize\textcolor{ablationbetter}{$\Delta$ -0.12\%}}} & \shortstack[r]{$\mathbf{1.0524}\pm.0044$\\[-1pt]{\scriptsize\textcolor{ablationbetter}{$\Delta$ -0.27\%}}} & \shortstack[r]{$1.1885\pm.0091$\\[-1pt]{\scriptsize\textcolor{ablationbetter}{$\Delta$ -1.31\%}}} & \shortstack[r]{$1.7702\pm.0168$\\[-1pt]{\scriptsize\textcolor{ablationbetter}{$\Delta$ -1.84\%}}}  \\
Scalar residual & 104.58 & \shortstack[r]{$\mathbf{0.8302}\pm.0038$\\[-1pt]{\scriptsize\textcolor{ablationbetter}{$\Delta$ -0.09\%}}} & \shortstack[r]{$\mathbf{1.2044}\pm.0056$\\[-1pt]{\scriptsize\textcolor{ablationbetter}{$\Delta$ -0.18\%}}} & \shortstack[r]{$0.7189\pm.0014$\\[-1pt]{\scriptsize\textcolor{ablationbetter}{$\Delta$ -0.03\%}}} & \shortstack[r]{$1.0549\pm.0015$\\[-1pt]{\scriptsize\textcolor{ablationbetter}{$\Delta$ -0.03\%}}} & \shortstack[r]{$1.2086\pm.0159$\\[-1pt]{\scriptsize\textcolor{ablationworse}{$\Delta$ +0.36\%}}} & \shortstack[r]{$1.8056\pm.0257$\\[-1pt]{\scriptsize\textcolor{ablationworse}{$\Delta$ +0.12\%}}}  \\
Dynamic residual & 121.11 & \shortstack[r]{$0.8370\pm.0082$\\[-1pt]{\scriptsize\textcolor{ablationworse}{$\Delta$ +0.73\%}}} & \shortstack[r]{$1.2069\pm.0070$\\[-1pt]{\scriptsize\textcolor{ablationworse}{$\Delta$ +0.02\%}}} & \shortstack[r]{$0.7217\pm.0033$\\[-1pt]{\scriptsize\textcolor{ablationworse}{$\Delta$ +0.35\%}}} & \shortstack[r]{$1.0599\pm.0040$\\[-1pt]{\scriptsize\textcolor{ablationworse}{$\Delta$ +0.45\%}}} & \shortstack[r]{$1.2081\pm.0111$\\[-1pt]{\scriptsize\textcolor{ablationworse}{$\Delta$ +0.32\%}}} & \shortstack[r]{$1.8091\pm.0191$\\[-1pt]{\scriptsize\textcolor{ablationworse}{$\Delta$ +0.32\%}}}  \\
SwiGLU ($d_{\textit{ff}}=2\mathrm{K}$) & 102.81 & \shortstack[r]{$0.8319\pm.0041$\\[-1pt]{\scriptsize\textcolor{ablationworse}{$\Delta$ +0.11\%}}} & \shortstack[r]{$1.2126\pm.0072$\\[-1pt]{\scriptsize\textcolor{ablationworse}{$\Delta$ +0.49\%}}} & \shortstack[r]{$0.7215\pm.0036$\\[-1pt]{\scriptsize\textcolor{ablationworse}{$\Delta$ +0.33\%}}} & \shortstack[r]{$1.0577\pm.0047$\\[-1pt]{\scriptsize\textcolor{ablationworse}{$\Delta$ +0.24\%}}} & \shortstack[r]{$\mathbf{1.1765}\pm.0077$\\[-1pt]{\scriptsize\textcolor{ablationbetter}{$\Delta$ -2.30\%}}} & \shortstack[r]{$\mathbf{1.7585}\pm.0120$\\[-1pt]{\scriptsize\textcolor{ablationbetter}{$\Delta$ -2.49\%}}}  \\
SwiGLU ($d_{\textit{ff}}=3\mathrm{K}$) & 137.61 & \shortstack[r]{$0.8310\pm.0068$\\[-1pt]{\scriptsize\textcolor{ablationworse}{$\Delta$ +0.01\%}}} & \shortstack[r]{$1.2066\pm.0100$\\[-1pt]{\scriptsize\textcolor{ablationneutral}{$\Delta$ +0.00\%}}} & \shortstack[r]{$0.7238\pm.0026$\\[-1pt]{\scriptsize\textcolor{ablationworse}{$\Delta$ +0.65\%}}} & \shortstack[r]{$1.0601\pm.0035$\\[-1pt]{\scriptsize\textcolor{ablationworse}{$\Delta$ +0.46\%}}} & \shortstack[r]{$1.1862\pm.0085$\\[-1pt]{\scriptsize\textcolor{ablationbetter}{$\Delta$ -1.49\%}}} & \shortstack[r]{$1.7806\pm.0147$\\[-1pt]{\scriptsize\textcolor{ablationbetter}{$\Delta$ -1.26\%}}}  \\
\bottomrule
\end{tabular}
\end{table}

\section{Classification with Frozen \textnormal{\textsc{Tafsut}} Representations}
\label{app:ucr-classification}

\paragraph{Methodology.}
We follow the frozen-feature evaluation protocol of
\citet{feofanov2026mantisv2} on the 128 UCR datasets~\citep{dau2019ucr},
including the 15 adjusted tasks and the official train and test splits.
On each training split, we fit a feature standardizer followed by
logistic regression with a 500-iteration cap. We freeze
\textsc{Tafsut}-Synth and \textsc{Tafsut}-GIFT, selecting the latter by
normalized MASE on GIFT-Eval-S. We resize each series to 1K samples by linear
interpolation and concatenate the coordinatewise maxima over the context
tokens at each of the 14 Transformer layers. This yields 10,752 features.

\begin{figure}[!htbp]
\centering
\captionsetup{skip=3pt}
\includegraphics[width=0.70\linewidth]{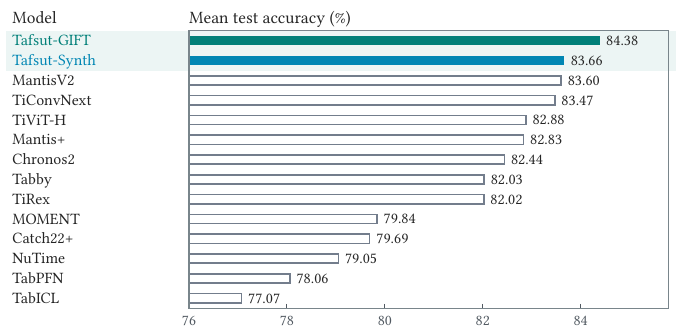}
\caption{Mean test accuracy across the 128 UCR datasets, with equal
weight per dataset. }
\label{fig:ucr-classification}
\end{figure}

\paragraph{Baselines.}
We evaluate Tabby~\citep{xie2026tabby} with the same downstream classifier
and its original 15,360-dimensional readout, about $1.43\times$ the
dimensionality of \textsc{Tafsut}. We reproduce MantisV2's published mean
test accuracy of 83.60\% to the reported precision using its 512-sample
inputs and third-layer readout, which concatenates the class token and
the mean context state. For the comparison in
Figure~\ref{fig:ucr-classification}, we use the published scores from Figure~1 of
\citet{feofanov2026mantisv2} for Mantis+ and MantisV2,
TiViT-H and TiConvNext~\citep{roschmann2025visionts},
Chronos-2~\citep{ansari2025chronos2}, TiRex~\citep{auer2025tirex},
MOMENT~\citep{goswami2024moment}, NuTime~\citep{lin2024nutime},
Catch22+~\citep{lubba2019catch22}, TabPFN~\citep{hollmann2022tabpfn},
and TabICL~\citep{qu2025tabicl}.

\paragraph{Discussion.}
\textsc{Tafsut}-Synth reaches 83.66\% mean test accuracy, close to MantisV2
at 83.60\%. \textsc{Tafsut}-GIFT reaches 84.38\%, and Tabby scores 82.03\%.
These results show that a backbone trained only to forecast synthetic series
can learn features that transfer to classification. 

\section{From Zero-Shot Forecasting to Test-Time Training}
\label{app:gift-ttt}

\paragraph{Adaptation Protocol.}
We evaluate in-domain adaptation on 371,330 forecast windows across
97 GIFT-Eval tasks. Each batch contains up to 128 windows from one task
and shares up to 20 updates of the last four encoder blocks, final encoder
norm, and output head of the pretrained \textsc{Tafsut}-GIFT model
(30,065,560 parameters). For a window with observed history length $C_i$
capped at 8K and benchmark horizon $H_i$, we predict the last
$h_i=\min(H_i,\lfloor C_i/2\rfloor,1024)$ history steps from the preceding
prefix. Each update averages the masked, normalized quantile loss across
eligible windows, with two-donor synthetic mixup applied to each example
with probability $0.5$. After each update, we forecast the benchmark
horizon using up to 32K history steps and 64 output patches. Windows
without a valid history split retain their frozen forecasts. Benchmark
targets and seasonal scales are supplied to the scorer. We test two
batching protocols. \emph{Shared-cutoff TTT} groups windows by task and exact
forecast cutoff. This protocol is leakage-free with respect to temporal
look-ahead within a batch because every history is available at the common
cutoff. \emph{Benchmark-order TTT} groups consecutive windows in the stored
order of each task and permits temporal leakage. For two windows of the
same series with different cutoffs, the later window can contain the
earlier window's forecast targets in its history. The shared update can
then pass information from those observations to the earlier forecast.
Of 2,961 benchmark-order batches, 2,853 mix cutoffs. Before mixup, the
historical pseudo-target of the later window covers the full scored horizon
of the earlier window in 93,467 within-batch pairs.

\begin{figure}[!htbp]
\centering
\captionsetup{skip=3pt}
\includegraphics[width=\linewidth]{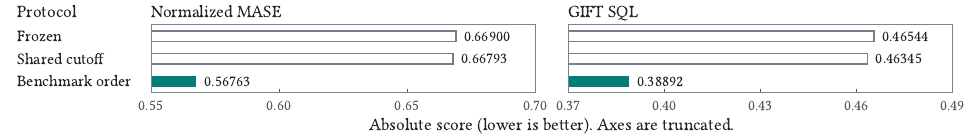}
\caption{Full GIFT-Eval scores for the frozen baseline and the two
test-time training protocols after 20 updates. Benchmark-order TTT permits
temporal look-ahead. Bar endpoints show the absolute scores, with the
lowest score in each panel highlighted in teal. Both metrics use
equal-task geometric aggregation after Seasonal Naive normalization.}
\label{fig:gift-ttt-best-step}
\end{figure}

\paragraph{Discussion.}
After 20 updates, normalized MASE is 0.66793 for shared-cutoff TTT and
0.56763 for benchmark-order TTT, compared with 0.669 for the frozen
model (Figure~\ref{fig:gift-ttt-best-step}). The small gain under shared
cutoff suggests limited benefit from leakage-free TTT in this setting.
The much larger gain in this experiment occurs under benchmark-order
batching, where later histories can inform earlier forecasts. The extent
of this exposure depends on how a benchmark batches its rolling windows.
Batch size and peer composition also differ
(76.5 versus 125.4 windows on average), so the comparison does not isolate
the contribution of look-ahead. Cross-learning~\citep{google2026timesfm3,ansari2025chronos2,meyer2026t0} can transfer information
between windows during inference even with frozen weights. It is valid
when the information shared with each forecast is available at its
cutoff. Rankings should therefore state whether forecasts are independent
or use cross-window information, and verify cutoff alignment even for
systems evaluated in zero-shot mode.

\end{document}